\documentclass{article}

    \PassOptionsToPackage{numbers, compress}{natbib}

 \usepackage{amsmath}

\usepackage[eandd, final]{neurips_2026}

\usepackage[utf8]{inputenc} %
\usepackage[T1]{fontenc}    %
\usepackage{url}            %
\usepackage{booktabs}       %
\usepackage{amsfonts}       %
\usepackage{nicefrac}       %
\usepackage{microtype}      %
\usepackage{xcolor}         %
\usepackage[utf8]{inputenc}
\usepackage[T1]{fontenc}
\usepackage[pagebackref,breaklinks,colorlinks,citecolor=black,linkcolor=red]{hyperref}
\usepackage{url}
\usepackage{booktabs}
\usepackage{amsfonts}
\usepackage{amsmath}
\usepackage{amssymb}
\usepackage{nicefrac}
\usepackage{microtype}
\usepackage{xcolor}
\usepackage{graphicx}
\usepackage{subcaption}
\usepackage{multirow}
\usepackage{makecell}
\usepackage{array}
\usepackage{wrapfig}
\usepackage{enumitem}
\usepackage{xspace}

\newcommand{\shadow}{\textsc{ShadowTransfer}\xspace}
\newcommand{\loco}{\textsc{LOCO}\xspace}
\newcommand{\sib}{\textsc{SIB}\xspace}

\newcolumntype{C}[1]{>{\centering\arraybackslash}p{#1}}
\newcolumntype{L}[1]{>{\raggedright\arraybackslash}p{#1}}

\title{ShadowTransfer: A Geographic Transfer Benchmark\\
for Overhead Shadow Detection}

\author{
Shagun Mittal \quad\quad Raymond A. Yeh\thanks{Equal senior advising} \quad\quad Satish Ukkusuri\footnotemark[1]\\
Purdue University\\
\texttt{\{mittal53, rayyeh, sukkusur\}@purdue.edu}
}

\begin{document}

\maketitle

\begin{abstract}
Overhead aerial imagery feeds dense-prediction pipelines that are routinely trained in one geographic region and deployed in another, yet no benchmark measures how shadow detection models transfer across regions. Existing overhead shadow datasets either lack a geographic transfer protocol or rely on automatically generated labels. We introduce \shadow, a benchmark of $4{,}500$ human-verified shadow masks across three U.S.\ cities at two native resolutions, paired with a leave-one-city-out evaluation protocol. %
Under vanilla training, holding out the test city costs $3.5$ to $7.8$ mIoU across four architectures, the gap remains significant when the training set grows from $225$ to $900$ images, and six of eight adaptation methods degrade more than vanilla training on average.
We propose a three-probe framework that diagnoses where and how the transfer breaks: the failure operates in feature space, beyond the reach of pixel-histogram matching, shows up as a confidence collapse on shadow pixels while background confidence stays intact, and is recoverable by retraining different architectural components depending on the model: encoder for some, decoder for others. These findings replicate on INRIA building footprints, suggesting they describe geographic transfer beyond shadows. 
Guided by the diagnosis, we build Spectral Information Bottleneck (SIB), a plug-in module with one component per finding. Across the $9$ (architecture, held-out city) cells, \sib{} has the smallest mean gap to the within-city upper bound of the ten methods we evaluate ($6.42$ mIoU), level with FDA. 
The benchmark, LOCO protocol, code, and trained models are released to support reproducible evaluation of geographic transfer in overhead dense prediction.
\end{abstract}

\section{Introduction}
\label{sec:intro}

Overhead aerial imagery feeds segmentation pipelines that are deployed across geographic regions. Shadow masks are a core input to several of these pipelines: solar irradiance estimation uses them to separate shaded from sunlit surfaces \cite{batchu2024satellite}, urban heat mapping uses them to correct surface temperature readings \cite{dashadebench, ukkusuri2024heatwaves}, and building footprint extraction uses them to resolve occlusion boundaries \cite{seo}. These pipelines are routinely trained on data from one region and deployed in another~\cite{mittal2025overcoming, verma2025spatial}.

Geographic shift is a known failure mode in segmentation models across domains. Medical imaging models underperform across hospitals due to scanner and protocol differences~\cite{ogier2022flamby}, and aerial land-cover models show similar drops on unseen regions~\cite{wang2021loveda, gong2025crossearth}. For overhead shadow detection, no comparable evaluation infrastructure exists. AISD, the largest human-verified overhead shadow dataset, contains 514 images and defines no train-test split across cities~\cite{aisd}. S-EO scales further but relies on automatically generated masks with documented alignment artifacts, and again reports no geographic holdout~\cite{seo}. Other widely used shadow benchmarks (SBU, ISTD) target ground-level imagery and do not match overhead geometry~\cite{sbu, istd}. Table~\ref{tab:dataset_comparison} summarizes the landscape. Without a benchmark that pairs human-verified labels with a defined geographic protocol, transfer failure in overhead shadow detection cannot be measured, diagnosed, or addressed.

\begin{table}[t]
  \centering
  \caption{
    \textbf{Comparison of shadow detection datasets.}
    No existing overhead dataset combines human-verified labels, multiple cities, multiple native resolutions, and a defined geographic transfer protocol.
    \shadow{} fills this gap.
    ISTD is included for context only; ground-level geometry is not a substitute for overhead evaluation.
    ``Transfer protocol'' = a defined geographic train-test split with formal evaluation guidelines.
    $\checkmark$~=~present; \texttimes{}~=~absent; n/r~=~not reported.
  }
  \label{tab:dataset_comparison}
  \small
  \setlength{\tabcolsep}{4pt}
  \begin{tabular}{@{}lllcccc@{}}
    \toprule
    Dataset & Domain & Scale & Human QC & Cities & Multi-res & Transfer protocol \\
    \midrule
    ISTD~\cite{istd}                & Ground-level             & 1{,}870     & $\checkmark$         & --          & \texttimes & \texttimes \\
    AISD~\cite{aisd}                & Aerial 0.3\,m   & 514         & $\checkmark$         & 5           & \texttimes & \texttimes \\
    Airborne-Shadow~\cite{azimi2023airborne} & Aerial & n/r         & $\checkmark$         & DE only     & \texttimes & \texttimes \\
    S-EO~\cite{seo}                 & Satellite 0.3\,m & $\sim$19{,}000 & \texttimes\ (auto) & 3 (US)      & \texttimes & \texttimes \\
    \midrule
    \shadow{} (ours)                & Aerial 0.3/0.6\,m & 4{,}500   & $\checkmark$         & 3 (US)      & $\checkmark$ & $\checkmark$\ \loco \\
    \bottomrule
  \end{tabular}
\end{table}
\begin{figure}[t]
  \centering
  \includegraphics[width=\linewidth]{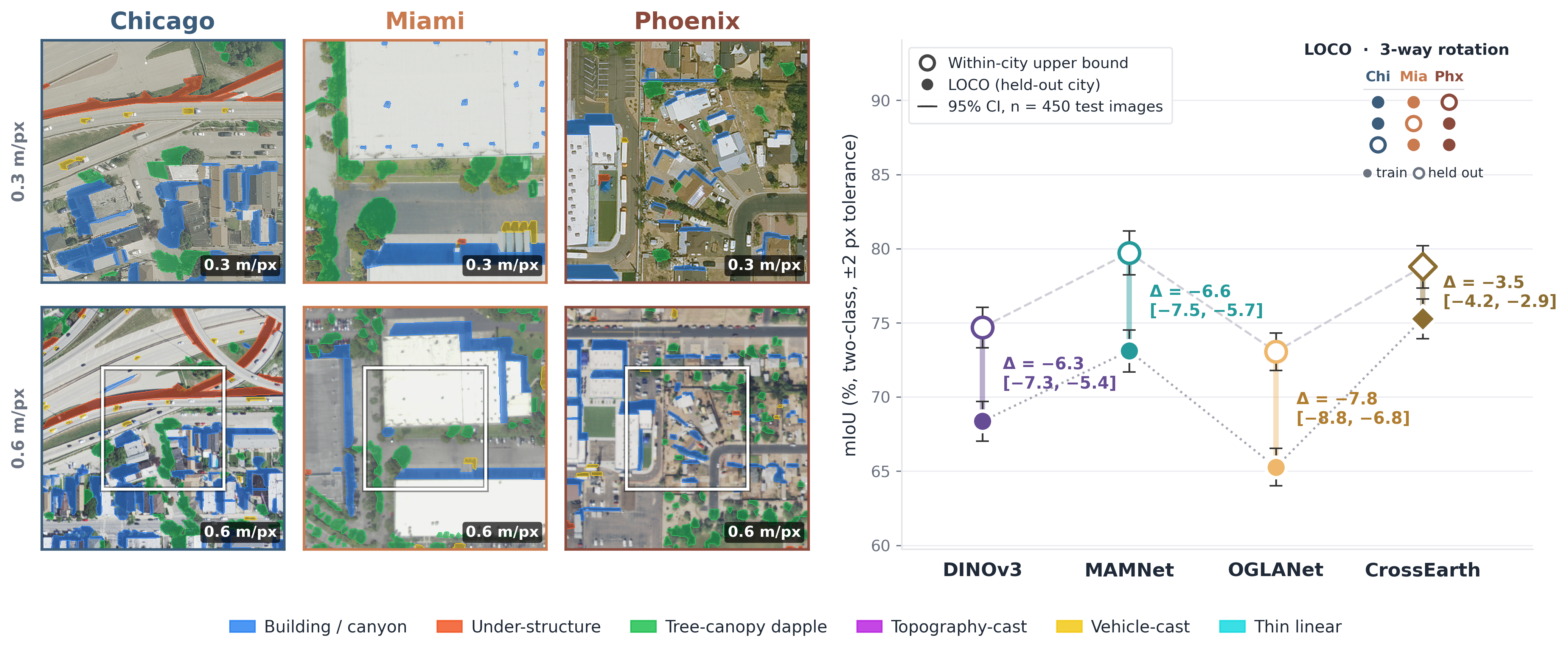}
  \caption{
    \textbf{\shadow and the geographic transfer gap.}
    \emph{Left:} paired test patches from three U.S.\ cities at two native NAIP
    resolutions ($0.3$\,m/px, $0.6$\,m/px); each column shares ground
    coordinates between rows, and the white box on each $0.6$\,m/px tile
    marks the paired $0.3$\,m/px footprint. Shadow masks are human-verified
    across six categories (bottom legend).
    \emph{Right:} two-class boundary-tolerant mIoU under vanilla training on the high-resolution cells, with each city held out in turn and results averaged over the three rotations (inset). Open marker $=$ within-city upper bound; filled marker $=$ LOCO held-out city; $\Delta =$ mean gap; error bars are $95\%$ CIs. Methods: Table~\ref{tab:loco-main}.
}
  \label{fig:teaser}
\end{figure}

To bridge this gap, we introduce \shadow, a benchmark for measuring geographic transfer in overhead shadow detection. It contains 4{,}500 human-verified shadow masks across three U.S. cities (Chicago, Miami, Phoenix) at two native sensor resolutions (0.3 m and 0.6 m), with a leave-one-city-out (LOCO) protocol that trains on two cities and tests on the third. 

Figure~\ref{fig:teaser} shows the size of the gap this protocol exposes: under vanilla training on the high-resolution cells, holding out the test city costs between $3.5$ mIoU (CrossEarth) and $7.8$ mIoU (OGLANet), and existing domain-adaptation methods help on some architecture-city pairs and fail catastrophically on others. The major \textbf{contributions} of this work are threefold:

\begin{enumerate}
    \item We release \shadow{}\footnote{Data: \url{https://huggingface.co/datasets/shagunmittal/ShadowTransfer} (DOI \href{https://doi.org/10.57967/hf/11483}{10.57967/hf/11483}); code: \url{https://github.com/shagun8/shadowtransfer}; project page: \url{https://shagun8.github.io/shadowtransfer/}; full datasheet in App.~\ref{app:datasheet}.} with its LOCO benchmarking and statistical reporting. 
    \item We provide a three-probe evaluation framework that characterizes \emph{how} models fail under geographic transfer: \textit{where} in the pipeline the shift acts, \textit{how} the prediction distribution distorts, and \textit{which} architectural components carry the failure
    \item We present \sib, a case-study module designed from the framework's findings, with the smallest mean gap to the within-city upper bound of the ten methods we evaluate, level with FDA. Together, the benchmark and framework give practitioners a way to measure geographic transfer risk before deployment and to diagnose where in a model that risk concentrates.
\end{enumerate}

\section{Related Work}
\label{sec:related}

\paragraph{Shadow detection in overhead imagery.}
Overhead shadow detection assigns a binary shadow label to every pixel of an aerial or satellite image. The field was established by AISD~\cite{aisd}, a 514-image manually annotated benchmark spanning five cities at 0.3\,m resolution. S-EO~\cite{seo} scales to roughly 19{,}000 images by deriving masks from digital surface models and sun geometry; the authors document temporal-misalignment artifacts and recommend the dataset for training rather than evaluation. Airborne-Shadow~\cite{azimi2023airborne} adds a manually annotated overhead set with a heavy/light shadow distinction, but is restricted to German imagery and does not report a geographic split. Recent methods (MAMNet, MBCSANet, CSDFormer, OGLANet) report F1 scores within roughly one point of each other on AISD~\cite{mamnet, mbcsanet, csdformer, oglanet}, indicating that the in-domain benchmark has plateaued. Additionally, ground-level shadow datasets~\cite{istd, sbu, cuhkshadow} are not substitutes for overhead evaluation: viewing angle, and shadow geometry differ. None of these overhead datasets defines a geographic train-test split, and the most recent overhead shadow survey identifies cross-dataset generalization as an open problem~\cite{hu2026unveiling}.

\paragraph{Domain adaptation for dense prediction.}
A range of methods make dense prediction robust to source-target appearance differences by intervening at one stage of the pipeline. At the input, FDA swaps low-frequency Fourier amplitude between images~\cite{fda}, IIM derives illumination-invariant inputs~\cite{yola}, and BSM mixes styles within a training batch through a frozen AdaIN network~\cite{luo2023bsm}. In feature space, ISW whitens photometrically sensitive covariance dimensions~\cite{isw}, SegDeSiC aligns encoder features through a geo-coordinate regression loss~\cite{segdesicnet}, MRFP$^+$ randomizes high- and low-frequency feature components~\cite{mrfp}, and FADA processes the Haar-wavelet bands of foundation-model features separately~\cite{fada}. At test time, TENT minimises prediction entropy over normalisation-layer parameters~\cite{wang2021tent}, and SAR stabilises this update with sharpness-aware minimisation and by excluding high-entropy samples~\cite{niu2023sar}. Each method applies its correction at the same locus for every architecture, and no prior work tests whether that locus is consistent across architectures for overhead dense prediction. Remote-sensing foundation models address the shift through the representation instead: CrossEarth trains adapters and a Mask2Former decoder on a frozen DINOv2 backbone, with style augmentation from an external image pool and masked image modelling~\cite{gong2025crossearth}.

\section{The \shadow Benchmark}
\label{sec:benchmark}

\subsection{Dataset Design and City Selection}
\label{sec:design}

\shadow contains $4{,}500$ RGB patches of size $384{\times}384$ pixels, sampled from National Agriculture Imagery Program (NAIP) orthoimagery across three U.S.\ cities: Chicago, Miami, and Phoenix. Each city contributes $1{,}500$ patches, split evenly across two native NAIP resolutions, $0.3$\,m and $0.6$\,m ground sampling distance ($750$ patches per city$\times$resolution pair). The three cities span distinct urban morphologies and climatic regimes, so the leave-one-city-out protocol in Section~\ref{sec:loco} measures transfer across genuine appearance shifts rather than nearby variants of the same scene type. Other potential confounders (sensor type, image format, file conventions) are held fixed by the shared NAIP source. We retain the two native resolutions rather than synthetically downsampling $0.3$\,m imagery, since NAIP releases them as separate native acquisitions with different sensor characteristics; collapsing these differences would understate the gap deployed systems face. Preserving both resolutions also supports future study of resolution shift under the same protocol. Within each city, patches are sampled uniformly inside the administrative boundary, with at most one patch per source NAIP tile to limit spatial autocorrelation. For each city-resolution pair, the $750$ patches are split into $450$ training, $150$ validation, and $150$ test images, fixed at the benchmark level and released with the dataset. To support resolution-paired analyses, we additionally co-locate $150$ coordinates per city imaged at both resolutions ($300$ paired patches per city). Sampling code, random seeds, and per-patch metadata (city, resolution, longitude/latitude, source tile ID) are released with the benchmark; within-city split distributional diagnostics are in Appendix~\ref{app:splits}.

\subsection{Annotation Pipeline}
\label{sec:annotation}

Pixel-level shadow masks are produced by $7$ trained annotators in a three-phase pipeline that jointly solves throughput and label quality. Phase~1 establishes the label distribution from scratch through human annotation. Phase~2 trains an interim shadow detector on Phase~1 labels and uses its predictions to seed subsequent batches, which annotators edit rather than redraw. Phase~3 is a retroactive quality-control pass that intersects each prior segment with the final-model prediction and returns disagreements for adjudication. Across all phases, every $100$-image batch includes $5$ gold images with frozen reference masks; Cohen's $\kappa$ on these images monitors annotator drift across batches. Each shadow segment carries one of six type labels (taxonomy in Appendix~\ref{app:data_stats}); the six-way typing is released for downstream analyses, but the LOCO benchmark in Section~\ref{sec:loco} evaluates binary masks (shadow vs.\ background). Tooling, per-phase throughput, inter-annotator agreement statistics, and Phase~3 adjudication rules are in Appendix~\ref{app:annotation}.

\begin{table}[t]
\centering
\small
\setlength{\tabcolsep}{2.4pt}
\caption{
\textbf{LOCO baseline performance across architectures and adaptation methods.} Two-class boundary-tolerant mIoU ($\pm 2$\,px). For each architecture-method pair we report mIoU pooled across the $6$ test cells per pair ($3$ held-out cities $\times$ $2$ native resolutions, $150$ paired test images per cell). \textit{Upper} is the within-city upper bound (per-architecture, pooled across the same cells). $\Delta$ is the mean gap to the upper bound (mIoU points; negative is degradation). \textit{Sig.} is the count of cells, out of $6$, with significant degradation under paired bootstrap ($p < 0.05$, $B = 10{,}000$). $^\dagger$High-resolution cells only ($3$ held-out cities); Sig.\ is given out of the evaluated cells. The \textit{Pooled} block pools MAMNet, OGLANet and DINOv3.
}
\label{tab:loco-main}
\begin{tabular}{l|cccc|cccc|cccc|cc}
\toprule
& \multicolumn{4}{c|}{\textit{LOCO mIoU}}
& \multicolumn{4}{c|}{$\Delta$ \textit{(mIoU)}}
& \multicolumn{4}{c|}{\textit{Sig.}}
& \multicolumn{2}{c}{\textit{Pooled}} \\
\textbf{Method} & \textbf{D} & \textbf{M} & \textbf{O} & \textbf{C}$^\dagger$
                & \textbf{D} & \textbf{M} & \textbf{O} & \textbf{C}$^\dagger$
                & \textbf{D} & \textbf{M} & \textbf{O} & \textbf{C}$^\dagger$
                & $\Delta$ & \textit{Sig./18} \\
\midrule
\textit{Upper} & $\mathit{77.6}$ & $\mathit{81.7}$ & $\mathit{74.9}$ & $\mathit{78.8}$ & --- & --- & --- & --- & --- & --- & --- & --- & --- & --- \\
\midrule
Vanilla   & 70.0 & 77.1 & 67.0 & 75.3 & $-7.6$ & $-4.6$ & $-7.9$ & $-3.5$ & 6 & 6 & 6 & 3/3 & $-6.68$ & 18 \\
FDA       & 69.4 & 77.0 & 67.0 & 75.1 & $-8.1$ & $-4.7$ & $-8.0$ & $-3.7$ & 6 & 6 & 6 & 3/3 & $-6.93$ & 18 \\
SegDeSiC  & 73.0 & 76.1 & 65.0 & -- & $-4.6$ & $-5.6$ & $-9.9$ & -- & 5 & 6 & 6 & -- & $-6.68$ & 17 \\
IIM       & 71.4 & 71.9 & 65.5 & -- & $-6.1$ & $-9.7$ & $-9.4$ & -- & 6 & 6 & 6 & -- & $-8.41$ & 18 \\
MRFP+     & 72.7 & 71.4 & 60.9 & -- & $-4.9$ & $-10.3$ & $-14.0$ & -- & 6 & 6 & 6 & -- & $-9.74$ & 18 \\
FADA      & 67.0 & 72.3 & 63.8 & -- & $-10.6$ & $-9.4$ & $-11.1$ & -- & 6 & 6 & 6 & -- & $-10.36$ & 18 \\
ISW       & 59.0 & 68.3 & 63.9 & -- & $-18.5$ & $-13.4$ & $-11.1$ & -- & 6 & 6 & 6 & -- & $-14.32$ & 18 \\
SAR$^\dagger$ & 68.4 & 73.1 & 65.4 & 75.3 & $-6.3$ & $-6.6$ & $-7.6$ & $-3.5$ & 3/3 & 3/3 & 3/3 & 3/3 & $-6.86$ & 9/9 \\
BSM$^\dagger$ & 66.2 & 72.8 & 62.3 & 75.7 & $-5.7$ & $-6.9$ & $-10.7$ & $-3.0$ & 2/2 & 3/3 & 3/3 & 3/3 & $-8.05$ & 8/8 \\
\midrule
\textit{Avg.\ $\Delta$} & --- & --- & --- & --- & $-8.6$ & $-8.2$ & $-10.2$ & --- & --- & --- & --- & --- & --- & --- \\
\textit{\# Sig./$42$} & --- & --- & --- & --- & --- & --- & --- & --- & 41 & 42 & 42 & --- & --- & --- \\
\bottomrule
\end{tabular}
\\[2pt]
\footnotesize
\textbf{D}: DINOv3 \quad \textbf{M}: MAMNet \quad \textbf{O}: OGLANet \quad \textbf{C}: CrossEarth. Avg.\ $\Delta$ and \# Sig.\ cover the first seven methods. --: not run; FADA, IIM, MRFP+, SegDeSiC and ISW are defined on a per-pixel head, which CrossEarth's query-based Mask2Former decoder lacks. CrossEarth \sib{} ($-4.4$, 3/3) is in Table~\ref{tab:loco-full-hr}.
\end{table}

\subsection{The LOCO Protocol}
\label{sec:loco}

We define the leave-one-city-out (LOCO) protocol over the city set $\mathcal{C} = \{\text{Chicago, Miami, Phoenix}\}$. For each holdout city $c^* \in \mathcal{C}$, the training set is sampled from the two non-holdout cities at a fixed $50/50$ ratio, yielding $n_{\text{train}} = 450$ images ($225$ per training city). The validation set is $n_{\text{val}} = 300$ images drawn from the same two training cities ($150$ per city), and the test set is the held-out city's $150$-image test pool. The within-city upper bound for each (city, resolution) is a model trained on that pair's $450$ training images and evaluated on its $150$ test images, so the LOCO-vs-upper-bound gap isolates geographic shift from training-set scale. All three holdout rotations are evaluated; results aggregate over rotations and the two native resolutions, giving $6$ test cells per (architecture, method) pair. The primary metric is boundary-tolerant mIoU, the mean of shadow and background IoU with a $\pm 2$-pixel don't-care band, since shadow boundaries are inherently imprecise to annotate at sub-pixel resolution; full metric definition and excluded-mask handling are in Appendix~\ref{app:metric}. Per-cell significance uses paired bootstrap on per-image $\Delta$ values ($B = 10{,}000$ resamples, two-tailed $p$-value against the within-city upper bound).

\subsection{Baselines}
\label{sec:baselines}

We evaluate four architectures spanning representative design families: \textbf{MAMNet}~\cite{mamnet} (CNN with criss-cross attention), \textbf{OGLANet}~\cite{oglanet} (CNN with multi-scale prediction-stage output aggregation), \textbf{DINOv3}~\cite{dinov3} (ViT with self-supervised pretraining, fine-tuned end-to-end), and \textbf{CrossEarth}~\cite{gong2025crossearth} (remote-sensing foundation model with a frozen DINOv2 backbone and a Mask2Former decoder). MAMNet, OGLANet and DINOv3 are each paired with nine configurations: \textbf{Vanilla} (source-only, no domain adaptation), \textbf{FDA}~\cite{fda}, \textbf{SegDeSiC}~\cite{segdesicnet}, \textbf{IIM}~\cite{yola}, \textbf{ISW}~\cite{isw}, \textbf{MRFP+}~\cite{mrfp}, \textbf{FADA}~\cite{fada}, \textbf{SAR}~\cite{niu2023sar}, and \textbf{BSM}~\cite{luo2023bsm}. Adaptation methods leave the architecture unchanged. The first seven are tested on all $7 \times 3 \times 3 \times 2 = 126$ (method, architecture, holdout city, resolution) configurations, and SAR, BSM and CrossEarth on the high-resolution cells, for $158$ LOCO test cells of $150$ images each. Architecture details, pretrained weight sources, and adaptation-method hyperparameters are in Appendix~\ref{app:training}. Table~\ref{tab:loco-main} summarizes results pooled across the three holdouts and two resolutions; per-cell numbers and confidence intervals are in Appendix~\ref{app:loco_extended}. $157$ of the $158$ test cells degrade significantly under geographic transfer ($p < 0.05$, paired bootstrap; $151$ after Bonferroni correction over the $158$-test family), with mean LOCO degradation of $-8.34$ mIoU and a worst single cell of $-28.50$ mIoU. A data-scaling test indicates that the gap is not a small-data effect: quadrupling the LOCO training set from $225$ to $900$ images narrows it by $0.80$ mIoU ($95\%$ CI $[0.46, 1.14]$), from $6.48$ to $5.50$, with the upper bound significantly ahead in every cell but one (high resolution, three seeds; Appendix~\ref{app:ladder}). Two findings in Table~\ref{tab:loco-main} shape the rest of the paper. First, the cost of an adaptation method depends on the architecture: ISW costs DINOv3 $-18.5$ mIoU against $-11.1$ on OGLANet, while MRFP$^+$ costs OGLANet $-14.0$ against $-4.9$ on DINOv3. Second, no adaptation method is uniformly safe: six of eight adaptation methods degrade more than Vanilla on average (five significantly, Holm $p < 0.05$), and every method except SegDeSiC degrades every cell it is evaluated on, on all four architectures. Without access to the target city, Table~\ref{tab:loco-main} gives a practitioner no basis for choosing an adaptation method. Section~\ref{sec:evaluation} asks \textit{where} in the architecture the failure lives and \textit{why} uniform corrections fail non-uniformly across them.

\section{Evaluating Geographic Transfer Failure}
\label{sec:evaluation}

\subsection{Framework}
\label{sec:framework}

\begin{figure*}[t]
  \centering
  \includegraphics[height=4cm, width=\linewidth]{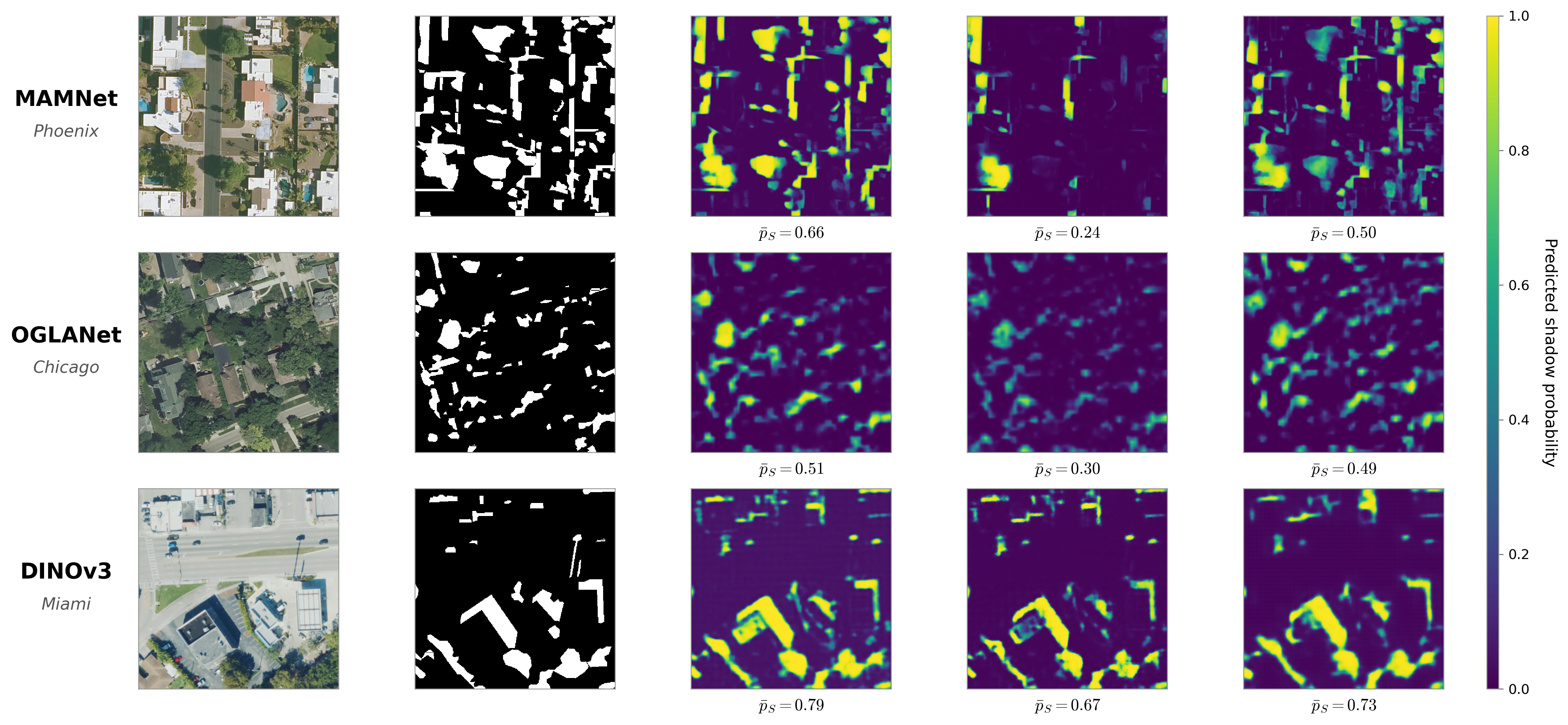}
  \caption{\textbf{Failure mode and recovery under geographic transfer.}
Rows: three architectures on tiles from their leave-one-city-out (LOCO) rotations, together spanning all three held-out cities. Columns: input, ground-truth mask (white $=$ shadow), and predicted shadow probability under Upper bound (within-city training), LOCO, and SIB (\S\ref{sec:sib}); shared colormap $[0, 1]$. $\bar{p}_S$: mean predicted shadow probability inside ground-truth shadow. \emph{What to look for:} LOCO collapses $\bar{p}_S$ while background stays dark — failure is washed-out predictions over true shadow regions, not hallucinated shadow elsewhere. SIB recovers foreground confidence on the same regions across all three architectures with varying magnitude.}
  \label{fig:qualitative}
\end{figure*}

\begin{figure*}[t]
  \centering
  \includegraphics[width=\linewidth]{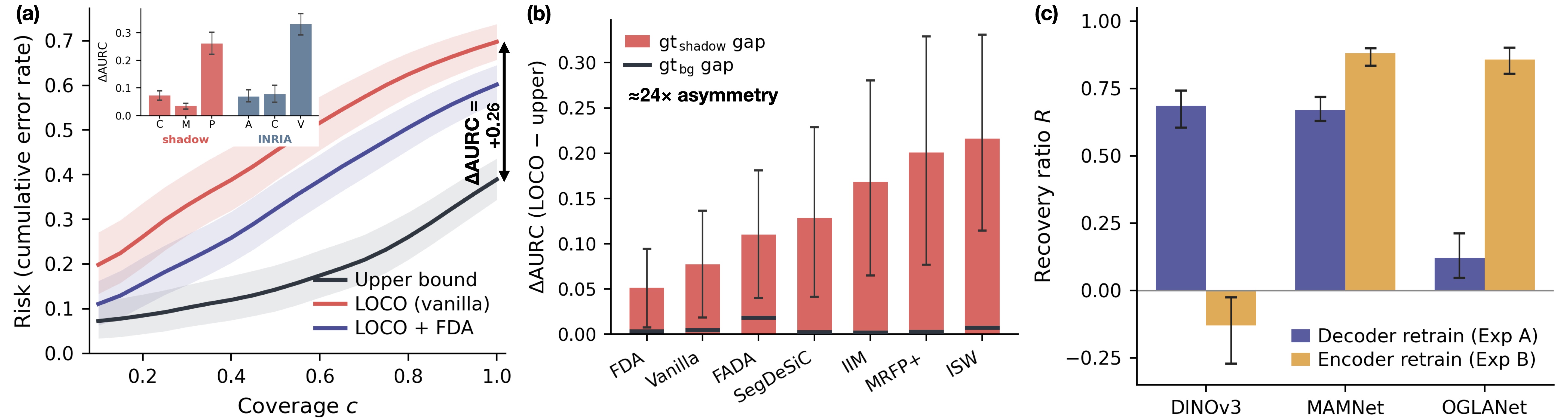}
  \caption{\textbf{Diagnostic framework.}
\textbf{(a)} Risk-coverage curves on foreground pixels (MAMNet, Phoenix held out): LOCO accumulates $\Delta$AURC$=+0.26$ over the upper bound; FDA narrows it. Bands: 95\% bootstrap CIs. \emph{Inset:} foreground gap replicates on INRIA building footprints under matched LOCO (C/M/P: Chicago/Miami/Phoenix; A/C/V: Austin/Chicago/Vienna).
\textbf{(b)} Per-method AURC gap across $9$ cells ($3$ architectures $\times$ $3$ cities): the gap is class-asymmetric, concentrated on foreground (coral) with a pooled background gap (black) $24\times$ smaller over the $63$ cells of the $7$ methods.
\textbf{(c)} Per-image recovery ratio $R$ under decoder (Exp~A) vs.\ encoder (Exp~B) retraining: the DINOv3$\leftrightarrow$OGLANet sign reversal shows the failure locus is architecture-dependent.
LOCO: leave-one-city-out; AURC: area under risk-coverage.}
  \label{fig:diagnostics}
\end{figure*}

Table~\ref{tab:loco-main} showed that no adaptation method is uniformly safe across architectures, suggesting the failure is not a single mechanism a single correction can address. Yet the gap is consistent enough within each architecture that we can ask \emph{where} in the model it lives. We answer this with three controlled interventions on the LOCO models from Table~\ref{tab:loco-main}, ordered by what a practitioner would try first. Section~\ref{sec:feature-space} tests the cheapest possible fix: aligning pixel statistics across cities, ruling in or out a whole class of preprocessing-based recoveries. Section~\ref{sec:sp-gap} then asks whether the model's outputs are still usable under a confidence threshold, since deployed systems consume thresholded predictions rather than raw probabilities. Section~\ref{sec:architecture-locus} localizes \emph{which component} of each architecture carries the failure. Figure~\ref{fig:qualitative} grounds what we are explaining: under LOCO, models keep spatial localization but lose foreground-class confidence: sharp upper-bound predictions become washed-out heatmaps over true shadow regions, not hallucinated shadow elsewhere.

\subsection{Failure operates in feature space, not pixel space}
\label{sec:feature-space}

\textbf{Test design.} If the LOCO failure were a pixel-statistics shortcut, aligning the test image's pixel histogram to the training-city distribution should at least partially recover the gap. We test this directly. For each (architecture, holdout city) pair under vanilla LOCO, we compute the aggregate intensity histogram (the mean of the three channels) of the source-city training images. We map each test image's intensity CDF onto this histogram, restore each pixel's ratio between its channels and its intensity, and run inference with the trained model unchanged. 

\textbf{Result.} Histogram matching makes things worse, uniformly. Across the $9$ vanilla LOCO cells, mean mIoU drops by $0.45$ relative to LOCO itself (paired bootstrap $95\%$ CI $[-0.523, -0.380]$, $p < 0.001$); not a single cell improves. To put the magnitude on a scale, we use the per-image recovery ratio $R_i = \frac{\text{IoU}_i^{\text{match}} - \text{IoU}_i^{\text{LOCO}}}{\text{IoU}_i^{\text{upper}} - \text{IoU}_i^{\text{LOCO}}}$ where $R = 1$ means full recovery; $R = 0$ means no change from LOCO; $R < 0$ means the intervention is worse than doing nothing. Per-architecture medians fall in $[-2.72, -1.55]$, with all $95\%$ CIs excluding zero: at the median, matched images lie $2.5$ to $3.7$ times as far from the upper bound as LOCO itself. The same rule improves none of $27$ vanilla LOCO models trained on $225$, $450$ or $900$ images (three architectures, high resolution; Appendix~\ref{app:ladder}).

\textbf{Training-time appearance augmentation.} We also retrain both the LOCO model and its upper bound with appearance-only augmentation (colour, illumination, contrast, blur, noise; geometry preserved) on the $9$ high-resolution (architecture, held-out city) cells. Retraining with a different random seed alone moves a cell's gap: across three vanilla seeds, the standard deviation of the gap ranges from $0.24$ to $4.26$ mIoU between cells. Before training the augmented models, we therefore set each cell's threshold at twice this standard deviation and count a gap as changed only if augmentation moves it by more. Six of nine gaps do not change. The mean gap moves from $-6.04$ to $-4.11$ mIoU, with OGLANet/Phoenix and DINOv3/Phoenix narrowing and DINOv3/Miami widening. Phoenix also has the largest seed variance of the three cities, and three cities cannot separate a Phoenix-specific appearance effect from that variance (Appendix~\ref{app:aug-stress}).

\textbf{Implication.} Geographic transfer failure is not a pixel-level shortcut. The features the LOCO model relies on encode the joint distribution of intensity, texture, and spatial pattern in source-city imagery. Aligning the marginal pixel histogram alone breaks this coupling without replacing it with anything useful. This rules out a whole class of cheap fixes a practitioner might try first (like gamma correction, contrast normalization, single-channel histogram standardization) and tells us any successful intervention has to operate at the feature level. 

\subsection{Class-asymmetric selective-prediction failure}
\label{sec:sp-gap}

\textbf{Test design.} A practitioner deploying segmentation under domain shift consumes confidence-thresholded predictions, not raw outputs. The relevant question is therefore not the average error but how the threshold-conditioned error rate degrades under shift. The selective-prediction (SP) framework~\cite{geifman2017selective, traub2024overcoming, cattelan2023fix} measures threshold-conditioned error: per image, sort pixels by confidence and compute the cumulative error rate at each coverage level $c \in \{0.10, 0.15, \ldots, 1.00\}$. The Area Under the Risk-Coverage curve (AURC) integrates this curve. The \emph{SP gap} = AURC$_{\text{LOCO}}$ $-$ AURC$_{\text{upper}}$ measures how much threshold-conditioned error rises under transfer; a gap of zero means selective deployment recovers within-city quality. We compute the gap on three pixel populations: the foreground class (GT-shadow, confidence $= p$), the background class (GT-background, confidence $= 1{-}p$), and all pixels (confidence $= \max(p, 1{-}p)$). All numbers in this subsection come from the $63$ LOCO cells (architecture $\times$ method $\times$ holdout city, for high-resolution); paired image-level bootstrap ($B = 1{,}000$) gives $95\%$ confidence intervals.

\textbf{The gap concentrates on the foreground class.} On shadow pixels the gap averages $+0.136$ across the $63$ cells. On background pixels it averages $+0.006$. The asymmetry is consistent across cells: the foreground gap exceeded the background gap in $55$ of the $63$ cells (sign-test $p = 4.9 \times 10^{-10}$). Two further measurements describe how the model's confidence shifts. At the logit level, true-positive logits drop on average by $0.85$ nats (the model becomes much less confident on its foreground predictions), while logits on false-positive predictions barely move ($-0.15$ nats); the drop is larger on the true-positive side in $55$ of $63$ cells. Calibration error rises asymmetrically in the same direction (Appendix~\ref{app:sp-recovery}). The LOCO failure is therefore not uniform confidence flattening: the model selectively loses commitment on the class it is trying to detect, while leaving its background commitment intact.

\textbf{Cross-task scope test on INRIA building footprints.} A natural objection is that these patterns are properties of shadow appearance, not geographic transfer. We replicate the protocol on INRIA building extraction~\cite{inria}, a dense-prediction task whose foreground is shape-defined rather than illumination-defined (MAMNet, holdout cities Austin/Chicago/Vienna, matched training recipe). The foreground SP gap and the foreground-vs-background asymmetry both transfer in direction across tasks (sign test at the strongest result the sample size permits, $p=0.016$ each; Cohen's $d>1.1$), confirming the SP-gap signature is a property of LOCO across tasks rather than of shadows. Figure~\ref{fig:diagnostics}(a) inset shows the gap distributions side by side.

\textbf{A recoverable threshold exists.} The selective-prediction framing is only useful if there is some confidence threshold at which LOCO predictions match within-city quality. We find one in $59$ of $63$ cells; the four exceptions are all on Phoenix, the most distant holdout, under the most aggressive adaptation methods. This makes coverage thresholding, rather than post-hoc temperature scaling, the natural label-free recovery recipe under geographic shift, and Section~\ref{sec:sib} builds on it directly. Method-ranking stability across architectures, the post-hoc temperature-scaling alternative, and per-cell threshold values are in Appendix~\ref{app:sp-recovery}.

\textbf{Implication.} The SP gap is large, foreground-concentrated, hidden by $0/1$ evaluation, recoverable by coverage thresholding, and consistent across architectures and across a second dense-prediction task. Section~\ref{sec:architecture-locus} asks where in the architecture the failure that produces this gap actually lives.

\subsection{Architecture-dependent failure locus}
\label{sec:architecture-locus}

Sections~\ref{sec:feature-space} and~\ref{sec:sp-gap} established that the failure is in feature space and that confidence collapses asymmetrically. Neither result tells us \emph{which part} of a given model produces the collapse: the encoder, the decoder, or both. This matters because uniform feature-level corrections (ISW whitens, MRFP$^+$ randomizes) act on a fixed locus, and Table~\ref{tab:loco-main} already showed they fail catastrophically on some architectures while helping others. If different architectures carry the failure in different components, that asymmetry is the explanation.

\textbf{Test design.} We localise the failure by targeted retraining. \textbf{Experiment A} freezes the LOCO-trained encoder, reinitializes the decoder, and retrains \emph{only} the decoder on $25\%$ of the holdout-city training set. \textbf{Experiment B} does the inverse: freezes the decoder, resets the encoder to its pretrained initialization, and retrains \emph{only} the encoder on the same subset. If a component carries the failure, retraining it on a small target sample should recover most of the within-city gap; retraining the other component should not. Recovery is measured by the per-image ratio $R$ defined earlier (training details in Appendix~\ref{app:locus-retraining}).

\textbf{Mirror-image recovery pattern.} Decoder retraining (Experiment A) recovers DINOv3 ($R = 0.69$) and MAMNet ($R = 0.67$) but barely moves OGLANet ($R = 0.12$, $[0.05, 0.21]$ - CI excludes meaningful recovery). Encoder retraining (Experiment B) shows the opposite pattern: it makes DINOv3 \textit{worse} than LOCO ($R = -0.13$), while recovering both CNNs ($R = 0.88$ MAMNet, $R = 0.86$ OGLANet). All CIs in Appendix~\ref{app:locus-retraining}. The DINOv3-vs-OGLANet sign reversal across the two experiments shows decoder works for one, encoder works for the other, and each architecture is harmed by the other's fix. This is the central evidence that the failure component is architecture-dependent. Figure~\ref{fig:diagnostics}(c) shows the pattern.

\textbf{Frozen encoder.} If the gap came only from fine-tuning the encoder on two source cities, freezing the encoder would remove it. With the DINOv3 encoder frozen and only the decoder trained, the upper bound still exceeds LOCO by $7.10$, $2.67$ and $2.75$ mIoU on Chicago, Miami and Phoenix (each significant after Holm correction over the three cities; Appendix~\ref{app:frozen}). Both models share the frozen encoder, so this gap arises in the decoder, consistent with the locus Experiment~A measures for DINOv3.

\textbf{Pretraining and fine-tuning regime.} DINOv3 also differs from the CNNs in self-supervised pretraining and a low fine-tuning learning rate. We test both on MAMNet with a ResNet-50 encoder by training four versions: ImageNet-supervised or self-supervised DINO pretraining, each fine-tuned at learning rate $10^{-4}$ or $10^{-5}$ (three held-out cities, three seeds each). Encoder retraining recovers all four versions (median $R$ $0.80$ to $0.87$, $95\%$ CIs within $[0.76, 0.90]$), against $-0.13$ for DINOv3. The self-supervised, low-learning-rate version recovers less than the supervised, high-learning-rate one (mean per-image difference $-0.13$, $95\%$ CI $[-0.22, -0.05]$, Holm $p = 0.005$), and its recoverable component stays in the encoder. Pretraining scale remains confounded: DINO-ResNet-50 is pretrained on ImageNet-1k ($1.28$M images) and DINOv3 on LVD-1689M (Appendix~\ref{app:armL-expA}).

\textbf{Implication.} The recoverable component is architecture-dependent: decoder for DINOv3, encoder for OGLANet, both for MAMNet. This directly explains the catastrophic failures in Table~\ref{tab:loco-main}. MRFP$^+$ perturbs encoder features uniformly without targeting OGLANet's encoder-locus failure, so it costs OGLANet $-14.0$ mIoU. ISW whitens features in a way that corrupts the representation DINOv3 depends on, so it costs DINOv3 $-18.5$ mIoU.

\subsection{External benchmark: S-EO}
\label{sec:seo}

We repeat the benchmark and the three probes on S-EO~\cite{seo}: WorldView-3 imagery of Omaha, Jacksonville and San Diego, four architectures, three seeds, and $3{,}500$ training patches per side (Appendix~\ref{app:seo}). Its masks are derived from digital surface models; both sides of every gap share them, so label noise largely cancels in the gap. The upper bound exceeds LOCO in all $12$ (architecture, held-out city) pairs (Holm $p < 0.05$), by $2.10$ to $7.95$ mIoU and $5.41$ $[5.07, 5.75]$ pooled. Histogram matching lowers mIoU in all $36$ vanilla cells, each significant after Holm correction, and the foreground AURC gap exceeds the background gap in all $12$ pairs, significantly in $8$. Decoder retraining raises foreground IoU in all $18$ MAMNet and DINOv3 cells and lowers it in all $9$ OGLANet cells; the size of this change tracks the gap of the LOCO model whose decoder is retrained ($R^2 = 0.62$), while its sign separates by architecture even between DINOv3 and OGLANet models with similar gaps. The background probability also shifts under LOCO in $10$ of $12$ pairs, so on S-EO the confidence change is not confined to the shadow class. The gap shows no dependence on the sun-elevation distance between test patches and training scenes (cluster-robust $p = 0.30$, three cities).

\section{Case Study: Spectral Information Bottleneck (SIB)}
\label{sec:sib}

Section~\ref{sec:evaluation} characterized three failure modes under geographic transfer. We now ask whether this characterization translates into design guidance. We present \sib, a plug-in module that maps each finding to a specific design choice. \sib{} has the smallest mean gap to the within-city upper bound of the ten methods we evaluate, level with FDA. \sib{} is designed from the same LOCO cells on which we evaluate it, so its results show that the diagnosis yields a usable design and do not test the diagnosis independently.

\subsection{Module design}
\label{sec:sib-design}

\begin{figure*}[t]
  \centering
  \includegraphics[width=\linewidth]{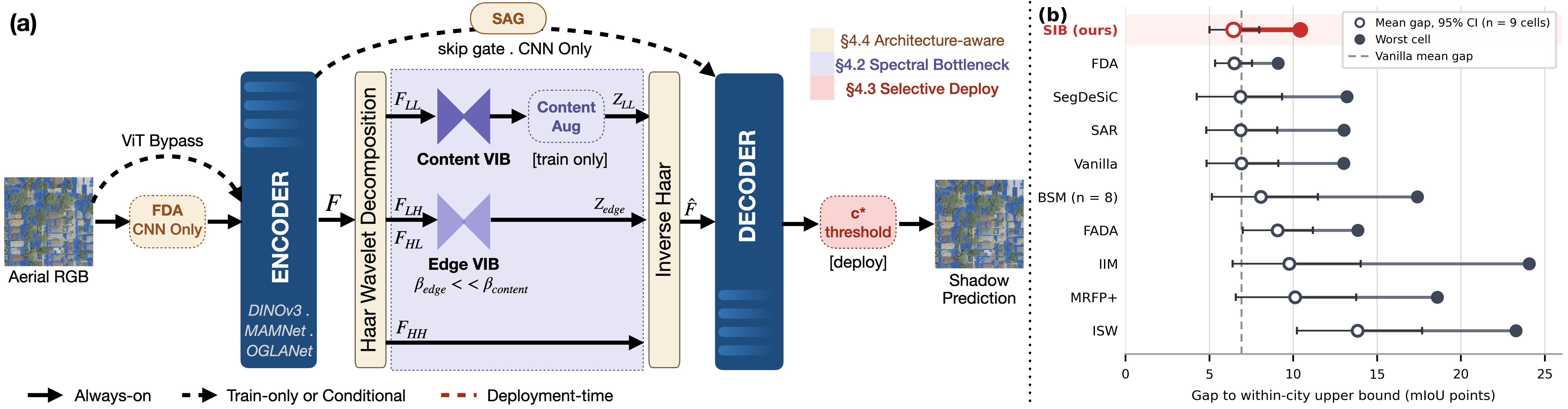}
  \caption{
\textbf{SIB module and gap to the upper bound.}
\textbf{(a)}~Spectral Information Bottleneck (SIB) plug-in module. Three color-tinted zones group components by the diagnostic finding they target: architecture-aware preprocessing (FDA on CNN inputs, SAG on U-Net skips, both bypassed for ViT); spectral bottleneck on the encoder output (Haar split of features $F$ into content $F_{LL}$ and boundary subbands, content compressed strongly, boundaries gently, $F_{HH}$ passed through); deployment-time coverage threshold $c^*$. Component-to-finding mapping in \S\ref{sec:sib-design}.
\textbf{(b)}~Per-method gap to the within-city upper bound across $9$ leave-one-city-out (LOCO) cells. Open markers: mean gap with $95\%$ CI; filled: worst cell; dashed grey line: Vanilla mean gap. Values in Table~\ref{tab:method-profile}.
FDA: Fourier Domain Adaptation; SAG: Skip Attention Gates.
}
  \label{fig:sib}
\end{figure*}

\textbf{Feature-space failure (\S\ref{sec:feature-space}) $\rightarrow$ band-asymmetric compression.} Based on ~\ref{sec:feature-space}, the intervention must operate on feature space. We need to compress domain-specific content while preserving boundary detail. Haar wavelet decomposition gives this separation directly: the low-frequency subband $F_{LL}$ carries surface appearance and city-specific texture; the high-frequency subbands $F_{LH}, F_{HL}, F_{HH}$ carry shadow geometry. Recent VFM-based domain generalization work confirms this content/style separation in feature space~\cite{fada}. We apply a variational information bottleneck~\cite{alemi2017deep} to $F_{LL}$ aggressively and to $F_{LH}, F_{HL}$ gently. $F_{HH}$ passes through. Aggressive compression on $F_{LL}$ removes city-specific content, while gentle compression on the boundary subbands prevents the bottleneck from erasing the edge structure shadow detection requires.

\textbf{Architecture-dependent locus (\S\ref{sec:architecture-locus}) $\rightarrow$ skip gates and CNN-only preprocessing.} The bottleneck acts on encoder output. In U-Net, skip connections route encoder features directly to the decoder, bypassing the bottleneck. We add Skip Attention Gates: channel-wise sigmoid gates on each skip, identity-initialized so they default to a no-op. The model learns to suppress city-specific channels routed around the bottleneck. CNN encoders also absorb pre-encoder pixel statistics no feature-level intervention can reach. We add Fourier Domain Adaptation~\cite{fda} as input preprocessing on the CNN branch only. ViT encoders carry their recoverable component in the decoder (\S\ref{sec:architecture-locus}) and receive neither component.

\textbf{Class-asymmetric SP failure (\S\ref{sec:sp-gap}) $\rightarrow$ coverage thresholding.} The model's confidence ranking survives geographic shift even when calibration fails. We exploit the surviving ranking with a deployment-time recipe: per cell, fit $c^*_{val}$ as the largest $c$ (coverage) at which selective shadow error on the source-city validation split is at most $(1-\tau) \cdot \text{err}_{val}(c{=}1.0)$, with $\tau = 0.5$ fixed globally, i.e., the criterion is to halve the source-val full-coverage error. Apply $c^*_{val}$ at deployment on the held-out city. The procedure uses no held-out labels. Rank preservation guarantees per-image selective error is bounded by full-coverage error, so the population test measures the magnitude of the recovery rather than its direction. Training loss and full hyperparameters in Appendix~\ref{app:sib-impl}.

\subsection{Results}
\label{sec:sib-results}

\begin{table}[t]
\centering
\small
\begin{minipage}[c]{0.45\textwidth}
\caption{
\textbf{\sib{} component ablations.} Mean $\Delta$ mIoU when each component is removed (or, for the inverse-evidence row, replaced), against each architecture's full-model reference from the same training batch, pooled over the three held-out cities ($n = 450$ paired images). Stars: Holm-adjusted $p$ over the $28$ ablation tests, $^{*}\,p<0.05$, $^{**}\,p<0.01$ (paired image bootstrap, $B = 10{,}000$). $^{\dagger}$MAMNet row removes every VIB. $^{\ddagger}$Run predicts almost no shadow. Full table in Appendix~\ref{app:sib-ablations}.
}
\label{tab:sib-ablations}
\end{minipage}\hfill
\begin{minipage}[c]{0.52\textwidth}
\centering
\setlength{\tabcolsep}{4pt}
\begin{tabular}{l|ccc}
\toprule
\textbf{Variant} & \textbf{MAM} & \textbf{OGL} & \textbf{DIN} \\
\midrule
$-$ SAG                      & $+0.23$ & $-2.06^{**}$ & N/A \\
$-$ FDA, contrast            & $-4.42^{**}$ & $-0.87$ & N/A \\
$-$ Haar                     & $+0.11$ & $-0.63$ & $-23.49^{**\ddagger}$ \\
$-$ VIB on $F_{LL}$          & $+0.29^{\dagger}$ & $-1.01^{**}$ & $-0.71$ \\
VIB on $F_{HL}$ (inv.)       & $-0.29$ & $-1.99^{**}$ & $-0.74$ \\
\bottomrule
\end{tabular}
\end{minipage}
\end{table}

\textbf{Gap to the upper bound.} Over the $9$ cells (3 architectures $\times$ 3 held-out cities; Table~\ref{tab:method-profile}), \sib{} has the smallest mean gap to the within-city upper bound of Vanilla and the eight adaptation methods, $6.42$ mIoU ($95\%$ CI $[5.01, 7.96]$), level with FDA at $6.47$ (difference $+0.05$, $[-0.42, +0.49]$). Its worst-case gap, $10.41$ mIoU (OGLANet, Miami), is second to FDA's $9.10$ and cannot be separated from it (difference $+1.31$, $[-1.01, +3.65]$). FDA and SegDeSiC use unlabeled target-city images in every cell, as \sib{} does on its CNN backbones through FDA. \sib{} leaves a significant gap in all $9$ cells, and on S-EO its worst cell scores $1.57$ mIoU below Vanilla's (Appendix~\ref{app:seo}). The deployment-time coverage recipe reduces selective shadow error on all $9$ cells (mean $\Delta = -0.199$, paired Wilcoxon $p = 0.002$), stable across $\tau \in \{0.3, 0.5, 0.7\}$ (Appendix~\ref{app:tau-sweep}).

\textbf{Component ablations.} Table~\ref{tab:sib-ablations} reports the change in mIoU when each \sib{} component is removed, against each architecture's full-model reference from the same training batch, with Holm correction over the $28$ ablation tests. The largest significant changes are on OGLANet, where removing SAG costs $2.06$ mIoU, compressing $F_{HL}$ in place of $F_{LL}$ costs $1.99$, and removing the content VIB costs $1.01$. On MAMNet the only significant row removes FDA preprocessing together with the contrast stage ($-4.42$ mIoU), and removing SAG changes mIoU by $+0.23$ (not significant). On DINOv3, removing the Haar decomposition gives $-23.49$ mIoU, but that run predicts no shadow on nearly every Chicago and Miami image, so the row records a training collapse and does not isolate the decomposition's contribution; the content-VIB and inverse-subband rows on DINOv3 are not significant after correction. Appendix~\ref{app:sib-ablations} lists every row together with the training differences between each ablation and its reference.

\section{Discussion and Conclusion}
\label{sec:conclusion}

We introduced \shadow, a $4{,}500$-image human-verified benchmark for geographic transfer in overhead shadow detection across three U.S.\ cities and two native resolutions, with a leave-one-city-out protocol and paired bootstrap reporting. Its three-probe framework indicates that the failure operates in feature space (Section~\ref{sec:feature-space}), appears as a class-asymmetric collapse of foreground confidence that $0/1$ metrics under-report by $2.3\times$ (Section~\ref{sec:sp-gap}), and is recoverable through an architecture-dependent component: decoder for DINOv3, encoder for OGLANet, both for MAMNet (Section~\ref{sec:architecture-locus}). Four of five selective-prediction findings replicate on INRIA building footprints. The gap remains significant in every $900$-image cell and on the external S-EO benchmark. As a case study, \sib{} has the smallest mean gap of the ten methods we evaluate ($6.42$ mIoU), level with FDA.

The benchmark covers three U.S.\ cities, RGB-only NAIP imagery, a single fall season, and four architectures. Non-grid morphologies, multispectral sensors, and untested architectures (e.g., SegFormer) remain open. We characterize where the failure occurs, how it distorts predictions, and which component recovers it, but do not isolate a causal mechanism that connects these characterizations. DINOv3's decoder locus also cannot be attributed to its ViT architecture alone, since no CNN in our tests matches its pretraining scale. We release \shadow, the LOCO splits, training code, and trained models as an evaluation instrument to be read with the three probes: its LOCO gaps measure geographic transfer risk in overhead dense prediction, and its two native resolutions support diagnosis of cross-resolution transfer failure. Together, the benchmark and framework give the field a shared instrument for the question that matters at deployment: not just whether a model works, but where and on which class it will fail when the city changes.

\begin{ack}
This work received no external funding. This work used Delta at the National Center for Supercomputing Applications (NCSA) through allocation CIS260282 from the Advanced Cyberinfrastructure Coordination Ecosystem: Services \& Support (ACCESS) program~\cite{boerner2023access}, which is supported by U.S.\ National Science Foundation grants \#2138259, \#2138286, \#2138307, \#2137603, and \#2138296. This research used the Delta advanced computing and data resource, which is supported by the National Science Foundation (award OAC 2005572) and the State of Illinois. Delta is a joint effort of the University of Illinois Urbana-Champaign and its National Center for Supercomputing Applications. The authors declare no competing interests.
\end{ack}

\clearpage
\bibliographystyle{unsrt}
\bibliography{ref}  %

\begin{thebibliography}{10}

\bibitem{batchu2024satellite}
Vishal Batchu, Alex Wilson, Betty Peng, Carl Elkin, Umangi Jain, Christopher
  Van~Arsdale, Ross Goroshin, and Varun Gulshan.
\newblock Satellite sunroof: High-res digital surface models and roof
  segmentation for global solar mapping.
\newblock In {\em NeurIPS 2024 Workshop on Tackling Climate Change with Machine
  Learning}, 2024.

\bibitem{dashadebench}
Longchao Da, Xiangrui Liu, Mithun Shivakoti, T~Pranav Kutralingam, Yezhou Yang,
  and Hua Wei.
\newblock Shadebench: A benchmark dataset and method for shade simulation in
  sustainable society.
\newblock In {\em NeurIPS 2025 UrbanAI: Harnessing Artificial Intelligence for
  Smart Cities}, 2025.

\bibitem{ukkusuri2024heatwaves}
S.~V. Ukkusuri, S.~U. Park, S.~Mittal, L.~Chapman, G.~Manoli, A.~Santos, et~al.
\newblock We need to prepare our transport systems for heatwaves---here's how.
\newblock {\em Nature}, 632(8024):253--256, 2024.

\bibitem{seo}
El{\'\i}as Masquil, Roger Mar{\'\i}, Thibaud Ehret, Enric Meinhardt-Llopis,
  Pablo Mus{\'e}, and Gabriele Facciolo.
\newblock S-eo: A large-scale dataset for geometry-aware shadow detection in
  remote sensing applications.
\newblock In {\em Proceedings of the Computer Vision and Pattern Recognition,
  EarthVision: Large Scale Computer Vision for Remote Sensing Imagery}, pages
  2383--2393, 2025.

\bibitem{mittal2025overcoming}
S.~Mittal and S.~V. Ukkusuri.
\newblock Overcoming data scarcity in transit planning: A novel framework
  combining machine learning and metaheuristics.
\newblock {\em Data Science for Transportation}, 7(1):4, 2025.

\bibitem{verma2025spatial}
R.~Verma, S.~Mittal, and S.~V. Ukkusuri.
\newblock Spatial {Access} of {America}: Multiple indicators of accessibility
  to opportunities.
\newblock {\em Scientific Data}, 12(1):1223, 2025.

\bibitem{ogier2022flamby}
Jean Ogier~du Terrail, Samy-Safwan Ayed, Edwige Cyffers, Felix Grimberg,
  Chaoyang He, Regis Loeb, Paul Mangold, Tanguy Marchand, Othmane Marfoq, Erum
  Mushtaq, et~al.
\newblock Flamby: Datasets and benchmarks for cross-silo federated learning in
  realistic healthcare settings.
\newblock {\em Advances in Neural Information Processing Systems},
  35:5315--5334, 2022.

\bibitem{wang2021loveda}
Junjue Wang, Zhuo Zheng, Ailong Ma, Xiaoyan Lu, and Yanfei Zhong.
\newblock Loveda: A remote sensing land-cover dataset for domain adaptive
  semantic segmentation.
\newblock {\em arXiv preprint arXiv:2110.08733}, 2021.

\bibitem{gong2025crossearth}
Ziyang Gong, Zhixiang Wei, Di~Wang, Xiaoxing Hu, Xianzheng Ma, Hongruixuan
  Chen, Yuru Jia, Yupeng Deng, Zhenming Ji, Xiangwei Zhu, et~al.
\newblock Crossearth: Geospatial vision foundation model for domain
  generalizable remote sensing semantic segmentation.
\newblock {\em IEEE Transactions on Pattern Analysis and Machine Intelligence},
  2025.

\bibitem{aisd}
Shuang Luo, Huifang Li, and Huanfeng Shen.
\newblock Deeply supervised convolutional neural network for shadow detection
  based on a novel aerial shadow imagery dataset.
\newblock {\em ISPRS Journal of Photogrammetry and Remote Sensing},
  167:443--457, 2020.

\bibitem{sbu}
Tom{\'a}s F~Yago Vicente, Le~Hou, Chen-Ping Yu, Minh Hoai, and Dimitris
  Samaras.
\newblock Large-scale training of shadow detectors with noisily-annotated
  shadow examples.
\newblock In {\em European conference on computer vision}, pages 816--832.
  Springer, 2016.

\bibitem{istd}
Jifeng Wang, Xiang Li, and Jian Yang.
\newblock Stacked conditional generative adversarial networks for jointly
  learning shadow detection and shadow removal.
\newblock In {\em Proceedings of the IEEE conference on computer vision and
  pattern recognition}, pages 1788--1797, 2018.

\bibitem{azimi2023airborne}
Seyed~Majid Azimi and Reza Bahmanyar.
\newblock Airborne-shadow: towards fine-grained shadow detection in aerial
  imagery.
\newblock In {\em DAGM German conference on pattern recognition}, pages 34--49.
  Springer, 2023.

\bibitem{mamnet}
Lei Zhang, Qing Zhang, Yu~Wu, Yanfeng Zhang, Shan Xiang, Donghai Xie, and Zeyu
  Wang.
\newblock A full-scale shadow detection network based on multiple attention
  mechanisms for remote-sensing images.
\newblock {\em Remote Sensing}, 16(24):4789, 2024.

\bibitem{mbcsanet}
Xueli Chang, Haiyang Shi, Tiejun Zhang, Huazhong Jin, and Ao~Xu.
\newblock Shadow detection in remote sensing images based on multibranch
  feature aggregation and channel-spatial attention.
\newblock {\em IEEE Journal of Selected Topics in Applied Earth Observations
  and Remote Sensing}, 18:2618--2630, 2024.

\bibitem{csdformer}
Jiayi Li and Qunming Wang.
\newblock Csdformer: A cloud and shadow detection method for landsat images
  based on transformer.
\newblock {\em International Journal of Applied Earth Observation and
  Geoinformation}, 129:103799, 2024.

\bibitem{oglanet}
Yakun Xie, Dejun Feng, Hongyu Chen, Ziyang Liao, Jun Zhu, Chuangnong Li, and
  Sung~Wook Baik.
\newblock An omni-scale global--local aware network for shadow extraction in
  remote sensing imagery.
\newblock {\em ISPRS Journal of Photogrammetry and Remote Sensing}, 193:29--44,
  2022.

\bibitem{cuhkshadow}
Xiaowei Hu, Tianyu Wang, Chi-Wing Fu, Yitong Jiang, Qiong Wang, and Pheng-Ann
  Heng.
\newblock Revisiting shadow detection: A new benchmark dataset for complex
  world.
\newblock {\em IEEE Transactions on Image Processing}, 30:1925--1934, 2021.

\bibitem{hu2026unveiling}
Xiaowei Hu, Zhenghao Xing, Tianyu Wang, Chi-Wing Fu, and Pheng-Ann Heng.
\newblock Unveiling deep shadows: A survey and benchmark on image and video
  shadow detection, removal, and generation in the deep learning era.
\newblock {\em International Journal of Computer Vision}, 134(4):158, 2026.

\bibitem{fda}
Yanchao Yang and Stefano Soatto.
\newblock Fda: Fourier domain adaptation for semantic segmentation.
\newblock In {\em Proceedings of the IEEE/CVF conference on computer vision and
  pattern recognition}, pages 4085--4095, 2020.

\bibitem{yola}
Mingbo Hong, Shen Cheng, Haibin Huang, Haoqiang Fan, and Shuaicheng Liu.
\newblock You only look around: learning illumination-invariant feature for
  low-light object detection.
\newblock {\em Advances in neural information processing systems},
  37:87136--87158, 2024.

\bibitem{luo2023bsm}
Muying Luo, Shunping Ji, and Shiqing Wei.
\newblock A diverse large-scale building dataset and a novel plug-and-play
  domain generalization method for building extraction.
\newblock {\em IEEE Journal of Selected Topics in Applied Earth Observations
  and Remote Sensing}, 16:4122--4138, 2023.

\bibitem{isw}
Sungha Choi, Sanghun Jung, Huiwon Yun, Joanne~T Kim, Seungryong Kim, and Jaegul
  Choo.
\newblock Robustnet: Improving domain generalization in urban-scene
  segmentation via instance selective whitening.
\newblock In {\em Proceedings of the IEEE/CVF conference on computer vision and
  pattern recognition}, pages 11580--11590, 2021.

\bibitem{segdesicnet}
Sachin Verma, Frank Lindseth, and Gabriel Kiss.
\newblock Segdesicnet: Lightweight semantic segmentation in remote sensing with
  geo-coordinate embeddings for domain adaptation.
\newblock In {\em 2025 IEEE/CVF Winter Conference on Applications of Computer
  Vision (WACV)}, pages 9093--9104. IEEE, 2025.

\bibitem{mrfp}
Sumanth Udupa, Prajwal Gurunath, Aniruddh Sikdar, and Suresh Sundaram.
\newblock Mrfp: Learning generalizable semantic segmentation from sim-2-real
  with multi-resolution feature perturbation.
\newblock In {\em Proceedings of the IEEE/CVF Conference on Computer Vision and
  Pattern Recognition}, pages 5904--5914, 2024.

\bibitem{fada}
Qi~Bi, Jingjun Yi, Hao Zheng, Haolan Zhan, Yawen Huang, Wei Ji, Yuexiang Li,
  and Yefeng Zheng.
\newblock Learning frequency-adapted vision foundation model for domain
  generalized semantic segmentation.
\newblock {\em Advances in Neural Information Processing Systems},
  37:94047--94072, 2024.

\bibitem{wang2021tent}
Dequan Wang, Evan Shelhamer, Shaoteng Liu, Bruno Olshausen, and Trevor Darrell.
\newblock Tent: Fully test-time adaptation by entropy minimization.
\newblock In {\em International Conference on Learning Representations}, 2021.

\bibitem{niu2023sar}
Shuaicheng Niu, Jiaxiang Wu, Yifan Zhang, Zhiquan Wen, Yaofo Chen, Peilin Zhao,
  and Mingkui Tan.
\newblock Towards stable test-time adaptation in dynamic wild world.
\newblock In {\em International Conference on Learning Representations}, 2023.

\bibitem{dinov3}
Oriane Sim{\'e}oni, Huy~V Vo, Maximilian Seitzer, Federico Baldassarre, Maxime
  Oquab, Cijo Jose, Vasil Khalidov, Marc Szafraniec, Seungeun Yi, Micha{\"e}l
  Ramamonjisoa, et~al.
\newblock Dinov3.
\newblock {\em arXiv preprint arXiv:2508.10104}, 2025.

\bibitem{geifman2017selective}
Yonatan Geifman and Ran El-Yaniv.
\newblock Selective classification for deep neural networks.
\newblock {\em Advances in neural information processing systems}, 30, 2017.

\bibitem{traub2024overcoming}
Jeremias Traub, Till~J Bungert, Carsten~T L{\"u}th, Michael Baumgartner,
  Klaus~H Maier-Hein, Lena Maier-Hein, and Paul~F J{\"a}ger.
\newblock Overcoming common flaws in the evaluation of selective classification
  systems.
\newblock {\em Advances in Neural Information Processing Systems},
  37:2323--2347, 2024.

\bibitem{cattelan2023fix}
Lu{\'\i}s Felipe~P Cattelan and Danilo Silva.
\newblock How to fix a broken confidence estimator: Evaluating post-hoc methods
  for selective classification with deep neural networks.
\newblock {\em arXiv preprint arXiv:2305.15508}, 2023.

\bibitem{inria}
Emmanuel Maggiori, Yuliya Tarabalka, Guillaume Charpiat, and Pierre Alliez.
\newblock Can semantic labeling methods generalize to any city? the inria
  aerial image labeling benchmark.
\newblock In {\em 2017 IEEE International geoscience and remote sensing
  symposium (IGARSS)}, pages 3226--3229. IEEE, 2017.

\bibitem{alemi2017deep}
Alexander~A Alemi, Ian Fischer, Joshua~V Dillon, and Kevin Murphy.
\newblock Deep variational information bottleneck.
\newblock In {\em International Conference on Learning Representations}, 2017.

\bibitem{boerner2023access}
Timothy~J. Boerner, Stephen Deems, Thomas~R. Furlani, Shelley~L. Knuth, and
  John Towns.
\newblock {ACCESS}: Advancing innovation: {NSF}'s advanced cyberinfrastructure
  coordination ecosystem: Services \& support.
\newblock In {\em Practice and Experience in Advanced Research Computing (PEARC
  '23)}. ACM, 2023.

\bibitem{gebru2021datasheets}
Timnit Gebru, Jamie Morgenstern, Briana Vecchione, Jennifer~Wortman Vaughan,
  Hanna Wallach, Hal~Daum{\'e} Iii, and Kate Crawford.
\newblock Datasheets for datasets.
\newblock {\em Communications of the ACM}, 64(12):86--92, 2021.

\bibitem{cvat}
{CVAT.ai Corporation}.
\newblock Computer vision annotation tool (cvat).

\bibitem{sam}
Alexander Kirillov, Eric Mintun, Nikhila Ravi, Hanzi Mao, Chloe Rolland, Laura
  Gustafson, Tete Xiao, Spencer Whitehead, Alexander~C Berg, Wan-Yen Lo, et~al.
\newblock Segment anything.
\newblock In {\em Proceedings of the IEEE/CVF international conference on
  computer vision}, pages 4015--4026, 2023.

\bibitem{mittal2026shadowtransferdata}
Shagun Mittal, Raymond~A. Yeh, and Satish Ukkusuri.
\newblock {ShadowTransfer} (revision 5bef90e), 2026.

\bibitem{landis1977measurement}
J~Richard Landis and Gary~G Koch.
\newblock The measurement of observer agreement for categorical data.
\newblock {\em biometrics}, pages 159--174, 1977.

\bibitem{kattenborn2022spatially}
Teja Kattenborn, Felix Schiefer, Julian Frey, Hannes Feilhauer, Miguel~D
  Mahecha, and Carsten~F Dormann.
\newblock Spatially autocorrelated training and validation samples inflate
  performance assessment of convolutional neural networks.
\newblock {\em ISPRS Open Journal of Photogrammetry and Remote Sensing},
  5:100018, 2022.

\bibitem{he2016deep}
Kaiming He, Xiangyu Zhang, Shaoqing Ren, and Jian Sun.
\newblock Deep residual learning for image recognition.
\newblock In {\em Proceedings of the IEEE conference on computer vision and
  pattern recognition}, pages 770--778, 2016.

\bibitem{oquab2024dinov2}
Maxime Oquab, Timoth{\'e}e Darcet, Th{\'e}o Moutakanni, Huy Vo, Marc
  Szafraniec, Vasil Khalidov, Pierre Fernandez, Daniel Haziza, Francisco Massa,
  Alaaeldin El-Nouby, et~al.
\newblock {DINOv2}: Learning robust visual features without supervision.
\newblock {\em Transactions on Machine Learning Research}, 2024.

\bibitem{wei2024rein}
Zhixiang Wei, Lin Chen, Yi~Jin, Xiaoxiao Ma, Tianle Liu, Pengyang Ling, Ben
  Wang, Huaian Chen, and Jinjin Zheng.
\newblock Stronger, fewer, \& superior: Harnessing vision foundation models for
  domain generalized semantic segmentation.
\newblock In {\em Proceedings of the IEEE/CVF Conference on Computer Vision and
  Pattern Recognition}, pages 28619--28630, 2024.

\bibitem{cheng2022mask2former}
Bowen Cheng, Ishan Misra, Alexander~G. Schwing, Alexander Kirillov, and Rohit
  Girdhar.
\newblock Masked-attention mask transformer for universal image segmentation.
\newblock In {\em Proceedings of the IEEE/CVF Conference on Computer Vision and
  Pattern Recognition}, pages 1280--1289, 2022.

\bibitem{huang2017adain}
Xun Huang and Serge Belongie.
\newblock Arbitrary style transfer in real-time with adaptive instance
  normalization.
\newblock In {\em Proceedings of the IEEE International Conference on Computer
  Vision}, pages 1510--1519, 2017.

\bibitem{foret2021sam}
Pierre Foret, Ariel Kleiner, Hossein Mobahi, and Behnam Neyshabur.
\newblock Sharpness-aware minimization for efficiently improving
  generalization.
\newblock In {\em International Conference on Learning Representations}, 2021.

\end{thebibliography}

\clearpage

\appendix

\section{Datasheet}
\label{app:datasheet}

Following \cite{gebru2021datasheets}, this datasheet documents \textsc{ShadowTransfer}.
Items already detailed in the main paper or other appendices are cross-referenced
rather than repeated.

\subsection{Motivation}
\textbf{Purpose.} \textsc{ShadowTransfer} was created to enable measurement of
geographic transfer in overhead shadow detection. No existing overhead shadow
dataset combines human-verified labels, multiple cities, native multi-resolution
coverage, and a defined cross-city evaluation protocol (Table~\ref{tab:dataset_comparison});
the most recent overhead-shadow survey~\citep{hu2026unveiling} identifies
cross-dataset generalization as an open problem.\\
\textbf{Creators and funding.} Shagun Mittal, Raymond A.\ Yeh and Satish Ukkusuri (Purdue University). The work received no external funding; compute was provided through ACCESS allocation CIS260282.

\subsection{Composition}
\textbf{Instances.} 4{,}500 image--mask pairs: 3 cities (Chicago, Miami, Phoenix)
$\times$ 2 native NAIP resolutions (0.3\,m/px, 0.6\,m/px) $\times$ 750 patches.
Each instance is a $384\!\times\!384$ RGB orthoimage patch and a co-registered
binary shadow mask; six-category multiclass masks accompany the patches where
reliable typing was possible (App.~\ref{app:taxonomy}).\\
\textbf{Sample vs.\ full.} A spatial sample of NAIP imagery within each city's
administrative boundary; protocol in App.~\ref{app:sampling}.\\
\textbf{Files per instance.} 8-bit RGB \texttt{.png}; 8-bit binary mask \texttt{.png}
($\{0, 255\}$); optional 8-bit multiclass mask (IDs 0--6); and a per-patch metadata
record (city, native resolution, lat/lon, source NAIP tile, paired-coordinate id,
shadow types present).\\
\textbf{Labels.} Pixel-level binary shadow (primary); six typed shadow categories
(secondary, App.~\ref{app:taxonomy}). Sparse types (vehicle-cast, thin-linear)
are not recommended as primary evaluation targets.\\
\textbf{Relationships between instances.} 300 paired patches per city share a
\texttt{pair\_id} linking the two native resolutions at identical ground coordinates
(App.~\ref{app:sampling}).\\
\textbf{Recommended splits.} Per-city: 450 train / 150 val / 150 test, fixed at the
benchmark level. Cross-city: the LOCO protocol (\S\ref{sec:loco}) defines three
rotations; pre-built folds are released alongside the per-city raw data.\\
\textbf{Errors and noise.} Shadow boundaries are inherently imprecise; we recommend
boundary-tolerant mIoU ($\pm 2$\,px, App.~\ref{app:metric}). A $\approx 3\%$ residual
inter-annotator disagreement persists after three-phase QC (App.~\ref{app:annotation}).\\
\textbf{Self-containment.} Self-contained; underlying NAIP imagery is U.S.\ public
domain (\S\ref{sec:datasheet-distribution}).\\
\textbf{Sensitive content or identifiable individuals.} None. Patches are land-cover
orthoimagery; no PII, no identifiable individuals, no confidential information.

\subsection{Collection Process}
\textbf{Acquisition.} Source NAIP tiles obtained from USDA-FSA APFO; patches cropped
within each city's administrative boundary under one-patch-per-tile sampling to
limit spatial autocorrelation (App.~\ref{app:sampling}).\\
\textbf{Annotation labor.} Seven trained annotators produced masks across 45
sessions in CVAT~\citep{cvat} with SAM-assisted boundary proposals~\citep{sam};
annotators were paid INR~20 per accepted image, and a second annotator who reviewed an image was paid a further INR~20, about \$0.50 per image in total (Appendix~\ref{app:annotator-instructions}).\\
\textbf{Time frame.} A single fall NAIP acquisition window per city; annotation
completed across the 45 sessions.\\
\textbf{IRB / ethics review.} Not required: no human subjects, no PII.

\subsection{Preprocessing, Cleaning, Labeling}
\textbf{Preprocessing.} Patches cropped to $384\!\times\!384$ at native resolution;
no radiometric, color, or geometric transforms applied. Source tile IDs and patch
coordinates are released, so raw inputs can be re-derived from NAIP.\\
\textbf{Labeling.} Three-phase pipeline (App.~\ref{app:annotation}): SAM-assisted
manual draw, model-in-the-loop acceleration, and a retroactive QC pass that
intersects each segment with a final-model prediction. Frozen gold masks
(5 per 100-image batch) drive Cohen's $\kappa$ monitoring; phase-3 mean $\kappa = 0.89$.

\subsection{Uses}
\textbf{Primary use.} Benchmarking shadow detection on overhead aerial imagery,
with explicit measurement of geographic transfer (LOCO) and in-domain performance
per city.\\
\textbf{Already used for.} Baseline, diagnostic, and case-study results in this
paper, including cross-task replication on INRIA building footprints
(\S\ref{sec:sp-gap}, App.~\ref{app:inria}).\\
\textbf{Other suggested uses.} Building footprint and fa\c{c}ade extraction
(occlusion priors), shadow-removal research, domain generalization for dense
prediction, and urban-morphology or solar-geometry analysis from the released
coordinates and shadow-type labels.\\
\textbf{Uses not supported.} Generalization beyond grid-pattern North American
urbanism, beyond NAIP RGB sensor characteristics, or across seasons (the dataset
captures a single fall window) is not supported. The 0.3\,m/px and 0.6\,m/px
subsets are separate native acquisitions; users studying resolution transfer
should not synthesize one from the other.

\subsection{Distribution}
\label{sec:datasheet-distribution}
\textbf{Where.} Hugging Face dataset record at
\url{https://huggingface.co/datasets/shagunmittal/ShadowTransfer};
DOI \href{https://doi.org/10.57967/hf/11483}{10.57967/hf/11483}~\cite{mittal2026shadowtransferdata}. Code at \url{https://github.com/shagun8/shadowtransfer}.\\
\textbf{When.} Public release at publication; the version corresponding to paper
results is Hugging Face revision \texttt{5bef90e}, tagged \texttt{v1.0}.\\
\textbf{License.} Annotations and metadata are released under \mbox{CC~BY~4.0}.
NAIP imagery is U.S.\ public domain (USDA-FSA APFO attribution requested for
derived products). Code is released under the MIT license.\\
\textbf{Export or regulatory restrictions.} None known.\\
\textbf{Croissant metadata.} A machine-readable MLCommons~Croissant description,
including the Responsible AI fields, is released with the dataset at
\url{https://huggingface.co/datasets/shagunmittal/ShadowTransfer/resolve/main/croissant.json}
and is consumable by \texttt{mlcroissant}, TFDS, and any Croissant-aware loader.

\subsection{Maintenance}
\textbf{Maintainer.} Shagun Mittal (\texttt{mittal53@purdue.edu}); correspondence is also handled via the
Hugging Face dataset record (issues/discussions) and the GitHub repository.\\
\textbf{Versioning.} Tagged releases on Hugging Face; \texttt{v1.0} freezes the
splits used for paper results. Errata are posted as new tags; older tags remain
accessible.\\
\textbf{Contributions.} External contributions accepted via pull request on the
GitHub repository, tagged separately from \texttt{v1.0}.\\
\textbf{Retention or individual rights.} N/A: no individuals or personal data
are present.

\section{Annotation Pipeline Details}
\label{app:annotation}

\begin{figure}[h!]
  \centering
  \includegraphics[width=\linewidth]{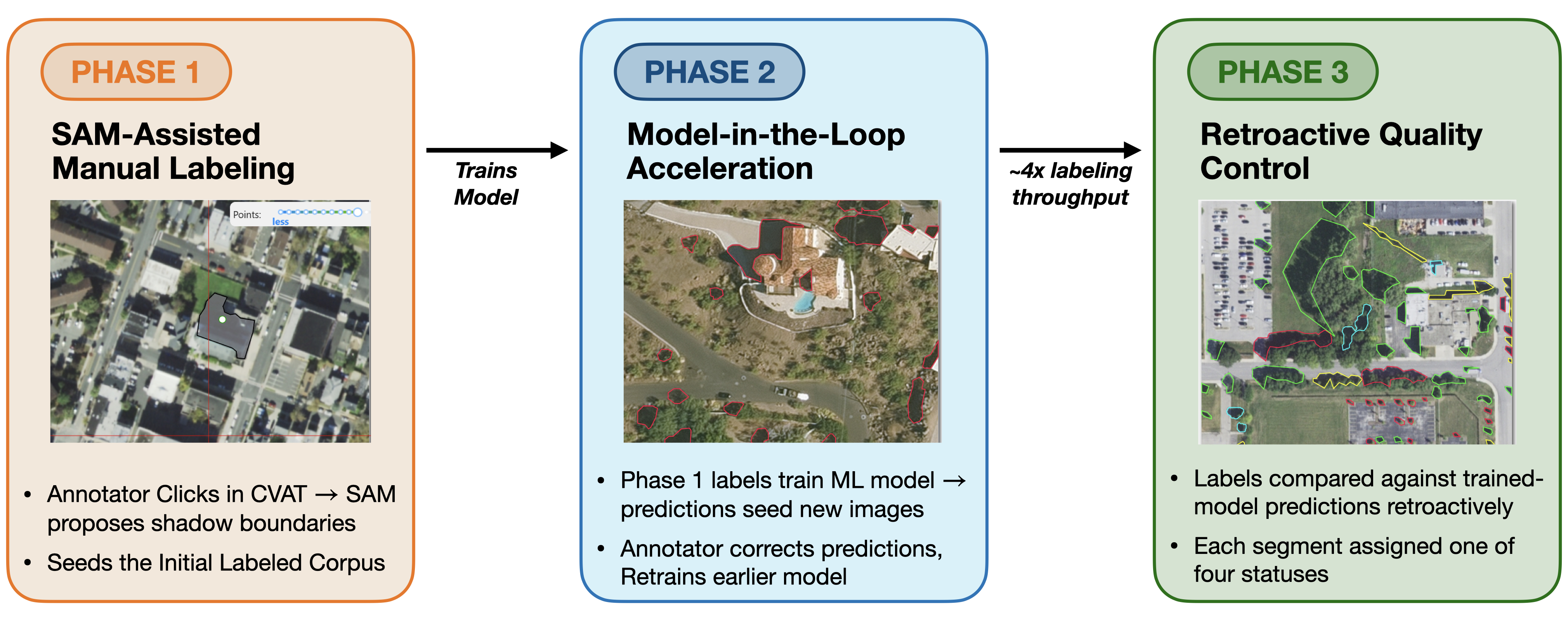}
  \caption{
    \textbf{Three-phase annotation pipeline.}
    Phase 1 (SAM-assisted manual labeling) trains an initial model; Phase 2
    (model-in-the-loop) accelerates throughput $\sim$4$\times$; Phase 3
    (retroactive quality control) reconciles labels against final model predictions.
    Each phase's output seeds the next, jointly solving scale and label quality.
    Inter-annotator agreement statistics and adjudication rules are in
    Appendix~\ref{app:annotation}.
  }
  \label{fig:annotation}
\end{figure}

This appendix documents the three-phase pipeline summarized in
Section~\ref{sec:annotation}: the per-phase tooling
(Figure~\ref{fig:annotation}), the Phase~3 adjudication rules that
reconcile human labels with model predictions
(Table~\ref{tab:adjudication}), and the gold-image protocol and
Cohen's $\kappa$ trajectory that monitor annotator drift across batches
(Table~\ref{tab:kappa}). Together these mechanisms support the
human-verified label claim made against S-EO~\cite{seo} in
Table~\ref{tab:dataset_comparison}.

\subsection{Three-Phase Pipeline}
\label{app:phases}

Seven trained annotators produced shadow masks in CVAT across
$45$ sessions of $105$ images each ($100$ working images $+$ $5$
gold images with frozen reference masks; \S\ref{app:gold}). Sessions
proceed in three sequential phases (Figure~\ref{fig:annotation}); each
phase's output seeds the next.

\textbf{Phase 1: SAM-assisted manual labeling.} Annotators draw shadow
masks on raw $384{\times}384$ patches. The CVAT AI Interactor \cite{cvat}, powered
by the Segment Anything Model (SAM)~\cite{sam}, proposes a
candidate boundary from a single annotator click; the annotator
accepts, refines, or rejects it. Manual polygon drawing is available
for thin-linear and irregular shadows where SAM under-segments. Each
segment receives a shadow-type label from the six-category taxonomy
(\S\ref{app:taxonomy}). Phase~1 spans the first $10$ sessions and
is restricted to Chicago and Miami imagery balanced across the two
native resolutions; Phoenix is admitted only after the label
distribution stabilizes (gold-image $\kappa \geq 0.7$ for a
majority of annotators). Mean throughput is 4.8 minutes per
image after a one-session per-annotator ramp-up.

\textbf{Phase 2: Model-in-the-loop acceleration.} After Phase~1, all
labels train an interim shadow detector. The detector is one of
\{MAMNet, OGLANet, DINOv3\}, drawn at random for each new $100$-image
batch; rotating architectures limits annotator entrainment to any
single model's failure modes. Predictions from the interim model are
loaded into CVAT as the starting layer; annotators delete false
positives, refine boundaries, and add missed segments rather than
redraw from scratch. The interim model is retrained at session
boundaries on the cumulative labeled corpus. Throughput rises to
$75$ seconds per image, a $\sim\!4\times$ acceleration over
Phase~1.

\textbf{Phase 3: Retroactive quality control.} After all working
sessions complete, a final shadow detector is trained on the full
Phase~1$+$2 corpus and used to revisit every annotated segment. For
each image we compute the per-segment Intersection-over-Union (IoU)
between each human annotation $j$ and each model prediction $i$. Each
(annotation, prediction) pair receives one of four statuses
(Table~\ref{tab:adjudication}). Statuses are color-coded in CVAT so
annotators triage at a glance: green segments are skipped; yellow,
blue, and red segments are routed to a second annotator pass for
adjudication. The status taxonomy is released with the dataset for
downstream provenance analyses; for the LOCO benchmark
(\S\ref{sec:loco}), all final segments are merged into the binary
foreground class regardless of their adjudication route.

\begin{table}[h]
  \centering
  \small
  \setlength{\tabcolsep}{4pt}
  \caption{
    \textbf{Phase 3 adjudication rules.}
    Each (annotation, prediction) pair on an image is assigned one of
    four statuses based on their IoU and overlap, and routed to the
    corresponding annotator action.
    $j$: human annotation; $i$: model prediction.
    Distribution column reports the share of all Phase 3 segments
    falling into each status.
  }
  \label{tab:adjudication}
  \begin{tabular}{@{}lp{3.1cm}p{2cm}p{3.4cm}c@{}}
    \toprule
    \textbf{Status (color)} & \textbf{Trigger} & \textbf{Kept segment} & \textbf{Annotator action} & \textbf{Share} \\
    \midrule
    Correct (green) &
      $j \cap i \neq \emptyset$, IoU\,$>0.5$ &
      $j$ &
      Verify, usually skip &
      $65\%$ \\
    Potential\_Annotated (yellow) &
      $j \cap i \neq \emptyset$, IoU\,$\leq 0.5$ &
      $j$ (edit shape) &
      Adjust boundary &
      $21\%$ \\
    Potential\_Model (blue) &
      $i$ has no overlapping $j$ &
      $i$ if validated &
      Confirm or reject; assign type &
      $9\%$ \\
    Incorrect (red) &
      $j$ has no overlapping $i$ &
      $j$ if validated &
      Verify segment is shadow &
      $5\%$ \\
    \bottomrule
  \end{tabular}
\end{table}

\subsection{Gold-Set Protocol and Inter-Annotator Agreement}
\label{app:gold}

\textbf{Gold images.} Every $100$-image working batch is paired with
$5$ gold images that carry frozen reference masks. Gold images are
drawn from previously closed batches that have already cleared
Phase~3, and their reference masks are independently re-adjudicated
by two senior annotators before the gold image enters circulation.
Once frozen, a gold mask is not modified. Gold images are interleaved
with working images and are not visually distinguishable from them in
CVAT, so annotator behavior on gold matches their behavior on the rest
of the batch.

\textbf{Cohen's $\kappa$ trajectory.} For each annotator and each
batch, we compute Cohen's $\kappa$ on binary shadow/background
agreement between the annotator's mask and the frozen gold reference,
pooled over the $5$ gold images. Mean $\kappa$ across the seven
annotators rises from Phase~1 to Phase~2 as model seeding stabilizes
the label distribution, and rises further after Phase~3 adjudication
(Table~\ref{tab:kappa}). All Phase~2 and Phase~3 values fall within the
substantial-to-almost-perfect range
($\kappa\!\geq\!0.80$)~\cite{landis1977measurement}.

\begin{table}[h]
  \centering
  \small
  \setlength{\tabcolsep}{6pt}
  \caption{
    \textbf{Inter-annotator agreement across phases.}
    Mean Cohen's $\kappa$ on gold images across the seven annotators,
    with the across-annotator range and the per-batch first-submission
    acceptance rate (\S\ref{app:gates}). Phase~3 reports
    post-adjudication agreement against the final reconciled corpus.
  }
  \label{tab:kappa}
  \begin{tabular}{@{}lcccc@{}}
    \toprule
    \textbf{Phase} & \textbf{Sessions} & \textbf{Mean $\kappa$} & \textbf{Range across annotators} & \textbf{First-submission accept} \\
    \midrule
    Phase 1 (manual)            & $1\text{--}10$  & $0.71$ & $[0.62,\,0.78]$ & $84\%$ \\
    Phase 2 (model-in-the-loop) & $11\text{--}45$ & $0.84$ & $[0.79,\,0.89]$ & $96\%$ \\
    Phase 3 (post-adjudication) & ---                   & $0.89$ & $[0.85,\,0.92]$ & --- \\
    \bottomrule
  \end{tabular}
\end{table}

\textbf{Acceptance gates.}
\label{app:gates}
A batch is accepted when its mean gold mask-IoU against the frozen
reference is at least $0.90$; spot-checked on $5$--$10$ randomly
selected working images by a senior annotator before entering the
corpus. A batch with mean gold IoU between $0.80$ and $0.90$ is
conditionally accepted: the annotator receives written feedback on the
failed gold images, and the next batch dispatched to that annotator
includes $5$ additional gold images. A batch with mean gold IoU below
$0.80$ is rejected and the failed images returned for redo; a second
failed submission triggers a paired-review session with a senior
annotator before the next solo batch is dispatched. Across all
sessions, $96\%$ of batches were accepted on first submission, and
$4\%$ were resubmitted.

\subsection{Annotator Recruitment and Instructions}
\label{app:annotator-instructions}

Annotation was paid, on-demand work carried out in India. The full text of the recruitment and task instructions given to annotators follows; amounts are in Indian rupees (INR). In practice every accepted image was paid INR~20, and each review by a second annotator was paid a further INR~20.

\begin{quote}\small
\textbf{Mark Shadow Areas on Images (Web Tool)}

\textit{Why this work matters.} We're building a high-quality research dataset that helps improve mapping and city planning. Your careful work directly contributes to real-world research used by universities and labs.

\textit{What you'll do.} Open a small set of images (satellite view) in a web tool. Mark areas that are in shadow by drawing/adjusting shapes. Save and move to the next image. That's it. No coding. No prior mapping experience required.

\textit{Training \& support.} Short written guide with pictures. Quick sample demo. We provide fast feedback so you can improve and earn the quality bonus.

\textit{Payment (per accepted image).} INR 20 per image. Each session is ${\sim}100$ images, so the typical total is INR 1,500--2,000. Payments are made after review of each session.

\textit{How we review quality.} Each batch passed through a validation AI model. We do a short spot-check on a few others. If any images need fixes, we'll send them back with quick notes; you'll be paid for the final images once approved.

\textit{Selection process (short test).} You'll get a mini test (5 images) in the same web tool. If the test looks good, we'll invite you for paid work immediately. Top performers get more sessions each week.

\end{quote}

\section{Per-City Dataset Statistics and Shadow Taxonomy}
\label{app:data_stats}

This appendix documents the six-category shadow taxonomy referenced in
Section~\ref{sec:annotation} (\S\ref{app:taxonomy}), the per-city and
per-resolution distributional statistics that motivate the LOCO design
(\S\ref{app:perCity}), and the spatial sampling protocol used to draw
patches within each city (\S\ref{app:sampling}).

\subsection{Shadow Taxonomy}
\label{app:taxonomy}

Each annotated shadow segment carries one of six type labels. The taxonomy
is released with the dataset for downstream analyses; the LOCO benchmark
(\S\ref{sec:loco}) collapses these to a binary mask. Definitions and
inclusion rules are in Table~\ref{tab:taxonomy}. Annotators applied a
single inclusion rule across all six types: a region is shadow if it is
solar-directional and consistent with local scene geometry. Sensor
artifacts, dark water, and cloud shadows lacking geometric consistency are
excluded.

\begin{table}[h]
  \centering
  \small
  \setlength{\tabcolsep}{4pt}
  \caption{
    \textbf{Six-category shadow taxonomy.}
    Every annotated segment receives exactly one label.
    The binary LOCO benchmark merges all six into the foreground class.
  }
  \label{tab:taxonomy}
  \begin{tabular}{@{}p{3.0cm}p{9.5cm}@{}}
    \toprule
    \textbf{Category} & \textbf{Definition and inclusion rule} \\
    \midrule
    Building / canyon &
    Cast shadows from buildings and built structures, including narrow
    inter-building canyons. \\
    Under-structure &
    Persistent occlusion shadows beneath elevated structures (bridges,
    overpasses, awnings, carports). \\
    Tree-canopy dapple &
    Shadow patterns cast by tree foliage, including dappled shadow on the
    ground beneath the canopy. \\
    Topography-cast &
    Shadows produced by terrain elevation changes (hills, embankments,
    cuttings). \\
    Vehicle-cast &
    Shadows cast by cars, trucks, and other ground vehicles. \\
    Thin linear &
    Shadows from thin vertical structures (poles, fences, antennas, wires)
    whose width is on the order of a few pixels. \\
    \bottomrule
  \end{tabular}
\end{table}

\subsection{Per-City and Per-Resolution Statistics}
\label{app:perCity}

\begin{figure}[h!]
  \centering
  \includegraphics[width=\linewidth]{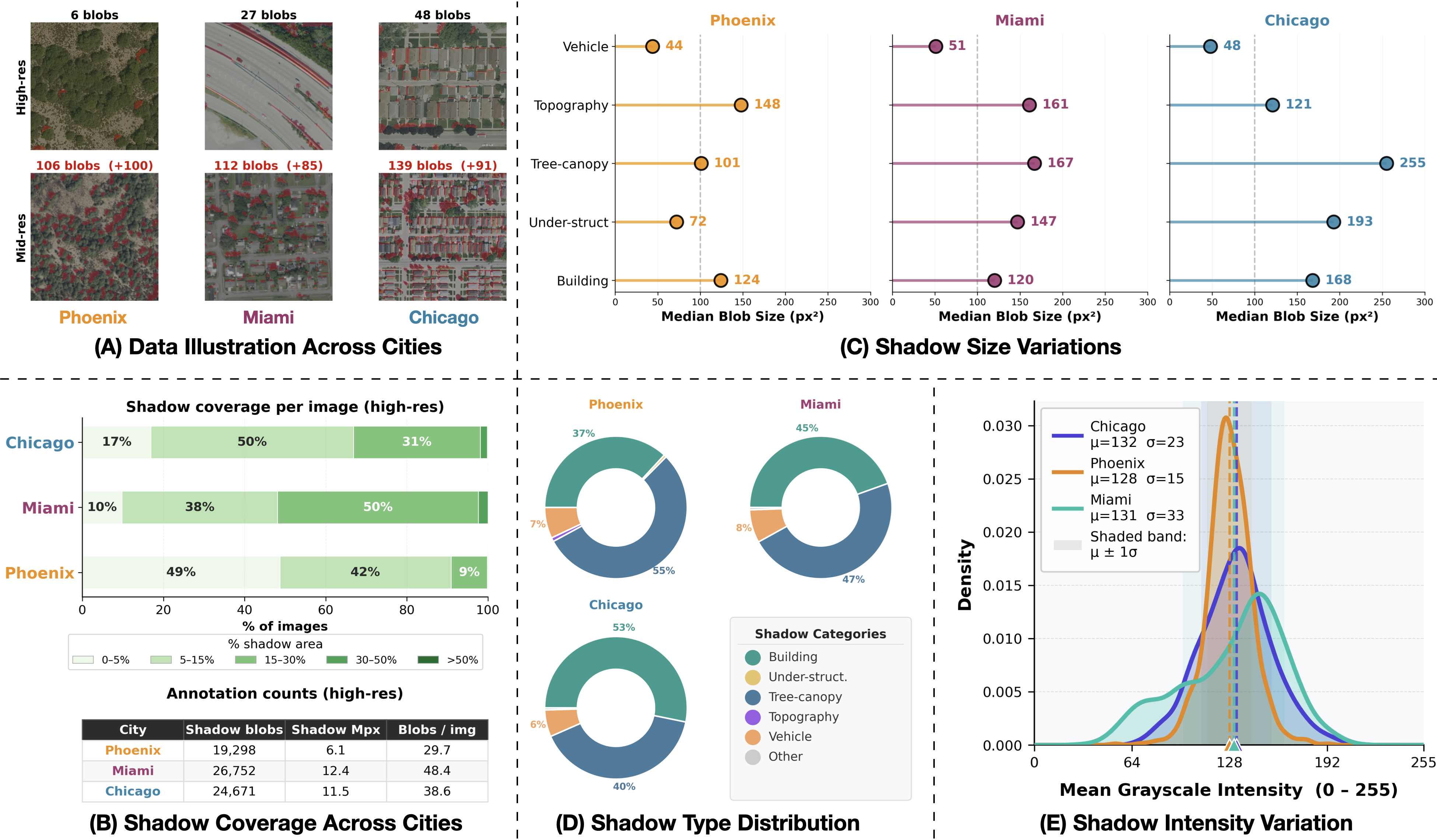}
  \caption{
    \textbf{\shadow dataset statistics.}
    (A) Sample patches across city-resolution pairs with shadow annotation overlays.
    (B) Shadow-coverage distribution per city (high-res); Phoenix patches are
    predominantly high-coverage ($\ge$30\% shadow area), driving the surface-albedo
    shortcut.
    (C) Median shadow blob size per category and city, showing strong city-level
    variation in shadow morphology.
    (D) Shadow-category distribution per city.
    (E) Mean grayscale intensity distributions per city ($\mu \approx 128$-$132$,
    $\sigma = 15$-$33$); Phoenix's lower standard deviation reflects more uniform
    high-albedo surfaces.
  }
  \label{fig:stats}
\end{figure}

Figure~\ref{fig:stats} reports the distributional properties of the
benchmark. Four observations bear on the LOCO design and on results
reported in Section~\ref{sec:evaluation}.

\textbf{Shadow coverage is city-dependent (panel B).}
Phoenix is dominated by low-coverage patches: $49\%$ of high-resolution
images contain $5$--$15\%$ shadow area and only $9\%$ exceed $30\%$. Miami
and Chicago are shifted toward higher coverage ($50\%$ and $31\%$ of
images above $30\%$ coverage, respectively). Per-image annotation density
follows the same trend ($29.7$ blobs/image in Phoenix, $48.4$ in Miami,
$38.6$ in Chicago).

\textbf{Shadow morphology varies by city (panels C, D).}
Chicago has the largest median tree-canopy blobs ($255$\,px$^2$) and
under-structure blobs ($193$\,px$^2$), reflecting a denser urban grid with
mature street trees. Phoenix shows the smallest under-structure blobs
($72$\,px$^2$) and a vehicle-shadow share of $7\%$, the highest among the
three cities. Miami's distribution is dominated by tree-canopy shadow
($47\%$ of segments) consistent with its dense subtropical canopy. Chicago
is dominated by building shadow ($53\%$).

\textbf{Surface intensity is city-dependent (panel E).}
Mean grayscale intensity distributions overlap in mean
($\mu\!\in\![128, 132]$) but differ in spread: Phoenix
($\sigma\!=\!15$) is roughly half as wide as Miami ($\sigma\!=\!33$). The
narrow Phoenix distribution is consistent with uniform high-albedo desert
surfaces; the wide Miami distribution reflects a mix of bright sand,
asphalt, and dark vegetation.

\textbf{Native resolution affects blob count, not city ranking (panel A).}
The same $384{\times}384$ ground footprint contains substantially more
shadow blobs at $0.6$\,m/px than at $0.3$\,m/px ($+85$ to $+100$ blobs in
the illustrated examples) because the lower-resolution patch covers
$4{\times}$ more area on the ground. The relative ordering across cities is
preserved across resolutions.

These statistics establish that the three cities differ along orthogonal
axes (coverage, morphology, surface intensity) rather than along a single
appearance axis, supporting the LOCO claim in Section~\ref{sec:design}
that the protocol measures transfer across genuine appearance shifts.

\subsection{Spatial Sampling Protocol}
\label{app:sampling}

Patches are drawn from NAIP orthoimagery in two streams: \emph{random}
patches that provide diverse independent samples, and \emph{paired}
patches that share ground coordinates across the two native resolutions.

\textbf{Random patches.}
Within each (city, resolution) pair, target patch counts are distributed
across NAIP tiles overlapping the city administrative boundary. If the
target count is at most the tile count, tiles are sampled without
replacement and each sampled tile contributes one patch; otherwise the
count is distributed evenly with the remainder allocated to randomly
chosen tiles. Within each selected tile, a candidate center is sampled
uniformly inside the tile geometry and accepted only if (i) the point
lies within the city boundary and (ii) a $384{\times}384$ window centered
at the point fits entirely inside the tile (a $192$-pixel buffer to all
edges). Up to $50$ candidates are tried per slot; unfilled slots are
skipped. The one-patch-per-tile rule limits spatial autocorrelation
between samples.

\textbf{Paired patches.}
For each city, $150$ ground coordinates are drawn under the random-patch
procedure on the mid-resolution stream ($0.6$\,m/px), with the additional
constraint that each new center is at least $384$ pixels (one patch
width) from every previously accepted center in the same tile. The
accepted longitude/latitude coordinates are written to a shared registry
and reused for the high-resolution stream ($0.3$\,m/px): for each saved
coordinate, the high-resolution NAIP tile containing the location is
identified and a $384{\times}384$ window is extracted centered on the
exact same geographic point. Each paired patch carries a shared
\texttt{pair\_id} linking its two resolutions. Paired patches support
resolution-matched analyses; results reported in the main paper use the
combined random$+$paired pool unless stated otherwise.

\textbf{Reproducibility.}
Sampling is organized into sequential sessions ($40$ random and $10$
paired patches per resolution per session), each with a distinct random
seed. Per-patch metadata (\texttt{city}, \texttt{scale\_gsd\_m},
\texttt{lon}, \texttt{lat}, \texttt{source\_tile\_id}, \texttt{pair\_id})
is released with the benchmark, enabling exact patch-level reproduction
of the train/val/test splits.

\section{Split Distribution Characterization}
\label{app:splits}

Overhead remote-sensing benchmarks rarely verify that train, validation, and test splits are distributionally matched within each domain, despite documented evidence that within-domain spatial autocorrelation can inflate apparent performance by up to $28\%$~\cite{kattenborn2022spatially}. We close this gap for \shadow{} with two complementary diagnostic suites: omnibus feature-space tests that ask whether the splits are distinguishable at all (\S\ref{app:splits-omnibus}), and a task-relevant feature panel that localizes any divergence to interpretable axes (\S\ref{app:splits-features}). The diagnostics confirm that the LOCO Upper Bound measures within-city generalization rather than within-city heterogeneity, and they isolate one bounded caveat (\S\ref{app:splits-implications}) for transparent reporting.

\subsection{Omnibus Diagnostics}
\label{app:splits-omnibus}

For each of the six $\langle$city, resolution$\rangle$ pairs we run four within-city tests comparing train against test. \textbf{(D1)} Two-sample Kolmogorov-Smirnov on label statistics (per-image shadow coverage, blob count, mean blob area), reporting KS distance $D$ with image-level bootstrap CI ($B{=}1{,}000$); $D > 0.15$ at $p < 0.05$ flagged. \textbf{(D2)} Domain-classifier covariate-shift test: ImageNet ResNet-50 features feed a logistic regression trained to distinguish train from test images of the same city under $5$-fold cross-validation; ROC-AUC $> 0.7$ flagged. \textbf{(D3)} Bootstrap maximum mean discrepancy (MMD) with RBF kernel under the median-heuristic bandwidth on DINOv2 features ($n = 200$ per split, $B = 1{,}000$); $p < 0.05$ flagged. \textbf{(D4)} Spatial coverage: per-split longitude/latitude range, mean test-to-nearest-train distance, and KS on each coordinate.

\begin{table}[h]
\centering
\small
\setlength{\tabcolsep}{6pt}
\caption{
\textbf{Within-city split diagnostics.}
\textit{Mean $D$}: mean KS distance over the three label statistics (D1).
\textit{AUC}: domain-classifier ROC-AUC (D2).
\textit{MMD $p$}: bootstrap $p$-value (D3).
\textit{Spat.}: KS on (lon, lat) (D4).
$\dagger$: flagged ($D > 0.15$, AUC $> 0.7$, or $p < 0.05$).
Five of six cells show no significant mismatch; Chicago highres is the single exception.
}
\label{tab:splits-omnibus}
\begin{tabular}{llcccc}
\toprule
\textbf{City} & \textbf{Res.} & \textbf{Mean $D$} & \textbf{AUC} & \textbf{MMD $p$} & \textbf{Spat.} \\
\midrule
Chicago & High ($0.3$\,m) & $0.18^{\dagger}$ & $0.73^{\dagger}$ & $0.02^{\dagger}$ & $0.21^{\dagger}$ \\
Chicago & Mid  ($0.6$\,m) & $0.09$           & $0.56$           & $0.41$           & $0.08$           \\
Miami   & High ($0.3$\,m) & $0.07$           & $0.54$           & $0.62$           & $0.11$           \\
Miami   & Mid  ($0.6$\,m) & $0.08$           & $0.58$           & $0.34$           & $0.13$           \\
Phoenix & High ($0.3$\,m) & $0.08$           & $0.51$           & $0.71$           & $0.07$           \\
Phoenix & Mid  ($0.6$\,m) & $0.13$           & $0.62$           & $0.18$           & $0.10$           \\
\bottomrule
\end{tabular}
\end{table}

Phoenix shows the cleanest splits of any city by every diagnostic. The single flagged cell, Chicago highres, traces to a structural property of the sampling protocol: test images come from the paired-coordinate subset (\S\ref{app:sampling}), which can only be drawn from regions where NAIP provides overlapping coverage at both native resolutions. In Chicago that dual-coverage zone is geographically narrower than the random-train footprint (test latitudes start at $41.687$ vs.\ $41.748$ for train), producing the spatial concentration captured by D4 and propagating to the feature-space tests.

\subsection{Per-Feature Divergence Panel}
\label{app:splits-features}

The omnibus suite answers whether train and test are distinguishable; for practitioners the more useful question is which task-relevant features are mismatched and by how much. We compute the train-vs-test KS distance separately on a panel of nine features that mechanically affect mIoU: shadow coverage fraction, blob count per image, mean blob area, small-object fraction, shadow-category mix (total variation in place of KS), and per-image mean intensity for each of R/G/B and the grayscale standard deviation.

\begin{table}[h]
\centering
\small
\setlength{\tabcolsep}{6pt}
\caption{
\textbf{Per-feature train-vs-test divergence.}
\textit{Max $D$}: largest single-feature KS distance over the nine-feature panel.
\textit{Mean $D$}: panel average.
\textit{Sig./$9$}: count of features with $D > 0.15$ and $p < 0.05$.
Chicago highres flags eight of nine features at substantial magnitudes; remaining cells flag at most five features, all at the threshold.
}
\label{tab:splits-features}
\begin{tabular}{llccc}
\toprule
\textbf{City} & \textbf{Res.} & \textbf{Max $D$} & \textbf{Mean $D$} & \textbf{Sig./$9$} \\
\midrule
Chicago & High & $0.21$ & $0.14$ & $8$ \\
Chicago & Mid  & $0.12$ & $0.07$ & $0$ \\
Miami   & High & $0.13$ & $0.08$ & $1$ \\
Miami   & Mid  & $0.14$ & $0.09$ & $2$ \\
Phoenix & High & $0.12$ & $0.08$ & $0$ \\
Phoenix & Mid  & $0.15$ & $0.11$ & $5$ \\
\bottomrule
\end{tabular}
\end{table}

The panel reproduces the omnibus picture at finer resolution and quantifies the size of the divergences. Chicago highres flags eight of nine features: shadow coverage shifts from $0.103$ in train to $0.123$ in test ($+19\%$ relative), median blob area shifts by $+42\%$, and blob count by $+18\%$. These shifts are consistent with the dual-coverage geographic concentration documented above, the paired test footprint over-samples denser-shadow neighborhoods in central Chicago. Phoenix midres flags five features, but each at the threshold ($D \in [0.13, 0.15]$, relative shifts $2$--$5\%$), an order of magnitude smaller than the Chicago highres effect. Phoenix highres flags zero features and posts the lowest mean $D$ in the benchmark.

\subsection{Implications for LOCO Reporting}
\label{app:splits-implications}

Two conclusions follow.

\textbf{Phoenix's LOCO degradation reflects cross-city transfer, not split heterogeneity.}
Phoenix's within-city splits are statistically matched on every diagnostic, and the LOCO protocol does not draw on Phoenix's training split when Phoenix is held out. Phoenix's lower Upper Bound is partly mechanical: shadow coverage in Phoenix ($5.7$--$6.7\%$) is roughly half that in Chicago/Miami ($8.7$--$15.1\%$), the small-object fraction ($0.80$--$0.82$ vs.\ $0.41$--$0.50$) is nearly twice as high, and mean R-channel intensity differs by $\sim 20$ units (\S\ref{app:perCity}). All three properties suppress mIoU regardless of model quality. The cross-city gap reported in Table~\ref{tab:loco-main} is therefore a genuine transfer phenomenon, not a split artifact.

\textbf{Chicago highres carries a known, bounded caveat.}
The paired-coordinate test set covers a narrower geographic band than the random training set, producing the $+19\%$ shift in test-set shadow coverage documented above. This affects Chicago's Upper Bound (and therefore the magnitude of $\Delta$ in cells where Chicago is held out) but does not introduce circularity into the held-out-city evaluation, since LOCO never co-locates Chicago train and Chicago test. The direction of bias is conservative for our headline claim: Chicago already posts the highest Upper Bound in Table~\ref{tab:loco-main}, so a slight under-estimate of its true Upper Bound would only widen the reported LOCO gap. We disclose the caveat explicitly and release per-patch metadata (\texttt{pair\_id}, \texttt{lon}, \texttt{lat}, \texttt{source\_tile\_id}) so downstream users can re-split if a fully matched protocol is required for their analysis.

\section{Metric Definition}
\label{app:metric}

This appendix formalizes the boundary-tolerant mIoU used as the primary
metric in Section~\ref{sec:loco}: the $\pm 2$-pixel don't-care band, the
per-image IoU on the valid region, the aggregation across images, and the
rule for degenerate cases.

\textbf{Don't-care band.}
For a binary ground-truth mask $Y \in \{0,1\}^{H \times W}$ and a tolerance
$K = 2$ pixels, let $\mathcal{B}_K$ be an elliptical structuring element of
radius $K$ (OpenCV \texttt{MORPH\_ELLIPSE}, kernel size
$(2K+1) \times (2K+1)$). The don't-care band $B$ and the valid evaluation
mask $V$ are
\begin{equation}
B = \text{dilate}(Y; \mathcal{B}_K) \setminus \text{erode}(Y; \mathcal{B}_K),
\qquad
V = \neg B.
\label{eq:band}
\end{equation}
Pixels inside $\text{erode}(Y; \mathcal{B}_K)$ are ``definite shadow''; pixels
outside $\text{dilate}(Y; \mathcal{B}_K)$ are ``definite background''; pixels
in $B$ are excluded from numerator and denominator alike, so errors inside
the band are neither rewarded nor penalized. Setting $K = 0$ recovers the
strict variant ($B = \emptyset$, $V$ covers the whole image).

\textbf{Boundary-tolerant IoU.}
For a binary prediction $\hat{Y}$, the per-image foreground IoU on the valid
region is
\begin{equation}
\text{IoU}^{\text{tol}}_K(\hat{Y}, Y) =
\frac{\sum_{p \in V} \mathbb{1}\!\left[\hat{Y}_p = 1 \,\wedge\, Y_p = 1\right]}
     {\sum_{p \in V} \mathbb{1}\!\left[\hat{Y}_p = 1 \,\vee\, Y_p = 1\right]}.
\label{eq:iou-tol}
\end{equation}
The background IoU $\text{IoU}^{\text{tol,bg}}_K$ is defined in the same way with the two
labels exchanged. Throughout the paper, ``mIoU'' denotes the two-class mean,
$\tfrac{1}{2}\big(\text{IoU}^{\text{tol}}_K + \text{IoU}^{\text{tol,bg}}_K\big)$, computed per
image; we report the foreground IoU of Equation~\ref{eq:iou-tol} separately where a statement
concerns the shadow class alone.

\textbf{Aggregation.}
We report the per-image mean across the $N = 150$ test images per cell,
\begin{equation}
\text{mIoU}^{\text{tol}}_K = \frac{1}{N} \sum_{i=1}^{N}
\tfrac{1}{2}\Big(\text{IoU}^{\text{tol}}_K(\hat{Y}_i, Y_i) + \text{IoU}^{\text{tol,bg}}_K(\hat{Y}_i, Y_i)\Big).
\label{eq:miou}
\end{equation}
We do not micro-pool TP/FP/FN counts across images. Per-image macro
aggregation gives each patch equal weight regardless of shadow coverage,
which matters because coverage varies by city (Appendix~\ref{app:perCity});
under micro-pooling, coverage-heavy patches would dominate the cell mean and
conflate geographic transfer with shadow-mass distribution. Per-image
scalars are also required by the paired-bootstrap protocol in
Section~\ref{sec:loco}: the bootstrap resamples per-image $\Delta$ values to
construct cell-level confidence intervals.

\textbf{Degenerate cases.}
Two cases require an explicit rule. \emph{(i)~Components consumed by the
band.} If a connected shadow component is thin or small enough that
$\text{erode}(Y; \mathcal{B}_K)$ removes it entirely, the component
contributes zero pixels to both numerator and denominator and is silently
dropped from Equation~\ref{eq:iou-tol}. We track such components in the
size-stratified ``tiny'' bin (Appendix~\ref{app:loco_extended}); under
$K = 2$ this affects components narrower than $\sim\!5$ pixels in any
direction. \emph{(ii)~No foreground in $V$.} If an image has no shadow on
$V$, the IoU denominator is zero; we assign $\text{IoU} = 1$ when the
prediction also has no positives on $V$ (perfect agreement) and
$\text{IoU} = 0$ otherwise. This case affects $13.0\%$ of test images
in \shadow{} ($117$ of $900$ at $K = 2$) and follows the standard convention in binary segmentation.

\textbf{Rationale for the $\pm 2$-pixel tolerance.}
Annotators draw boundaries on the pixel grid and SAM-proposed boundaries
inherit the same quantization, but the true projected shadow edge depends
on the sub-pixel position of the occluder and the sun angle, so pixel-exact
boundary agreement is unachievable in principle. At $0.3$\,m/px, two pixels
correspond to $\sim\!0.6$\,m on the ground --- smaller than the typical
width of a tree trunk or roof eave, comparable to the annotator-quantization
scale on gold images, and well below the scale of genuine geometric errors
that LOCO transfer should expose. Every number in the paper uses $K = 2$.

\section{Training configurations}
\label{app:training}

This appendix documents the architecture and training configuration for
each of the four architectures (MAMNet, OGLANet, DINOv3, CrossEarth)
and the eight adaptation methods (FDA, SegDeSiC, IIM, ISW, MRFP+, FADA,
BSM, SAR).  SegDeSiC, IIM, ISW, MRFP+ and FADA are not run on
CrossEarth, whose query-based Mask2Former decoder has no per-pixel head
these methods are defined on.  Each adaptation block lists only those
hyperparameters that differ from the corresponding base configuration;
values matching the base are omitted.  

\subsection{MAMNet}
\label{sec:mamnet-base}

\textbf{Architecture.}
ResNet-34~\cite{he2016deep} backbone with pretrained weights
\texttt{ResNet34\_Weights.IMAGENET1K\_V1} (torchvision); \texttt{conv1}
stride changed from 2 to 1 and \texttt{maxpool} replaced by
\texttt{nn.Identity}, so the encoder returns five feature maps at
channels $\{64,128,256,512,512\}$ with output strides
$\{1,2,4,8,16\}$ relative to the $384\times384$ input
(\texttt{feat5}: $512{\times}24{\times}24$).
The MSCAF module operates on \texttt{feat5} before the decoder.
The decoder comprises four \texttt{DecoderBlock} stages
($512{\to}256{\to}128{\to}64{\to}64$); each block applies Criss-Cross
Attention (CCA) to the incoming skip-connection feature, bilinearly
upsamples the previous decoder map, concatenates, and refines with
two $3{\times}3$-BN-ReLU convolutions.  CCA uses $1{\times}1$
projections to $C/8$ channels for $Q,K$ and $C$ for $V$; row- and
column-wise attention maps are summed and gated by a learnable scalar
$\gamma$ (init.\ 0): $\text{out} = \gamma(H{+}W) + x$.  Three
auxiliary branches (AUX1--3) attach to decoder outputs at $256/128/64$
channels (Conv$3{\times}3$-BN-ReLU-Dropout(0.3)-Conv$1{\times}1$,
bilinearly upsampled to input resolution).
Input is $384{\times}384$ four-channel RGBC; the contrast-channel
\texttt{conv1} weight is initialised as the channel-mean of the
pretrained RGB weights.

\begin{table}[h]
\centering
\caption{MAMNet training configuration.
$^\dagger$Shell script overrides the Python-file default of
$1{\times}10^{-3}$.}
\label{tab:mamnet-config}
\small
\begin{tabular}{@{}ll@{}}
\toprule
\textbf{Hyperparameter} & \textbf{Value} \\
\midrule
\multicolumn{2}{@{}l}{\textit{Optimisation}} \\
Optimizer        & Adam \\
Learning rate    & $1{\times}10^{-4}$$^\dagger$ \\
Weight decay     & $1{\times}10^{-4}$ \\
Batch size       & 8 \\
Epochs           & 100 \\
LR scheduler     & \texttt{ReduceLROnPlateau}
                   (mode=\texttt{max}, factor=0.5, patience=3) \\
Decision metric  & Per-image tolerant mIoU ($\pm$2\,px) \\
Early stopping   & patience = 10 epochs \\
\midrule
\multicolumn{2}{@{}l}{\textit{Loss}} \\
Main loss        & \texttt{nn.CrossEntropyLoss} (2-class, unweighted) \\
Total loss       & $\mathcal{L}_\text{main} + 0.4\cdot
                    \tfrac{1}{3}\sum_{k=1}^{3}\mathcal{L}_{\text{aux}_k}$ \\
\midrule
\multicolumn{2}{@{}l}{\textit{Data}} \\
Input size       & $384{\times}384$ \\
Channels         & 4 (RGBC) \\
Train aug.       & H-flip $p{=}0.5$, V-flip $p{=}0.5$ \\
Val/test aug.    & None (resize + ImageNet normalisation only) \\
Post-processing  & Remove connected components $<10$\,px \\
\midrule
\multicolumn{2}{@{}l}{\textit{Infrastructure}} \\
Hardware         & 1$\times$ NVIDIA A100 (NCSA Delta, \texttt{gpuA100x4}) \\
\bottomrule
\end{tabular}
\end{table}

\subsection{MAMNet + FDA}

\textbf{MAMNet\,+\,FDA.}
FDA~\cite{fda} is applied at the input stage, before resize and
normalisation: each source training image (uint8 PIL $\to$ numpy) is
passed through \texttt{FDATransform} on CPU.  Per-channel 2-D FFT
separates amplitude and phase; the source amplitude rectangle of
half-width $b{=}\lfloor\min(H,W)\cdot L\rfloor$ centred on the shifted
spectrum is replaced element-wise by the corresponding target
rectangle (no blending), and the image is reconstructed from the
mutated amplitude with the source phase
$\hat{\mathbf{s}} = \operatorname{Re}\bigl(\mathcal{F}^{-1}(\hat{\mathcal{A}}_s
e^{i\mathcal{P}_s})\bigr)$, clipped to $[0,255]$.
With $L{=}0.01$ and $\min(H,W){=}384$, the swapped region is $6{\times}6$
per channel.  Validation and test images are unmodified.  Run in LOCO
mode only; the target pool is up to 100 unlabelled images sampled
uniformly per iteration from the held-out city's \texttt{train/images}.

\begin{table}[h]
\centering
\caption{MAMNet\,+\,FDA hyperparameters; settings matching
Table~\ref{tab:mamnet-config} are omitted.}
\label{tab:mamnet-fda}
\small
\begin{tabular}{@{}ll@{}}
\toprule
\textbf{Hyperparameter} & \textbf{Value} \\
\midrule
FDA low-freq.\ ratio $L$ & 0.01 \\
Target pool size         & $\leq 100$ images \\
Target domain            & Held-out city \texttt{train/images} (unlabelled) \\
Application point        & Input (pre-resize, train only) \\
Target sampling          & Uniform random per iteration \\
Early-stopping patience  & 15 epochs \\
Modes run                & LOCO only \\
\bottomrule
\end{tabular}
\end{table}

\subsection{MAMNet + SegDeSiC}

\textbf{MAMNet\,+\,SegDeSiC.}
SegDeSiC~\cite{segdesicnet} attaches a geographic coordinate
branch to \texttt{feat5} ($512{\times}24{\times}24$): global average
pool $\to$ MLP
($512{\to}256{\to}256{\to}40$, ReLU + Dropout(0.3) between layers)
$\to$ $\ell_2$ normalisation, predicting a 40-d GRID encoding of
$(\varphi,\lambda)$ produced by sphere projection
($x{=}\cos\varphi\cos\lambda$, $y{=}\cos\varphi\sin\lambda$) and
$[\sin(\omega_k x),\cos(\omega_k x),\sin(\omega_k y),\cos(\omega_k y)]$
across 10 log-spaced wavelengths
$\lambda_k\in[10^{-4},10^{4}]$ ($\omega_k{=}2\pi/\lambda_k$).
The domain loss is
$\mathcal{L}_{\text{dom}}{=}1-\cos\langle\hat{e},e^{*}\rangle$,
batch-mean.
In LOCO mode the loss is computed on both the labelled source batch
and an unlabelled target batch (held-out city, no segmentation labels);
both terms are summed before the $\alpha$ scaling.  No coordinates are
passed at validation or test time.  

\begin{table}[h]
\centering
\caption{MAMNet\,+\,SegDeSiC hyperparameters; settings matching
Table~\ref{tab:mamnet-config} are omitted.
$^\dagger$No submit script provided; \texttt{argparse} default may have
been overridden at launch.}
\label{tab:mamnet-segdesic}
\small
\begin{tabular}{@{}ll@{}}
\toprule
\textbf{Hyperparameter} & \textbf{Value} \\
\midrule
\multicolumn{2}{@{}l}{\textit{Geographic branch}} \\
Attachment point         & \texttt{feat5} ($512{\times}24{\times}24$) \\
GRID scales / range      & 10 / $[10^{-4},10^{4}]$ log-spaced \\
Encoding dim             & 40 ($4{\times}10$) \\
MLP hidden dim / dropout & 256 / 0.3 (both layers) \\
\midrule
\multicolumn{2}{@{}l}{\textit{Loss}} \\
Domain loss              & $1-\cos\langle\hat{e},e^{*}\rangle$, batch mean \\
$\alpha$ (domain weight) & 0.5 \\
Total loss               & $\mathcal{L}_{\text{seg}}
                            + \alpha(\mathcal{L}_{\text{dom,src}}
                                    + \mathcal{L}_{\text{dom,tgt}})$ \\
CE ignore index          & 255 \\
\midrule
\multicolumn{2}{@{}l}{\textit{Optimisation}} \\
Learning rate            & $1{\times}10^{-3}$ \\
Early-stopping patience  & 15 epochs \\
\midrule
\multicolumn{2}{@{}l}{\textit{Data}} \\
Target domain (LOCO)     & Held-out city \texttt{train/images} (unlabelled) \\
Geo.\ metadata           & Per-image lat/lon in \texttt{mapping.json} \\
Modes run                & LOCO only \\
\bottomrule
\end{tabular}
\end{table}

\subsection{MAMNet + IIM}

\textbf{MAMNet\,+\,IIM.}
IIM~\cite{yola} is an input-level preprocessor placed before the
backbone: input RGB is denormalised, log-transformed, and convolved by
$K{=}8$ learnable $5{\times}5$ kernels in three cross-channel pairs
($W{*}\log R-W{*}\log B$, $W{*}\log R-W{*}\log G$,
$W{*}\log G-W{*}\log B$), producing 24 channels.  A FuseConv block
fuses these with the original RGB through two parallel
$3{\times}3$ paths (each $\to$ BN-ReLU), concatenation, and two
$3{\times}3$ convolutions, returning a 3-channel output that replaces
the RGB input to MAMNet.  After every \texttt{optimizer.step()} each
kernel is projected to zero-mean
($W{\leftarrow}W-\overline{W}$) to enforce illumination cancellation
in the log-domain.
A consistency loss
$\mathcal{L}_{\text{II}} = \operatorname{SmoothL1}\!\bigl(
\text{IIM}(I),\,\text{IIM}(\sigma_\gamma I)\bigr)$
with random $\gamma\sim U[0.5,2.0]$ regularises kernel invariance; an
adaptive weight
$\eta = r\cdot\mathcal{L}_{\text{task}}/(\mathcal{L}_{\text{II}}{+}\epsilon)$
holds the II contribution at $r{=}1\%$ of the task loss
(gradients flow only through $\mathcal{L}_{\text{II}}$).
IIM adds 0.008\,M parameters.

\begin{table}[h]
\centering
\caption{MAMNet\,+\,IIM hyperparameters; settings matching
Table~\ref{tab:mamnet-config} are omitted.}
\label{tab:mamnet-iim}
\small
\begin{tabular}{@{}ll@{}}
\toprule
\textbf{Hyperparameter} & \textbf{Value} \\
\midrule
\multicolumn{2}{@{}l}{\textit{IIM module}} \\
Num.\ kernels $K$        & 8 \\
Kernel size              & $5{\times}5$ \\
Cross-channel pairs      & (R,B), (R,G), (G,B) $\to$ 24 features \\
Zero-mean projection     & After every optimizer step \\
\midrule
\multicolumn{2}{@{}l}{\textit{Loss}} \\
II loss                  & SmoothL1 ($\beta{=}1.0$) on IIM features \\
II weighting             & Adaptive, target ratio $r{=}0.01$ \\
Gamma range              & $\gamma\sim U[0.5,2.0]$ \\
Total loss               & $\mathcal{L}_{\text{task}}
                            + \eta\,\mathcal{L}_{\text{II}}$ \\
\bottomrule
\end{tabular}
\end{table}

\subsection{MAMNet + ISW}

\textbf{MAMNet\,+\,ISW.}
ISW~\cite{isw} is a training-only feature-covariance
regulariser; the inference architecture is unchanged.
\emph{Phase 1 (offline)}: each training image and its photometric
augmentation (\texttt{ColorJitter(0.4,0.4,0.4,0.1)}\,+\,\texttt{Gaussian\-Blur}
$p{=}0.5$) are passed through the encoder; the per-instance covariance
of standardised features at \texttt{feat1}, \texttt{feat2},
\texttt{feat3} is computed, and the variance of each covariance
element across the original/augmented pair is averaged over the
training set.  Upper-triangular variance values are partitioned by
$k$-means ($k{=}3$); the top $k{-}m{=}2$ clusters are flagged as
style-sensitive, producing a binary mask $\widetilde M$ per layer.
\emph{Phase 2 (training)}: at each step, forward hooks capture
features at the same three layers; per-image standardised covariance
is masked element-wise with $\widetilde M$ and reduced to
$\mathcal{L}_{\text{ISW}}^{\ell}{=}\|\widetilde M\odot\Sigma_s\|_1$
(per-image mean), then averaged across the three layers.
Total loss is $\mathcal{L}_{\text{seg}}+\lambda\,\mathcal{L}_{\text{ISW}}$.

\begin{table}[h]
\centering
\caption{MAMNet\,+\,ISW hyperparameters; settings matching
Table~\ref{tab:mamnet-config} are omitted.}
\label{tab:mamnet-isw}
\small
\begin{tabular}{@{}ll@{}}
\toprule
\textbf{Hyperparameter} & \textbf{Value} \\
\midrule
\multicolumn{2}{@{}l}{\textit{ISW configuration}} \\
$\lambda$ (ISW weight)         & 0.6 \\
Hooked layers                  & \texttt{feat1, feat2, feat3} \\
$k$-means $k$ / $m$            & 3 / 1 (bottom 1 cluster $=$ content) \\
Photometric aug.\ (mask precomp.)\!\!
  & \texttt{ColorJitter(0.4,0.4,0.4,0.1)} \\
  & + \texttt{GaussianBlur} $p{=}0.5$ \\
Loss reduction                 & Per-image mean, then layer mean \\
Total loss                     & $\mathcal{L}_{\text{seg}}
                                  + \lambda\,\mathcal{L}_{\text{ISW}}$ \\
\bottomrule
\end{tabular}
\end{table}

\subsection{MAMNet + MRFP+}

\textbf{MAMNet\,+\,MRFP+.}
MRFP+~\cite{mrfp} adds three training-only stochastic
perturbations.  HRFP attaches after \texttt{conv1}-\texttt{bn1}-\texttt{relu}
and before \texttt{layer1} (64\,ch, $H/2$): a 4-encoder/4-decoder
overcomplete autoencoder with $\sim$1.2$\times$ bilinear upsampling per
encoder layer; its conv weights are He-reinitialised and BN
$\gamma,\beta$ resampled from $\mathcal{N}(0,0.5^2)$ \emph{at every
forward pass}, and its output is added to the encoder stream under
\texttt{no\_grad}+\texttt{detach}.  NP+ attaches at \texttt{feat4}
($512{\times}48{\times}48$): channel-mean perturbation
$y = \alpha x + \delta(\beta-\alpha)\mu_c$ with $\alpha,\beta\sim
\mathcal{N}(0,1)$ and $\delta$ a normalised cross-batch variance.
HRFP+ adds the max-resolution intermediate feature from the HRFP
encoder to decoder stage \texttt{dec3}; three parameter-free
\texttt{InstanceNorm2d} layers (affine=False) are applied on
\texttt{feat1}/\texttt{feat2}/\texttt{feat3}.  Each module is toggled
independently with $p{=}0.5$.  At inference all modules are disabled
and the model is identical to base MAMNet (HRFP weights add $\sim$74\,K
frozen parameters never saved).

\begin{table}[h]
\centering
\caption{MAMNet\,+\,MRFP+ hyperparameters; settings matching
Table~\ref{tab:mamnet-config} are omitted.}
\label{tab:mamnet-mrfp}
\small
\begin{tabular}{@{}ll@{}}
\toprule
\textbf{Hyperparameter} & \textbf{Value} \\
\midrule
\multicolumn{2}{@{}l}{\textit{Perturbation modules (training only)}} \\
HRFP attach point        & After stem (64\,ch, $H/2$) \\
NP+ attach point         & \texttt{feat4} (512\,ch, $H/32$) \\
HRFP+ attach point       & Decoder \texttt{dec3} (64\,ch) \\
Inst.\ Norm layers       & \texttt{feat1}, \texttt{feat2}, \texttt{feat3}
                           (affine=False) \\
HRFP / NP+ / HRFP+ prob.\ & 0.5 / 0.5 / 0.5 (independent) \\
HRFP arch.               & 4 enc + 4 dec, $\sim$1.2$\times$ scale per layer \\
HRFP BN std $\sigma$     & 0.5 \\
HRFP weights             & Frozen, re-randomised every forward pass \\
\bottomrule
\end{tabular}
\end{table}

\subsection{MAMNet + FADA}

\textbf{MAMNet\,+\,FADA.}
FADA~\cite{fada} freezes the ResNet-34 encoder
(\texttt{requires\_grad=False}; BN in \texttt{eval} mode using
pretrained running statistics) and inserts a frequency adapter at
\texttt{feat3}, \texttt{feat4}, and \texttt{feat5}.  Each FADA block
applies a 2-D Haar DWT $\to$ $\{$LL, LH, HL, HH$\}$;
the LL subband is processed by a low-frequency branch (LoRA tokens
$T_L\in\mathbb{R}^{m\times c}$ with $T{=}AB$, rank $r$; scaled-dot
attention; MLP projection; skip fusion) for content stabilisation;
the concatenated LH/HL/HH are processed by a high-frequency branch
(separate LoRA tokens $T_H\in\mathbb{R}^{3m\times c}$; scaled-dot
attention; \emph{Instance Normalisation applied to the similarity
map}; MLP projection; skip fusion) for style suppression; the four
adapted subbands are recombined by inverse Haar DWT.  Stages 1 and 2
are excluded for memory reasons (spatial dims $384^2$ and $192^2$
yield prohibitive token attention).
Trainable parameters: $\approx$8.4\,M (FADA $\approx$2.5\,M, MSCAF +
decoder + AUX $\approx$5.9\,M); frozen $\approx$21.3\,M.

\begin{table}[h]
\centering
\caption{MAMNet\,+\,FADA hyperparameters; settings matching
Table~\ref{tab:mamnet-config} are omitted.}
\label{tab:mamnet-fada}
\small
\begin{tabular}{@{}ll@{}}
\toprule
\textbf{Hyperparameter} & \textbf{Value} \\
\midrule
\multicolumn{2}{@{}l}{\textit{FADA configuration}} \\
Adapter stages           & \texttt{feat3, feat4, feat5} \\
LoRA rank $r$            & 16 \\
Token length $m$         & 100 \\
LF branch                & LoRA tokens on LL (content) \\
HF branch                & LoRA tokens + IN on similarity map (LH/HL/HH) \\
Decomposition            & Haar DWT (orthogonal, invertible) \\
\midrule
\multicolumn{2}{@{}l}{\textit{Optimisation overrides}} \\
Backbone state           & Frozen (BN in eval mode) \\
Trainable groups         & FADA blocks + MSCAF + decoder + AUX \\
\bottomrule
\end{tabular}
\end{table}

\subsection{OGLANet}
\label{sec:oglanet-base}

\textbf{Architecture.}
OGLANet~\cite{oglanet} is a four-stage encoder--decoder with
dense skip connections.  The encoder consists of five Global-Local
Aware Module (GLAM) stages on top of a ResNet-34 backbone (pretrained
on ImageNet via \texttt{torchvision.models.resnet34}); the original
paper specifies ResNet-101, but ResNet-34 is substituted to match
MAMNet's parameter budget.  Each GLAM wraps one ResNet layer (LFEM)
between two GFEM self-attention modules ($1{\times}1$ Q/K/V
projections, sigmoid-gated, followed by $3{\times}3$ conv with
stride 2 for downsampling or stride 1 for refinement).  GLAM5 is an
additional stage (not in the original figure) that expands 512$\to$1024
channels with a stride-2 $3{\times}3$ conv plus a non-downsampling
GFEM.  The DFFM forms four decoder maps $S^D_1$--$S^D_4$ by
concatenating multi-scale encoder features iteratively downsampled
with stride-2 $3{\times}3$ convs, each followed by a $1{\times}1$
channel-reduction conv (BN-ReLU).  The decoder bilinearly upsamples
DFFM features through Conv$3{\times}3$-BN-ReLU blocks, producing
outputs at $\{12{\times}12, 48{\times}48, 96{\times}96,
384{\times}384\}$.  The OAM produces six $1{\times}1$-conv predictions:
$P_1$--$P_4$ from the four decoder scales, $P_5$ from
$\operatorname{concat}(P_1{:}P_4)$, $P_6$ from
$\operatorname{concat}(P_1{:}P_5)$.  All six are deeply supervised at
training; only $P_6$ is used at inference.  Input is $384{\times}384
{\times}4$ (RGBC); the contrast-channel \texttt{conv1} weight is
initialised as the channel-mean of the pretrained RGB weights.
Output stride 1.  

\textbf{DFFM reimplementation defect.}\label{app:oglanet-dffm}
We reimplemented the DFFM from the published architecture figure, and the
reimplementation makes OGLANet inference depend on the random state and the
batch size.  Its $20$ downsampling convolutions and their batch-normalization
layers are constructed inside \texttt{forward()} on every call and never
registered as modules.  They are absent from the checkpoint, receive fresh
random weights on every forward pass, and stay in training mode under
\texttt{model.eval()}, so they normalize with the statistics of the current
batch.  With a fixed seed, inference is bit-identical across jobs and nodes.
Over repeated unseeded passes of one checkpoint on the same $150$ test images,
the standard deviation of the cell mean reaches $0.42$ mIoU.  Changing the
inference seed among three values moves the cell mean by up to $0.66$ mIoU on
an OGLANet-\sib{} checkpoint and by up to $1.37$ mIoU on a published Vanilla
checkpoint, both on Chicago at high resolution.  Every OGLANet value in the
paper is one inference pass at its run's inference batch size, and its
confidence interval does not include this variation.  We left the module
unchanged because every reported OGLANet number was produced with it, and the
released code contains the same module.

\begin{table}[h]
\centering
\caption{OGLANet training configuration.}
\label{tab:oglanet-config}
\small
\begin{tabular}{@{}ll@{}}
\toprule
\textbf{Hyperparameter} & \textbf{Value} \\
\midrule
\multicolumn{2}{@{}l}{\textit{Optimisation}} \\
Optimizer        & Adamax \\
Learning rate    & $5{\times}10^{-4}$ \\
Weight decay     & Adamax default \\
Batch size       & 8 (script override; \texttt{train.py} default 12) \\
Epochs           & 100 \\
LR scheduler     & \texttt{ReduceLROnPlateau}
                   (mode=\texttt{max}, factor=0.5, patience=5) \\
Decision metric  & Per-image tolerant mIoU ($\pm$2\,px) \\
Early stopping   & patience = 10 epochs \\
\midrule
\multicolumn{2}{@{}l}{\textit{Loss}} \\
Loss             & \texttt{OGLANetLoss}: six CE terms
                   ($\mathcal{L}_1$--$\mathcal{L}_6$ on $P_1$--$P_6$),
                   equal weights \\
Post-processing  & Remove connected components $<10$\,px \\
\midrule
\multicolumn{2}{@{}l}{\textit{Data}} \\
Input            & $384{\times}384{\times}4$ (RGBC) \\
\midrule
\multicolumn{2}{@{}l}{\textit{Infrastructure}} \\
Hardware         & 1$\times$ NVIDIA A100 40\,GB
                   (NCSA Delta, \texttt{gpuA100x4}) \\
Mixed precision  & Disabled \\
Gradient clip.   & Disabled \\
\bottomrule
\end{tabular}
\end{table}

\subsection{OGLANet + FDA}

\textbf{OGLANet\,+\,FDA.}
FDA is applied identically to MAMNet\,+\,FDA (see \S E.2): per-channel
2-D FFT, central $2b{\times}2b$ amplitude swap with $b{=}\lfloor
\min(H,W)\cdot L\rfloor$, source phase preserved, reconstruction
clipped to $[0,255]$.  Applied as input-level preprocessing in the
dataset pipeline before standard augmentation; model architecture and
optimisation are unchanged.  Run in LOCO mode only; the target pool is
sampled uniformly per iteration from the held-out city's
\texttt{train/images}.

\begin{table}[h]
\centering
\caption{OGLANet\,+\,FDA hyperparameters; settings matching
Table~\ref{tab:oglanet-config} are omitted.}
\label{tab:oglanet-fda}
\small
\begin{tabular}{@{}ll@{}}
\toprule
\textbf{Hyperparameter} & \textbf{Value} \\
\midrule
FDA low-freq.\ ratio $L$ & 0.01 \\
Target pool size         & $\leq 100$ images \\
Target domain            & Held-out city \texttt{train/images} (unlabelled) \\
Target sampling          & Uniform random per iteration \\
Application point        & Input (dataset, pre-augmentation) \\
Modes run                & LOCO only \\
\bottomrule
\end{tabular}
\end{table}

\subsection{OGLANet + SegDeSiC}

\textbf{OGLANet\,+\,SegDeSiC.}
The SegDeSiC module is attached in parallel to \texttt{feat5}
($1024{\times}6{\times}6$, GLAMEncoder output): global average pool
$\to$ Linear(1024, 256) $\to$ ReLU+Dropout(0.3)
$\to$ Linear(256, 256) $\to$ ReLU+Dropout(0.3)
$\to$ Linear(256, 40) $\to$ $\ell_2$ normalisation; weights Xavier
uniform.  GRID encoding ($S{=}10$, $\lambda{\in}[10^{-4},10^{4}]$) and
the cosine-dissimilarity domain loss are identical to
MAMNet\,+\,SegDeSiC (\S E.3).  Total loss combines the deep-supervision
six-CE segmentation loss (equal weights $\tfrac{1}{6}$) with source and
target geographic terms scaled by $\alpha{=}0.5$ (target term
LOCO-only).  Coordinates are not supplied at validation or test time.

\begin{table}[h]
\centering
\caption{OGLANet\,+\,SegDeSiC hyperparameters; settings matching
Table~\ref{tab:oglanet-config} are omitted.}
\label{tab:oglanet-segdesic}
\small
\begin{tabular}{@{}ll@{}}
\toprule
\textbf{Hyperparameter} & \textbf{Value} \\
\midrule
\multicolumn{2}{@{}l}{\textit{SegDeSiC module}} \\
Attachment point         & \texttt{feat5} ($1024{\times}6{\times}6$) \\
GRID scales / range      & 10 / $[10^{-4},10^{4}]$ log-spaced \\
Encoding dim             & 40 \\
MLP hidden dim / dropout & 256 / 0.3 \\
\midrule
\multicolumn{2}{@{}l}{\textit{Loss}} \\
Seg.\ loss               & Mean CE over $P_1$--$P_6$, equal weights \\
Domain loss              & $1-\cos\langle\hat{e},e^{*}\rangle$, batch mean \\
$\alpha$                 & 0.5 (both source and target terms) \\
Target domain (LOCO)     & Held-out city \texttt{train} split (unlabelled) \\
\midrule
\multicolumn{2}{@{}l}{\textit{Optimisation overrides}} \\
Learning rate            & $3{\times}10^{-4}$ (vs.\ $5{\times}10^{-4}$ base) \\
Batch size               & 4 (vs.\ 8 base) \\
Modes run                & LOCO only \\
\bottomrule
\end{tabular}
\end{table}

\subsection{OGLANet + IIM}

\textbf{OGLANet\,+\,IIM.}
The IIM preprocessor (identical to \S E.4: 8 learnable $5{\times}5$
zero-mean kernels, three cross-channel pairs, FuseConv, adaptive II
loss with $r{=}0.01$ and $\gamma\sim U[0.5,2.0]$) is inserted before
the OGLANet stem.  The downstream architecture (GLAMEncoder, DFFM,
decoder, OAM) is unchanged.  The optimiser is switched from Adamax to
Adam to handle the gradient-scale gap between the small IIM module
($\approx$0.008\,M) and the pretrained ResNet-34 backbone
($\approx$21\,M).

\begin{table}[h]
\centering
\caption{OGLANet\,+\,IIM hyperparameters; settings matching
Table~\ref{tab:oglanet-config} are omitted.}
\label{tab:oglanet-iim}
\small
\begin{tabular}{@{}ll@{}}
\toprule
\textbf{Hyperparameter} & \textbf{Value} \\
\midrule
\multicolumn{2}{@{}l}{\textit{IIM module}} \\
Num.\ kernels $K$        & 8 \\
Kernel size              & $5{\times}5$ \\
Cross-channel pairs      & (R,B), (R,G), (G,B) $\to$ 24 features \\
Zero-mean projection     & After every optimizer step \\
\midrule
\multicolumn{2}{@{}l}{\textit{Loss}} \\
II loss                  & SmoothL1 ($\beta{=}1.0$) on IIM features \\
II weighting             & Adaptive, target ratio $r{=}0.01$ \\
Gamma range              & $\gamma\sim U[0.5,2.0]$ \\
\midrule
\multicolumn{2}{@{}l}{\textit{Optimisation overrides}} \\
Optimizer                & Adam (vs.\ Adamax) \\
Learning rate            & $1{\times}10^{-4}$ (vs.\ $5{\times}10^{-4}$) \\
Weight decay             & $1{\times}10^{-4}$ \\
\bottomrule
\end{tabular}
\end{table}

\subsection{OGLANet + ISW}

\textbf{OGLANet\,+\,ISW.}
ISW is applied as in \S E.5 with hooks on \texttt{glam1},
\texttt{glam2}, \texttt{glam3} (the first three GLAMEncoder outputs).
Phase 1 precomputes binary masks
$\widetilde M$ once per LOCO fold from the photometric augmentation
covariance variance, partitioned by $k$-means ($k{=}3$, $m{=}1$);
Phase 2 adds $\lambda\,\mathcal{L}_{\text{ISW}}$ to the six-CE
segmentation loss, with $\mathcal{L}_{\text{ISW}}$ averaged
per-image and across the three layers.  Inference is unchanged.

\begin{table}[h]
\centering
\caption{OGLANet\,+\,ISW hyperparameters; settings matching
Table~\ref{tab:oglanet-config} are omitted.}
\label{tab:oglanet-isw}
\small
\begin{tabular}{@{}ll@{}}
\toprule
\textbf{Hyperparameter} & \textbf{Value} \\
\midrule
$\lambda$ (ISW weight)         & 0.6 \\
Hooked layers                  & \texttt{glam1, glam2, glam3} \\
$k$-means $k$ / $m$            & 3 / 1 \\
Photometric aug.\ (mask precomp.)\!\!
  & \texttt{ColorJitter(0.4,0.4,0.4,0.1)} \\
  & + \texttt{GaussianBlur} $p{=}0.5$ \\
Loss reduction                 & Per-image mean, then layer mean \\
\bottomrule
\end{tabular}
\end{table}

\subsection{OGLANet + MRFP+}

\textbf{OGLANet\,+\,MRFP+.}
HRFP attaches after \texttt{conv1}-\texttt{bn1}-\texttt{relu} of the
GLAMEncoder stem (64\,ch, $H/2$); NP+ attaches at \texttt{glam4}
(512\,ch, $H/32$); HRFP+ adds the max-resolution intermediate HRFP
feature to the decoder's \texttt{s1\_d\_up} stage (64\,ch,
$384{\times}384$).  Modules are toggled independently with $p{=}0.5$.
\emph{The InstanceNorm layers specified in the original MRFP+ paper
are removed} (Option A): the GLAM stages end with two GFEM
self-attention modules whose channel-wise statistics are essential
to subsequent processing, and applying \texttt{InstanceNorm2d}
(affine=False) after GFEM caused class collapse (mIoU $<$1\%) in all
three cities.  HRFP and NP+ alone span both ends of the perturbation
spectrum.  HRFP weights ($\approx$74\,K) are frozen, re-randomised
each forward pass, and discarded at inference.

\begin{table}[h]
\centering
\caption{OGLANet\,+\,MRFP+ hyperparameters; settings matching
Table~\ref{tab:oglanet-config} are omitted.}
\label{tab:oglanet-mrfp}
\small
\begin{tabular}{@{}ll@{}}
\toprule
\textbf{Hyperparameter} & \textbf{Value} \\
\midrule
\multicolumn{2}{@{}l}{\textit{Perturbation modules (training only)}} \\
HRFP attach point        & After stem (64\,ch, $H/2$) \\
NP+ attach point         & \texttt{glam4} (512\,ch, $H/32$) \\
HRFP+ attach point       & Decoder \texttt{s1\_d\_up} (64\,ch, $384{\times}384$) \\
Inst.\ Norm layers       & Removed (Option A) \\
HRFP / NP+ / HRFP+ prob.\ & 0.5 / 0.5 / 0.5 \\
HRFP arch.               & 4 enc + 4 dec, $\sim$1.2$\times$ scale per layer \\
HRFP BN std $\sigma$     & 0.5 \\
\midrule
\multicolumn{2}{@{}l}{\textit{Optimisation overrides}} \\
Learning rate            & $1{\times}10^{-4}$ (vs.\ $5{\times}10^{-4}$) \\
Weight decay             & $1{\times}10^{-4}$ \\
Loss reduction           & Per-image mean CE over $P_1$--$P_6$ \\
\bottomrule
\end{tabular}
\end{table}

\subsection{OGLANet + FADA}

\textbf{OGLANet\,+\,FADA.}
Only the ResNet-34 core inside each GLAM stage
(\texttt{conv1}, \texttt{bn1}, \texttt{layer1}--\texttt{layer4}) is
frozen; the GFEM modules, \texttt{glam5\_conv}, \texttt{glam5\_gfem},
DFFM, decoder, and OAM remain trainable because they are not pretrained
and freezing them collapsed mIoU to $<$1\% in pilot runs.  FADA blocks
are inserted between the frozen LFEM output and the trainable GFEM at
stages 3 and 4 (256\,ch and 512\,ch, $24{\times}24$ and $12{\times}12$);
stages 1, 2, and 5 are excluded for either spatial-cost or
non-pretrained-component reasons.  The block internals (Haar DWT,
LL LoRA-token branch, IN-on-similarity-map HF branch, inverse Haar
DWT) are identical to MAMNet\,+\,FADA (\S E.7).  Trainable parameters:
$\approx$7--9\,M; frozen $\approx$21.3\,M.

\begin{table}[h]
\centering
\caption{OGLANet\,+\,FADA hyperparameters; settings matching
Table~\ref{tab:oglanet-config} are omitted.}
\label{tab:oglanet-fada}
\small
\begin{tabular}{@{}ll@{}}
\toprule
\textbf{Hyperparameter} & \textbf{Value} \\
\midrule
\multicolumn{2}{@{}l}{\textit{FADA configuration}} \\
Adapter stages           & GLAM 3, 4 (LFEM output, pre-GFEM) \\
LoRA rank $r$            & 16 \\
Token length $m$         & 100 \\
Decomposition            & Haar DWT \\
\midrule
\multicolumn{2}{@{}l}{\textit{Optimisation overrides}} \\
Optimizer                & Adam (vs.\ Adamax) \\
Learning rate            & $1{\times}10^{-4}$ (vs.\ $5{\times}10^{-4}$) \\
Weight decay             & $1{\times}10^{-4}$ \\
LR scheduler             & \texttt{ReduceLROnPlateau} (factor=0.5, patience=3) \\
Frozen                   & ResNet-34 core only (BN in eval mode) \\
Trainable                & FADA + GFEM + DFFM + decoder + OAM \\
\bottomrule
\end{tabular}
\end{table}

\subsection{DINOv3}
\label{sec:dinov3-base}

\textbf{Architecture.}
DINOv3 ViT-S/16 backbone (\texttt{dinov3\_vits16}): patch size
$16{\times}16$, embedding dimension $d{=}384$, 12 transformer blocks.
Pretrained weights
\texttt{dinov3\_vits16\_pretrain\_lvd1689m-08c60483.pth}
(\texttt{github.com/facebookresearch/dinov3}); all backbone parameters
trainable (\texttt{frozen\_stages}${=}{-1}$).
Intermediate features extracted after blocks 3, 6, 9, and 11 at
$H/16{\times}W/16$, concatenated along the channel axis, and fused by
a $1{\times}1$-Conv-BN-ReLU ($4{\times}384{\to}384$,
\texttt{bias=False}).  The decoder applies four
\texttt{ConvTranspose2d} ($2{\times}2$, stride 2) stages
($\tfrac{1}{16}{\to}\tfrac{1}{8}{\to}\tfrac{1}{4}{\to}\tfrac{1}{2}{\to}1$)
each followed by Conv-BN-ReLU; channel sequence
$[384, 256, 128, 64, 32]$.  No encoder skip connections (all ViT
features share the same spatial resolution); a final $1{\times}1$ conv
yields 2-class logits.  No auxiliary branches; single output head.
Input is $384{\times}384$ RGB, yielding exactly $24{\times}24$ patches
with no padding; output stride 16.  

\begin{table}[h]
\centering
\caption{DINOv3 training configuration.}
\label{tab:dinov3-config}
\small
\begin{tabular}{@{}ll@{}}
\toprule
\textbf{Hyperparameter} & \textbf{Value} \\
\midrule
\multicolumn{2}{@{}l}{\textit{Optimisation}} \\
Optimizer        & AdamW ($\beta_1{=}0.9$, $\beta_2{=}0.999$) \\
Learning rate    & $5{\times}10^{-5}$
                   (CLI; argparse default $3{\times}10^{-4}$) \\
Weight decay     & 0.05 \\
Batch size       & 8 \\
Epochs (max)     & 100 \\
LR scheduler     & Cosine + linear warmup (5 ep.,
                   $\eta_\text{min}{=}10^{-6}$) \\
Decision metric  & Per-image strict mIoU
                   (\texttt{DetailedEvaluator}) \\
Early stopping   & 20 (single, LOCO); 10 (all) \\
\midrule
\multicolumn{2}{@{}l}{\textit{Loss}} \\
Loss             & \texttt{nn.CrossEntropyLoss} (2-class, no weights,
                   no aux.) \\
Post-processing  & Remove connected components $<10$\,px \\
\midrule
\multicolumn{2}{@{}l}{\textit{Data}} \\
Input size       & $384{\times}384$ \\
Channels         & 3 (RGB) \\
Train aug.       & H-flip $p{=}0.5$, V-flip $p{=}0.5$ (PIL, train split) \\
Normalisation    & ImageNet $\mu, \sigma$ \\
\midrule
\multicolumn{2}{@{}l}{\textit{Infrastructure}} \\
Hardware         & 1$\times$ A100 (NCSA Delta, \texttt{gpuA100x4}) \\
Sampling seed    & 42 (dataset subsampling only) \\
\bottomrule
\end{tabular}
\end{table}

\subsection{DINOv3 + FDA}

\textbf{DINOv3\,+\,FDA.}
FDA is applied as in \S E.2 (per-channel 2-D FFT, central
$2b{\times}2b$ amplitude swap with $b{=}\lfloor\min(H,W)\cdot
L\rfloor$, source phase preserved, reconstruction clipped).
Inserted in the dataset pipeline before PIL$\to$tensor and ImageNet
normalisation; validation and test images are unmodified.  FDA and
SegDeSiC are mutually exclusive in all launch scripts.  Run in LOCO
mode only.

\begin{table}[h]
\centering
\caption{DINOv3\,+\,FDA hyperparameters; settings matching
Table~\ref{tab:dinov3-config} are omitted.}
\label{tab:dinov3-fda}
\small
\begin{tabular}{@{}ll@{}}
\toprule
\textbf{Hyperparameter} & \textbf{Value} \\
\midrule
FDA low-freq.\ ratio $L$ & 0.01 \\
Target pool size         & $\leq 100$ images \\
Target domain            & Held-out city \texttt{train/images} (unlabelled) \\
Application point        & Input (pre-tensor, pre-normalisation) \\
Early-stopping patience  & 10 (all modes) \\
Modes run                & LOCO only \\
\bottomrule
\end{tabular}
\end{table}

\subsection{DINOv3 + SegDeSiC}

\textbf{DINOv3\,+\,SegDeSiC.}
A \texttt{SegDesicModule} is attached to \texttt{feat\_block11}
($384{\times}24{\times}24$, the deepest extracted ViT feature).  The
module predicts a 40-d GRID encoding ($S{=}10$,
$\lambda{\in}[10^{-4},10^{4}]$) via global average pool and a 256-dim
hidden MLP, with cosine-dissimilarity loss against the ground-truth
encoding from $(\varphi,\lambda)$.  In LOCO mode an additional
unlabelled forward pass over the held-out city's training images
contributes a target-domain cosine loss; no segmentation loss is
applied to the target.  Total loss:
$\mathcal{L} = \mathcal{L}_\text{seg} + \alpha(\mathcal{L}^{\text{src}}_\text{dom}
+ \mathcal{L}^{\text{tgt}}_\text{dom})$.

\begin{table}[h]
\centering
\caption{DINOv3\,+\,SegDeSiC hyperparameters; settings matching
Table~\ref{tab:dinov3-config} are omitted.}
\label{tab:dinov3-segdesic}
\small
\begin{tabular}{@{}ll@{}}
\toprule
\textbf{Hyperparameter} & \textbf{Value} \\
\midrule
\multicolumn{2}{@{}l}{\textit{SegDeSiC module}} \\
Attachment point         & \texttt{feat\_block11} ($384{\times}24{\times}24$) \\
GRID scales / range      & 10 / $[10^{-4},10^{4}]$ \\
Encoding dim             & 40 \\
MLP hidden dim / dropout & 256 / 0.3 \\
\midrule
\multicolumn{2}{@{}l}{\textit{Loss}} \\
Domain loss              & $1-\cos\langle\hat{e},e^{*}\rangle$ \\
$\alpha$                 & 0.5 (both terms) \\
Auxiliary weight         & 0 (no aux.\ branches) \\
Target domain (LOCO)     & Held-out city \texttt{train} split (unlabelled) \\
\midrule
\multicolumn{2}{@{}l}{\textit{Optimisation overrides}} \\
Learning rate            & $3{\times}10^{-4}$ (vs.\ $5{\times}10^{-5}$) \\
Batch size               & 4 (vs.\ 8) \\
Early-stopping patience  & 10 (all modes) \\
\bottomrule
\end{tabular}
\end{table}

\subsection{DINOv3 + IIM}

\textbf{DINOv3\,+\,IIM.}
The IIM preprocessor (8 learnable $5{\times}5$ zero-mean kernels,
three cross-channel pairs, FuseConv, adaptive II loss with
$r{=}0.01$, $\gamma\sim U[0.5,2.0]$; identical to \S E.4) is inserted
before the ViT-S/16 stem.  Output is the 3-channel fused image
re-normalised with ImageNet statistics.  The base DINOv3 strict mIoU
decision metric is replaced by tolerant per-image mIoU
($\pm$2\,px) for consistency with the IIM ablation protocol.

\begin{table}[h]
\centering
\caption{DINOv3\,+\,IIM hyperparameters; settings matching
Table~\ref{tab:dinov3-config} are omitted.}
\label{tab:dinov3-iim}
\small
\begin{tabular}{@{}ll@{}}
\toprule
\textbf{Hyperparameter} & \textbf{Value} \\
\midrule
\multicolumn{2}{@{}l}{\textit{IIM module}} \\
Num.\ kernels $K$        & 8 \\
Kernel size              & $5{\times}5$ \\
Cross-channel pairs      & (R,B), (R,G), (G,B) $\to$ 24 features \\
Zero-mean projection     & After every optimizer step \\
\midrule
\multicolumn{2}{@{}l}{\textit{Loss}} \\
II loss                  & SmoothL1 ($\beta{=}1.0$) \\
II weighting             & Adaptive, $r{=}0.01$ \\
Gamma range              & $\gamma\sim U[0.5,2.0]$ \\
\midrule
\multicolumn{2}{@{}l}{\textit{Optimisation overrides}} \\
Decision metric          & Per-image tolerant mIoU ($\pm$2\,px) \\
Early-stopping patience  & 15 epochs \\
\bottomrule
\end{tabular}
\end{table}

\subsection{DINOv3 + ISW}

\textbf{DINOv3\,+\,ISW.}
ISW hooks are registered on the outputs of ViT blocks 3, 6, and 9
(block 11, used by the decoder, is left untouched).  The
\texttt{[B, N+1, D]} token output is stripped of the CLS token and
the $576$ patch tokens reshaped to \texttt{[B, 384, 24, 24]} before
the standardised covariance is computed.  Mask precomputation and
loss formulation are identical to \S E.5.  Inference is unchanged.

\begin{table}[h]
\centering
\caption{DINOv3\,+\,ISW hyperparameters; settings matching
Table~\ref{tab:dinov3-config} are omitted.}
\label{tab:dinov3-isw}
\small
\begin{tabular}{@{}ll@{}}
\toprule
\textbf{Hyperparameter} & \textbf{Value} \\
\midrule
$\lambda$ (ISW weight)         & 0.6 \\
Hooked blocks                  & ViT blocks 3, 6, 9 \\
Token reshape                  & Strip CLS; $576\to24{\times}24$ \\
$k$-means $k$ / $m$            & 3 / 1 \\
Photometric aug.\ (mask precomp.)\!\!
  & \texttt{ColorJitter(0.4,0.4,0.4,0.1)} \\
  & + \texttt{GaussianBlur} $p{=}0.5$ \\
Loss reduction                 & Per-image mean, then block mean \\
Decision metric                & Per-image tolerant mIoU ($\pm$2\,px) \\
Early-stopping patience        & 15 epochs \\
\bottomrule
\end{tabular}
\end{table}

\subsection{DINOv3 + MRFP+}

\textbf{DINOv3\,+\,MRFP+.}
HRFP attaches at \texttt{feat\_block3} ($384{\times}24{\times}24$);
NP+ attaches at \texttt{feat\_block11} ($384{\times}24{\times}24$);
HRFP+ projects the max-resolution intermediate HRFP feature
($\sim$$H/8$, 384\,ch) through a frozen $1{\times}1$ conv
($384{\to}64$) and adds it to the decoder's \texttt{up3} output
(64\,ch, $192{\times}192$).  Modules are toggled independently with
$p{=}0.5$.  As in \S E.13, the original-paper InstanceNorm layers are
removed: in DINOv3 the four block features feed a single
\texttt{Conv}-\texttt{BatchNorm2d}-\texttt{ReLU} fusion, and applying
unconditional IN to blocks 6 and 9 during training corrupted that BN's
running statistics, producing decoder checkerboard artefacts and zero
mIoU at evaluation.  HRFP and NP+ alone span both ends of the
perturbation spectrum.  HRFP+HRFP+ projection adds $\approx$99\,K
frozen parameters; inference is identical to base DINOv3.

\begin{table}[h]
\centering
\caption{DINOv3\,+\,MRFP+ hyperparameters; settings matching
Table~\ref{tab:dinov3-config} are omitted.}
\label{tab:dinov3-mrfp}
\small
\begin{tabular}{@{}ll@{}}
\toprule
\textbf{Hyperparameter} & \textbf{Value} \\
\midrule
\multicolumn{2}{@{}l}{\textit{Perturbation modules (training only)}} \\
HRFP attach point        & \texttt{feat\_block3} (384\,ch, $24{\times}24$) \\
NP+ attach point         & \texttt{feat\_block11} (384\,ch, $24{\times}24$) \\
HRFP+ attach point       & Decoder \texttt{up3} (64\,ch, $192{\times}192$) \\
HRFP+ projection         & Frozen $1{\times}1$ conv $384{\to}64$ \\
Inst.\ Norm layers       & Removed \\
HRFP / NP+ / HRFP+ prob.\ & 0.5 / 0.5 / 0.5 \\
HRFP arch.               & 4 enc + 4 dec, $\sim$1.2$\times$ scale per layer \\
HRFP BN std $\sigma$     & 0.5 \\
\midrule
\multicolumn{2}{@{}l}{\textit{Optimisation overrides}} \\
Decision metric          & Per-image tolerant mIoU ($\pm$2\,px) \\
Early-stopping patience  & 15 epochs \\
\bottomrule
\end{tabular}
\end{table}

\subsection{DINOv3 + FADA}

\textbf{DINOv3\,+\,FADA.}
The DINOv3 ViT-S/16 backbone is fully frozen
(\texttt{requires\_grad=False}); ViT uses LayerNorm so no
\texttt{eval}-mode override is needed.  FADA blocks are inserted after
ViT blocks 3, 6, 9, and 11 (all four feature-extraction blocks), each
operating on the spatial patch tokens \texttt{[B, 384, 24, 24]}: 2-D
Haar DWT, LL LoRA-token branch (content), HF LoRA-token branch with
Instance Normalisation on the similarity map (style suppression),
inverse Haar DWT, and re-injection into the token stream.  Gradients
propagate through frozen blocks (parameters not updated) to upstream
adapters.  Trainable parameters: $\approx$3.4\,M
($4{\times}\!\approx$0.48\,M FADA + $\approx$1.5\,M decoder).

\begin{table}[h]
\centering
\caption{DINOv3\,+\,FADA hyperparameters; settings matching
Table~\ref{tab:dinov3-config} are omitted.}
\label{tab:dinov3-fada}
\small
\begin{tabular}{@{}ll@{}}
\toprule
\textbf{Hyperparameter} & \textbf{Value} \\
\midrule
\multicolumn{2}{@{}l}{\textit{FADA configuration}} \\
Adapter blocks           & ViT blocks 3, 6, 9, 11 \\
LoRA rank $r$            & 16 \\
Token length $m$         & 100 \\
LF branch                & LoRA tokens on LL \\
HF branch                & LoRA tokens + IN on similarity map (LH/HL/HH) \\
Decomposition            & Haar DWT \\
\midrule
\multicolumn{2}{@{}l}{\textit{Optimisation overrides}} \\
Optimizer                & Adam (vs.\ AdamW) \\
Learning rate            & $1{\times}10^{-4}$ (vs.\ $5{\times}10^{-5}$) \\
Weight decay             & $1{\times}10^{-4}$ (vs.\ 0.05) \\
LR scheduler             & \texttt{ReduceLROnPlateau}
                           (factor=0.5, patience=3) \\
Backbone state           & Frozen (LayerNorm; no eval-mode override) \\
Decision metric          & Per-image tolerant mIoU ($\pm$2\,px) \\
Early-stopping patience  & 15 epochs \\
\bottomrule
\end{tabular}
\end{table}

\subsection{CrossEarth}
\label{sec:crossearth-base}

\textbf{CrossEarth.}
We train CrossEarth~\cite{gong2025crossearth} from its official release
on our splits, starting from the release's
DINOv2~\cite{oquab2024dinov2} ViT-L/14 backbone initialisation at patch
size 16.  The backbone is frozen; the trainable components are the Rein
adapters~\cite{wei2024rein} (token length 100, LoRA rank 16) and a
Mask2Former decoder~\cite{cheng2022mask2former} with a two-class head.
The code release omits four components described in the CrossEarth paper
(the Geospatial Semantic Extractor, the cross-attention Injector, the
masked-image-modelling branch with its ASPP decoder, and Earth-Style
Injection), so our cells train the released Rein and Mask2Former
configuration without the Million-AID style pool.  No CrossEarth
per-benchmark checkpoint is used.  Table~\ref{tab:crossearth-config}
lists the configuration and our deviations from the release, which
adapt it to our $384{\times}384$ patches, our $450$-image training
sets, the 4-hour partition limit, and source-only validation.

\begin{table}[h]
\centering
\caption{CrossEarth configuration.  Every CrossEarth cell
(Vanilla, FDA, SIB, BSM; LOCO and within-city upper bound) uses these
settings, high resolution only, seed 0.}
\label{tab:crossearth-config}
\small
\begin{tabular}{@{}ll@{}}
\toprule
\textbf{Hyperparameter} & \textbf{Value} \\
\midrule
\multicolumn{2}{@{}l}{\textit{Architecture}} \\
Backbone                 & DINOv2 ViT-L/14 at patch 16, frozen ($303.7$\,M parameters) \\
Adapters                 & Rein, token length 100, LoRA rank 16 \\
Decoder                  & Mask2Former, 2 classes \\
Trainable parameters     & $23.6$\,M \\
\midrule
\multicolumn{2}{@{}l}{\textit{Optimisation}} \\
Optimizer                & AdamW ($\beta_1{=}0.9$, $\beta_2{=}0.999$, $\epsilon{=}10^{-8}$) \\
Learning rate            & $1{\times}10^{-4}$ \\
Weight decay             & 0.05 \\
LR scheduler             & Polynomial (power 0.9, horizon 40{,}000 iterations) \\
Batch size               & 4 (release: 1) \\
Iterations               & 10{,}000 (release: 30{,}000) \\
Crop                     & $384{\times}384$ (release: $512{\times}512$) \\
Precision                & bf16 autocast; deformable attention in fp32 \\
Validation               & Every 500 iterations, source cities only \\
Checkpoint selection     & Best two-class tolerant mIoU ($\pm$2\,px) on source-city validation \\
Inference                & Whole image (release: sliding window) \\
\midrule
\multicolumn{2}{@{}l}{\textit{Infrastructure}} \\
Hardware                 & 1$\times$ NVIDIA A100 40\,GB (NCSA Delta, \texttt{gpuA100x4}) \\
\bottomrule
\end{tabular}
\end{table}

\textbf{CrossEarth\,+\,FDA.}  FDA is applied as for the other
architectures: low-frequency ratio $L{=}0.01$ ($6{\times}6$ swapped
region per channel), a pool of 100 unlabelled images from the held-out
city's \texttt{train/images}, input stage, training only.

\textbf{CrossEarth\,+\,SIB.}  SIB on CrossEarth is the Haar and VIB path
of Appendix~\ref{app:sib-impl}; FDA preprocessing and skip-attention
gates are CNN-only components and are not applied.  One shared-weight
SIB block is inserted before the Mask2Former pixel decoder and applied
to all four backbone feature scales (strides 4, 8, 16, 32), adding
$2.4$\,M trainable parameters.  The VIB weights, hidden widths ($C/2$
and $C/4$ at $C{=}1024$), KL warm-up and coverage criterion follow
Appendix~\ref{app:sib-impl}; $c^*_{val}$ is fitted on source-city
validation per cell.

\subsection{BSM}
\label{sec:bsm}

\textbf{BSM.}
No official BSM code has been released, so we reimplemented it from
the paper~\cite{luo2023bsm}.  At each training step a frozen AdaIN
style-transfer network~\cite{huang2017adain} re-renders every image in
the batch with the style of another image from the same batch (a random
permutation with distinct pairs), and the result is mixed with the
original at a fixed coefficient $\alpha$.  The style source is the
training batch itself, so no external or held-out-city imagery enters
training.  BSM is applied to the RGB input before the network, with the
architecture's input normalisation inverted before mixing and reapplied
after; the contrast channel of MAMNet and OGLANet passes through
unmodified.  BSM is disabled at evaluation.  Each architecture otherwise
keeps its base configuration (Tables~\ref{tab:mamnet-config},
\ref{tab:oglanet-config}, \ref{tab:dinov3-config},
\ref{tab:crossearth-config}).  The DINOv3/Chicago cell collapsed at
seed 0 (best source-city validation mIoU $5.51$, below the training
floor of $10.0$) and is excluded from Table~\ref{tab:loco-main} without
imputation.

\begin{table}[h]
\centering
\caption{BSM hyperparameters; settings matching each architecture's
base configuration are omitted.}
\label{tab:bsm}
\small
\begin{tabular}{@{}ll@{}}
\toprule
\textbf{Hyperparameter} & \textbf{Value} \\
\midrule
Style network            & AdaIN: VGG-19 encoder to \texttt{relu4\_1} + mirror decoder, frozen \\
Style-network weights    & Public pretrained AdaIN release; 0 trainable parameters \\
Style source             & Permuted training batch \\
Mixing coefficient $\alpha$ & 0.5 (fixed; checked for non-degeneracy on source-city validation) \\
Application probability  & 1.0 per training batch \\
Batch size               & 8 (CrossEarth: 4) \\
Early-stopping patience  & 20 epochs (vs.\ 10; CrossEarth: fixed 10{,}000 iterations) \\
Modes run                & LOCO, high resolution only \\
\bottomrule
\end{tabular}
\end{table}

\subsection{SAR}
\label{sec:sar}

\textbf{SAR.}
SAR~\cite{niu2023sar} trains nothing.  It adapts the normalisation-layer
affine parameters of each cell's Vanilla LOCO checkpoint (for MAMNet,
the matched-450 retrain) on each test image's own prediction, by
minimising the entropy of reliable pixels under sharpness-aware
minimisation~\cite{foret2021sam}.  We run it episodically at batch size
1: parameters are restored before every image, so nothing accumulates
across the target set, and no held-out-city label or source image is
used.  Normalisation layers stay in evaluation mode on source running
statistics, because MAMNet and OGLANet return auxiliary and
deep-supervision outputs in training mode.  We keep upstream's skip list
and collect the affine parameters of every remaining BatchNorm,
LayerNorm and GroupNorm layer.  The two-class entropy constants are
upstream's, rescaled from $C{=}1000$ to $C{=}2$ and never tuned on
target mIoU.  When the reliability filter admits no pixel, the step is
skipped.  Each SAR row in Table~\ref{tab:loco-main} is the Vanilla
cell's per-image values plus the change SAR produces within its own
pass (final step minus step 0), so the row isolates the adaptation
effect from any difference between the SAR pass and the stored Vanilla
inference.

\begin{table}[h]
\centering
\caption{SAR hyperparameters.}
\label{tab:sar}
\small
\begin{tabular}{@{}ll@{}}
\toprule
\textbf{Hyperparameter} & \textbf{Value} \\
\midrule
Adapted parameters       & Affine of BatchNorm/LayerNorm/GroupNorm layers \\
Excluded (upstream)      & \texttt{layer4}, ViT blocks 9--11, final \texttt{norm} \\
Normalisation statistics & Source running statistics (evaluation mode) \\
Optimizer                & SAM ($\rho{=}0.05$) over SGD (momentum 0.9) \\
Learning rate            & $2.5{\times}10^{-4}$ (upstream ViT default, not batch-scaled) \\
Steps per image          & 1 (MAMNet, OGLANet, DINOv3); 10 (CrossEarth) \\
Entropy margin $E_0$     & $0.4\ln 2 = 0.277$ \\
Model-recovery threshold & $0.0201$ nats \\
EMA momentum             & 0.9 \\
Batch size               & 1, episodic \\
\midrule
Hardware                 & 1$\times$ NVIDIA A100 40\,GB (NCSA Delta, \texttt{gpuA100x4}) \\
\bottomrule
\end{tabular}
\end{table}

\textbf{Selection.}  For MAMNet, OGLANet and DINOv3, we selected the
learning rate ($\{1, 10, 100\}\times$ the base rate) and step count
($\{1, 10\}$) on source-city validation in a pilot on the Phoenix fold
that pooled DINOv3 and MAMNet.  Upstream additionally scales the learning
rate by batch size over 64, which at batch size 1 gives
$3.9{\times}10^{-6}$; we do not apply that scaling.

\textbf{SAR on CrossEarth.}  CrossEarth exposes no logits, so SAR
minimises the entropy of the logarithm of the normalised two-class
Mask2Former score.  Its recipe (base rate, 10 steps) is borrowed from
the S-EO pilot, which selected on S-EO source-city validation with
MAMNet, OGLANet and DINOv3 and did not include CrossEarth; no internal
CrossEarth cell selected it.  The SAR pass runs in fp32 while the parent
cells were inferred under bf16 autocast.  Before adaptation, the SAR
pass differs from the stored Vanilla value by $+0.0063$ (Chicago),
$+0.1053$ (Miami) and $+0.0070$ (Phoenix) mIoU; most of the Miami
difference comes from one image with 46 foreground pixels.  This
difference does not enter the row, which adds only the within-pass
change: $-0.0011$, $-0.0054$ and $+0.0020$ mIoU.  The run's divergence
guard, which fires when more than $20\%$ of images predict a foreground
fraction below $0.01$, fires on Miami and Phoenix.  The number of such
images is the same before and after adaptation in every city (11, 42 and
53 of 150), so the flag reflects the parent model.

\section{Full Per-Cell LOCO Results}
\label{app:loco-full}
\label{app:loco_extended} %

This appendix reports the full per-cell breakdown of the $158$ LOCO test
cells summarized in Table~\ref{tab:loco-main}. Tables~\ref{tab:loco-full-mamnet}--\ref{tab:loco-full-dinov3}
cover the $126$ cells of MAMNet, OGLANet and DINOv3 under Vanilla and the six training-time methods, and
Table~\ref{tab:loco-full-hr} the $32$ high-resolution SAR, BSM and CrossEarth cells.  Each table reports, for each (architecture, method, held-out city, native resolution)
cell, the within-city upper-bound mIoU, the LOCO mIoU, the per-image mean
$\Delta$ to the upper bound, the $95\%$ paired-bootstrap confidence interval
on $\Delta$, and the two-tailed bootstrap $p$-value
($B = 10{,}000$ resamples on per-image $\Delta$ values). All values are
two-class boundary-tolerant mIoU ($\pm 2$\,px, Appendix~\ref{app:metric}).
$p < .0002$ indicates the bootstrap resolution floor; cells marked $\dagger$
are not significant at $\alpha = 0.05$.
Tables~\ref{tab:loco-per-city}--\ref{tab:loco-method-summary} report pooled
summaries of the $126$ cells by held-out city, native resolution, and method.

\paragraph{Cell provenance and caveats.} The high-resolution MAMNet cells and
the high-resolution OGLANet SegDeSiC cells are $450$-image retrains that match
the training budget of the other high-resolution cells; every mid-resolution
cell is the published run at its published budget.  Each cell is one training
seed.  OGLANet cells are single inference passes at the published batch size;
OGLANet inference is nondeterministic (Appendix~\ref{app:oglanet-dffm}), with
cell means varying by up to $0.42$ mIoU over repeated passes, and the intervals do not include this.  SAR
rows are the Vanilla cell's per-image values plus SAR's within-pass change
(Appendix~\ref{sec:sar}).  BSM DINOv3/Chicago collapsed in training and has no
row.  CrossEarth cells are compared against CrossEarth's own upper bound.  The
intervals in Figure~\ref{fig:teaser} come from the same paired image bootstrap
over the $450$ high-resolution test images per architecture; the four
architectures share bootstrap draws, so overlapping intervals there are not a
test between architectures, and no multiplicity correction is applied.

\subsection{Per-Cell Tables}

\begin{table}[h]
\centering
\footnotesize
\setlength{\tabcolsep}{4pt}
\caption{
\textbf{Full per-cell LOCO results: MAMNet.}
$n = 150$ paired test images per cell. Upper, LOCO in mIoU (\%); $\Delta$ in
mIoU points; CI is the $95\%$ paired-bootstrap interval on $\Delta$.
$\dagger$: not significant at $\alpha = 0.05$.
}
\label{tab:loco-full-mamnet}
\begin{tabular}{@{}llcrrrcr@{}}
\toprule
\textbf{Method} & \textbf{City} & \textbf{Res.} & \textbf{Upper} & \textbf{LOCO} & $\Delta$ & \textbf{95\% CI} & $p$ \\
\midrule
Vanilla   & Chicago & HR & 86.31 & 80.45 & $-5.87$  & $[-6.89,\,-4.88]$   & $<\!.0002$ \\
          & Chicago & MR & 86.35 & 83.22 & $-3.13$  & $[-3.96,\,-2.39]$   & $<\!.0002$ \\
          & Miami   & HR & 79.11 & 74.67 & $-4.44$  & $[-6.04,\,-2.93]$   & $<\!.0002$ \\
          & Miami   & MR & 78.89 & 76.75 & $-2.14$  & $[-3.23,\,-1.13]$   & $<\!.0002$ \\
          & Phoenix & HR & 73.75 & 64.23 & $-9.52$  & $[-11.50,\,-7.53]$  & $<\!.0002$ \\
          & Phoenix & MR & 85.64 & 83.40 & $-2.24$  & $[-3.50,\,-1.07]$   & $<\!.0002$ \\
\cmidrule(l){1-8}
FDA       & Chicago & HR & 86.31 & 79.77 & $-6.54$  & $[-7.82,\,-5.31]$   & $<\!.0002$ \\
          & Chicago & MR & 86.35 & 82.88 & $-3.47$  & $[-4.19,\,-2.75]$   & $<\!.0002$ \\
          & Miami   & HR & 79.11 & 73.10 & $-6.01$  & $[-7.73,\,-4.35]$   & $<\!.0002$ \\
          & Miami   & MR & 78.89 & 75.71 & $-3.18$  & $[-4.31,\,-2.06]$   & $<\!.0002$ \\
          & Phoenix & HR & 73.75 & 67.89 & $-5.86$  & $[-7.61,\,-4.16]$   & $<\!.0002$ \\
          & Phoenix & MR & 85.64 & 82.64 & $-2.99$  & $[-4.45,\,-1.63]$   & $<\!.0002$ \\
\cmidrule(l){1-8}
SegDeSiC  & Chicago & HR & 86.31 & 82.01 & $-4.30$  & $[-5.19,\,-3.40]$   & $<\!.0002$ \\
          & Chicago & MR & 86.35 & 81.50 & $-4.85$  & $[-5.64,\,-4.08]$   & $<\!.0002$ \\
          & Miami   & HR & 79.11 & 70.15 & $-8.96$  & $[-11.88,\,-6.26]$  & $<\!.0002$ \\
          & Miami   & MR & 78.89 & 75.99 & $-2.90$  & $[-4.12,\,-1.79]$   & $<\!.0002$ \\
          & Phoenix & HR & 73.75 & 63.54 & $-10.21$ & $[-12.33,\,-8.12]$  & $<\!.0002$ \\
          & Phoenix & MR & 85.64 & 83.49 & $-2.15$  & $[-3.22,\,-1.12]$   & $<\!.0002$ \\
\cmidrule(l){1-8}
IIM       & Chicago & HR & 86.31 & 79.25 & $-7.06$  & $[-8.15,\,-6.00]$   & $<\!.0002$ \\
          & Chicago & MR & 86.35 & 82.31 & $-4.04$  & $[-4.91,\,-3.23]$   & $<\!.0002$ \\
          & Miami   & HR & 79.11 & 67.95 & $-11.16$ & $[-13.33,\,-9.14]$  & $<\!.0002$ \\
          & Miami   & MR & 78.89 & 74.52 & $-4.37$  & $[-5.74,\,-3.12]$   & $<\!.0002$ \\
          & Phoenix & HR & 73.75 & 49.69 & $-24.06$ & $[-26.70,\,-21.37]$ & $<\!.0002$ \\
          & Phoenix & MR & 85.64 & 77.85 & $-7.78$  & $[-9.87,\,-5.80]$   & $<\!.0002$ \\
\cmidrule(l){1-8}
MRFP+     & Chicago & HR & 86.31 & 74.81 & $-11.50$ & $[-12.93,\,-10.10]$ & $<\!.0002$ \\
          & Chicago & MR & 86.35 & 81.29 & $-5.06$  & $[-5.91,\,-4.26]$   & $<\!.0002$ \\
          & Miami   & HR & 79.11 & 70.87 & $-8.24$  & $[-10.87,\,-5.73]$  & $<\!.0002$ \\
          & Miami   & MR & 78.89 & 67.09 & $-11.80$ & $[-13.84,\,-9.85]$  & $<\!.0002$ \\
          & Phoenix & HR & 73.75 & 55.16 & $-18.59$ & $[-21.14,\,-16.01]$ & $<\!.0002$ \\
          & Phoenix & MR & 85.64 & 78.94 & $-6.70$  & $[-8.49,\,-5.00]$   & $<\!.0002$ \\
\cmidrule(l){1-8}
FADA      & Chicago & HR & 86.31 & 77.89 & $-8.42$  & $[-9.50,\,-7.36]$   & $<\!.0002$ \\
          & Chicago & MR & 86.35 & 80.03 & $-6.32$  & $[-7.40,\,-5.33]$   & $<\!.0002$ \\
          & Miami   & HR & 79.11 & 65.27 & $-13.84$ & $[-16.62,\,-11.24]$ & $<\!.0002$ \\
          & Miami   & MR & 78.89 & 72.35 & $-6.54$  & $[-7.88,\,-5.24]$   & $<\!.0002$ \\
          & Phoenix & HR & 73.75 & 61.79 & $-11.96$ & $[-13.97,\,-10.01]$ & $<\!.0002$ \\
          & Phoenix & MR & 85.64 & 76.41 & $-9.23$  & $[-10.94,\,-7.55]$  & $<\!.0002$ \\
\cmidrule(l){1-8}
ISW       & Chicago & HR & 86.31 & 74.65 & $-11.66$ & $[-12.65,\,-10.65]$ & $<\!.0002$ \\
          & Chicago & MR & 86.35 & 71.26 & $-15.09$ & $[-16.98,\,-13.26]$ & $<\!.0002$ \\
          & Miami   & HR & 79.11 & 68.04 & $-11.07$ & $[-13.37,\,-8.93]$  & $<\!.0002$ \\
          & Miami   & MR & 78.89 & 69.34 & $-9.55$  & $[-11.40,\,-7.84]$  & $<\!.0002$ \\
          & Phoenix & HR & 73.75 & 50.47 & $-23.28$ & $[-25.81,\,-20.66]$ & $<\!.0002$ \\
          & Phoenix & MR & 85.64 & 76.04 & $-9.59$  & $[-11.22,\,-8.05]$  & $<\!.0002$ \\
\bottomrule
\end{tabular}
\end{table}

\begin{table}[h]
\centering
\footnotesize
\setlength{\tabcolsep}{4pt}
\caption{
\textbf{Full per-cell LOCO results: OGLANet.}
Conventions as in Table~\ref{tab:loco-full-mamnet}.
}
\label{tab:loco-full-oglanet}
\begin{tabular}{@{}llcrrrcr@{}}
\toprule
\textbf{Method} & \textbf{City} & \textbf{Res.} & \textbf{Upper} & \textbf{LOCO} & $\Delta$ & \textbf{95\% CI} & $p$ \\
\midrule
Vanilla   & Chicago & HR & 79.12 & 68.47 & $-10.66$ & $[-12.08,\,-9.21]$  & $<\!.0002$ \\
          & Chicago & MR & 80.77 & 65.08 & $-15.70$ & $[-17.66,\,-13.81]$ & $<\!.0002$ \\
          & Miami   & HR & 73.89 & 66.18 & $-7.71$  & $[-10.20,\,-5.35]$  & $<\!.0002$ \\
          & Miami   & MR & 74.19 & 70.73 & $-3.45$  & $[-4.76,\,-2.24]$   & $<\!.0002$ \\
          & Phoenix & HR & 66.15 & 61.18 & $-4.98$  & $[-6.23,\,-3.72]$   & $<\!.0002$ \\
          & Phoenix & MR & 75.39 & 70.32 & $-5.07$  & $[-6.42,\,-3.76]$   & $<\!.0002$ \\
\cmidrule(l){1-8}
FDA       & Chicago & HR & 79.12 & 71.07 & $-8.05$  & $[-9.10,\,-6.97]$   & $<\!.0002$ \\
          & Chicago & MR & 80.77 & 60.19 & $-20.58$ & $[-22.54,\,-18.63]$ & $<\!.0002$ \\
          & Miami   & HR & 73.89 & 67.30 & $-6.59$  & $[-8.34,\,-4.89]$   & $<\!.0002$ \\
          & Miami   & MR & 74.19 & 69.82 & $-4.36$  & $[-5.63,\,-3.17]$   & $<\!.0002$ \\
          & Phoenix & HR & 66.15 & 63.27 & $-2.88$  & $[-4.07,\,-1.69]$   & $<\!.0002$ \\
          & Phoenix & MR & 75.39 & 70.09 & $-5.30$  & $[-6.65,\,-4.01]$   & $<\!.0002$ \\
\cmidrule(l){1-8}
SegDeSiC  & Chicago & HR & 79.12 & 68.99 & $-10.13$ & $[-11.72,\,-8.52]$  & $<\!.0002$ \\
          & Chicago & MR & 80.77 & 59.40 & $-21.37$ & $[-23.64,\,-19.13]$ & $<\!.0002$ \\
          & Miami   & HR & 73.89 & 60.69 & $-13.20$ & $[-16.02,\,-10.49]$ & $<\!.0002$ \\
          & Miami   & MR & 74.19 & 69.91 & $-4.28$  & $[-5.62,\,-3.03]$   & $<\!.0002$ \\
          & Phoenix & HR & 66.15 & 60.77 & $-5.38$  & $[-6.72,\,-4.09]$   & $<\!.0002$ \\
          & Phoenix & MR & 75.39 & 70.27 & $-5.12$  & $[-6.59,\,-3.68]$   & $<\!.0002$ \\
\cmidrule(l){1-8}
IIM       & Chicago & HR & 79.12 & 70.01 & $-9.11$  & $[-10.43,\,-7.79]$  & $<\!.0002$ \\
          & Chicago & MR & 80.77 & 62.21 & $-18.56$ & $[-20.84,\,-16.34]$ & $<\!.0002$ \\
          & Miami   & HR & 73.89 & 66.36 & $-7.53$  & $[-9.50,\,-5.61]$   & $<\!.0002$ \\
          & Miami   & MR & 74.19 & 69.13 & $-5.06$  & $[-6.50,\,-3.76]$   & $<\!.0002$ \\
          & Phoenix & HR & 66.15 & 55.25 & $-10.90$ & $[-12.30,\,-9.50]$  & $<\!.0002$ \\
          & Phoenix & MR & 75.39 & 70.28 & $-5.11$  & $[-6.43,\,-3.86]$   & $<\!.0002$ \\
\cmidrule(l){1-8}
MRFP+     & Chicago & HR & 79.12 & 63.82 & $-15.30$ & $[-16.93,\,-13.68]$ & $<\!.0002$ \\
          & Chicago & MR & 80.77 & 52.27 & $-28.50$ & $[-30.86,\,-26.14]$ & $<\!.0002$ \\
          & Miami   & HR & 73.89 & 57.30 & $-16.59$ & $[-19.17,\,-13.99]$ & $<\!.0002$ \\
          & Miami   & MR & 74.19 & 66.64 & $-7.54$  & $[-9.09,\,-6.08]$   & $<\!.0002$ \\
          & Phoenix & HR & 66.15 & 58.03 & $-8.13$  & $[-9.63,\,-6.59]$   & $<\!.0002$ \\
          & Phoenix & MR & 75.39 & 67.26 & $-8.14$  & $[-9.65,\,-6.69]$   & $<\!.0002$ \\
\cmidrule(l){1-8}
FADA      & Chicago & HR & 79.12 & 66.17 & $-12.96$ & $[-14.49,\,-11.43]$ & $<\!.0002$ \\
          & Chicago & MR & 80.77 & 53.43 & $-27.35$ & $[-29.43,\,-25.30]$ & $<\!.0002$ \\
          & Miami   & HR & 73.89 & 65.20 & $-8.69$  & $[-11.19,\,-6.19]$  & $<\!.0002$ \\
          & Miami   & MR & 74.19 & 69.17 & $-5.02$  & $[-6.43,\,-3.72]$   & $<\!.0002$ \\
          & Phoenix & HR & 66.15 & 60.26 & $-5.89$  & $[-7.28,\,-4.47]$   & $<\!.0002$ \\
          & Phoenix & MR & 75.39 & 68.71 & $-6.69$  & $[-8.06,\,-5.40]$   & $<\!.0002$ \\
\cmidrule(l){1-8}
ISW       & Chicago & HR & 79.12 & 64.19 & $-14.93$ & $[-16.27,\,-13.56]$ & $<\!.0002$ \\
          & Chicago & MR & 80.77 & 59.74 & $-21.03$ & $[-23.13,\,-18.95]$ & $<\!.0002$ \\
          & Miami   & HR & 73.89 & 67.78 & $-6.11$  & $[-7.66,\,-4.57]$   & $<\!.0002$ \\
          & Miami   & MR & 74.19 & 64.81 & $-9.38$  & $[-10.96,\,-7.86]$  & $<\!.0002$ \\
          & Phoenix & HR & 66.15 & 59.69 & $-6.46$  & $[-8.03,\,-4.91]$   & $<\!.0002$ \\
          & Phoenix & MR & 75.39 & 66.98 & $-8.42$  & $[-9.56,\,-7.29]$   & $<\!.0002$ \\
\bottomrule
\end{tabular}
\end{table}

\begin{table}[h]
\centering
\footnotesize
\setlength{\tabcolsep}{4pt}
\caption{
\textbf{Full per-cell LOCO results: DINOv3.}
Conventions as in Table~\ref{tab:loco-full-mamnet}.
}
\label{tab:loco-full-dinov3}
\begin{tabular}{@{}llcrrrcr@{}}
\toprule
\textbf{Method} & \textbf{City} & \textbf{Res.} & \textbf{Upper} & \textbf{LOCO} & $\Delta$ & \textbf{95\% CI} & $p$ \\
\midrule
Vanilla   & Chicago & HR & 80.27 & 67.26 & $-13.01$ & $[-14.37,\,-11.67]$ & $<\!.0002$ \\
          & Chicago & MR & 83.63 & 75.23 & $-8.40$  & $[-9.71,\,-7.17]$   & $<\!.0002$ \\
          & Miami   & HR & 76.00 & 72.61 & $-3.39$  & $[-5.36,\,-1.67]$   & $<\!.0002$ \\
          & Miami   & MR & 75.76 & 68.11 & $-7.65$  & $[-9.24,\,-6.16]$   & $<\!.0002$ \\
          & Phoenix & HR & 67.82 & 65.24 & $-2.58$  & $[-4.30,\,-0.88]$   & $.0028$ \\
          & Phoenix & MR & 81.85 & 71.52 & $-10.33$ & $[-12.00,\,-8.61]$  & $<\!.0002$ \\
\cmidrule(l){1-8}
FDA       & Chicago & HR & 80.27 & 72.26 & $-8.01$  & $[-9.15,\,-6.92]$   & $<\!.0002$ \\
          & Chicago & MR & 83.63 & 75.70 & $-7.94$  & $[-8.85,\,-7.04]$   & $<\!.0002$ \\
          & Miami   & HR & 76.00 & 70.81 & $-5.19$  & $[-6.30,\,-4.04]$   & $<\!.0002$ \\
          & Miami   & MR & 75.76 & 71.95 & $-3.81$  & $[-5.15,\,-2.62]$   & $<\!.0002$ \\
          & Phoenix & HR & 67.82 & 58.72 & $-9.10$  & $[-11.21,\,-6.97]$  & $<\!.0002$ \\
          & Phoenix & MR & 81.85 & 67.01 & $-14.84$ & $[-16.56,\,-13.13]$ & $<\!.0002$ \\
\cmidrule(l){1-8}
SegDeSiC  & Chicago & HR & 80.27 & 75.52 & $-4.75$  & $[-6.10,\,-3.35]$   & $<\!.0002$ \\
          & Chicago & MR & 83.63 & 77.30 & $-6.33$  & $[-7.44,\,-5.32]$   & $<\!.0002$ \\
          & Miami   & HR & 76.00 & 71.18 & $-4.82$  & $[-6.76,\,-3.10]$   & $<\!.0002$ \\
          & Miami   & MR & 75.76 & 70.72 & $-5.04$  & $[-6.60,\,-3.63]$   & $<\!.0002$ \\
          & Phoenix & HR & 67.82 & 68.04 & $+0.22$  & $[-1.22,\,+1.66]$   & $.7644^{\dagger}$ \\
          & Phoenix & MR & 81.85 & 75.13 & $-6.72$  & $[-8.10,\,-5.37]$   & $<\!.0002$ \\
\cmidrule(l){1-8}
IIM       & Chicago & HR & 80.27 & 77.43 & $-2.85$  & $[-4.04,\,-1.69]$   & $<\!.0002$ \\
          & Chicago & MR & 83.63 & 73.58 & $-10.05$ & $[-11.68,\,-8.54]$  & $<\!.0002$ \\
          & Miami   & HR & 76.00 & 72.23 & $-3.78$  & $[-5.75,\,-2.06]$   & $<\!.0002$ \\
          & Miami   & MR & 75.76 & 69.42 & $-6.34$  & $[-7.98,\,-4.79]$   & $<\!.0002$ \\
          & Phoenix & HR & 67.82 & 56.51 & $-11.31$ & $[-13.57,\,-9.01]$  & $<\!.0002$ \\
          & Phoenix & MR & 81.85 & 79.52 & $-2.33$  & $[-3.79,\,-0.86]$   & $.0034$ \\
\cmidrule(l){1-8}
MRFP+     & Chicago & HR & 80.27 & 77.46 & $-2.81$  & $[-3.67,\,-1.94]$   & $<\!.0002$ \\
          & Chicago & MR & 83.63 & 78.92 & $-4.71$  & $[-5.72,\,-3.78]$   & $<\!.0002$ \\
          & Miami   & HR & 76.00 & 73.20 & $-2.80$  & $[-4.13,\,-1.52]$   & $<\!.0002$ \\
          & Miami   & MR & 75.76 & 72.21 & $-3.55$  & $[-4.98,\,-2.17]$   & $<\!.0002$ \\
          & Phoenix & HR & 67.82 & 60.78 & $-7.04$  & $[-9.30,\,-4.72]$   & $<\!.0002$ \\
          & Phoenix & MR & 81.85 & 73.47 & $-8.37$  & $[-9.99,\,-6.84]$   & $<\!.0002$ \\
\cmidrule(l){1-8}
FADA      & Chicago & HR & 80.27 & 70.15 & $-10.12$ & $[-11.56,\,-8.62]$  & $<\!.0002$ \\
          & Chicago & MR & 83.63 & 63.10 & $-20.54$ & $[-22.32,\,-18.83]$ & $<\!.0002$ \\
          & Miami   & HR & 76.00 & 71.43 & $-4.57$  & $[-5.64,\,-3.59]$   & $<\!.0002$ \\
          & Miami   & MR & 75.76 & 66.73 & $-9.03$  & $[-10.80,\,-7.36]$  & $<\!.0002$ \\
          & Phoenix & HR & 67.82 & 62.81 & $-5.01$  & $[-6.66,\,-3.39]$   & $<\!.0002$ \\
          & Phoenix & MR & 81.85 & 67.56 & $-14.29$ & $[-16.02,\,-12.60]$ & $<\!.0002$ \\
\cmidrule(l){1-8}
ISW       & Chicago & HR & 80.27 & 57.60 & $-22.68$ & $[-24.35,\,-20.92]$ & $<\!.0002$ \\
          & Chicago & MR & 83.63 & 61.66 & $-21.98$ & $[-23.39,\,-20.55]$ & $<\!.0002$ \\
          & Miami   & HR & 76.00 & 59.86 & $-16.15$ & $[-18.16,\,-14.19]$ & $<\!.0002$ \\
          & Miami   & MR & 75.76 & 58.01 & $-17.75$ & $[-19.91,\,-15.59]$ & $<\!.0002$ \\
          & Phoenix & HR & 67.82 & 55.68 & $-12.14$ & $[-13.87,\,-10.43]$ & $<\!.0002$ \\
          & Phoenix & MR & 81.85 & 61.41 & $-20.44$ & $[-21.91,\,-18.90]$ & $<\!.0002$ \\
\bottomrule
\end{tabular}
\end{table}

\clearpage

\begin{table}[h]
\centering
\footnotesize
\setlength{\tabcolsep}{4pt}
\caption{
\textbf{Full per-cell LOCO results: SAR, BSM and CrossEarth} (high resolution).
Conventions as in Table~\ref{tab:loco-full-mamnet}.  CrossEarth cells are
measured against CrossEarth's own upper bound.
}
\label{tab:loco-full-hr}
\begin{tabular}{@{}llcrrrcr@{}}
\toprule
\textbf{Arch.} & \textbf{Method} & \textbf{City} & \textbf{Upper} & \textbf{LOCO} & $\Delta$ & \textbf{95\% CI} & $p$ \\
\midrule
MAMNet & SAR & Chicago & 86.31 & 80.45 & $-5.87$ & $[-6.89,\,-4.88]$ & $<\!.0002$ \\
 &  & Miami & 79.11 & 74.66 & $-4.45$ & $[-6.05,\,-2.93]$ & $<\!.0002$ \\
 &  & Phoenix & 73.75 & 64.22 & $-9.53$ & $[-11.51,\,-7.54]$ & $<\!.0002$ \\
\cmidrule(l){1-8}
OGLANet & SAR & Chicago & 79.12 & 68.70 & $-10.42$ & $[-11.99,\,-8.79]$ & $<\!.0002$ \\
 &  & Miami & 73.89 & 66.36 & $-7.53$ & $[-10.00,\,-5.13]$ & $<\!.0002$ \\
 &  & Phoenix & 66.15 & 61.18 & $-4.98$ & $[-6.28,\,-3.66]$ & $<\!.0002$ \\
\cmidrule(l){1-8}
DINOv3 & SAR & Chicago & 80.27 & 67.25 & $-13.02$ & $[-14.38,\,-11.67]$ & $<\!.0002$ \\
 &  & Miami & 76.00 & 72.61 & $-3.39$ & $[-5.36,\,-1.67]$ & $<\!.0002$ \\
 &  & Phoenix & 67.82 & 65.24 & $-2.58$ & $[-4.30,\,-0.88]$ & $.0028$ \\
\cmidrule(l){1-8}
MAMNet & BSM & Chicago & 86.31 & 78.09 & $-8.22$ & $[-9.37,\,-7.08]$ & $<\!.0002$ \\
 &  & Miami & 79.11 & 72.59 & $-6.52$ & $[-8.21,\,-4.92]$ & $<\!.0002$ \\
 &  & Phoenix & 73.75 & 67.70 & $-6.05$ & $[-7.74,\,-4.45]$ & $<\!.0002$ \\
\cmidrule(l){1-8}
OGLANet & BSM & Chicago & 79.12 & 61.73 & $-17.40$ & $[-19.54,\,-15.30]$ & $<\!.0002$ \\
 &  & Miami & 73.89 & 60.65 & $-13.24$ & $[-15.85,\,-10.70]$ & $<\!.0002$ \\
 &  & Phoenix & 66.15 & 64.58 & $-1.57$ & $[-2.92,\,-0.29]$ & $.0156$ \\
\cmidrule(l){1-8}
DINOv3 & BSM & Miami & 76.00 & 69.74 & $-6.27$ & $[-8.52,\,-4.29]$ & $<\!.0002$ \\
 &  & Phoenix & 67.82 & 62.66 & $-5.16$ & $[-6.94,\,-3.36]$ & $<\!.0002$ \\
\cmidrule(l){1-8}
CrossEarth & Vanilla & Chicago & 85.01 & 82.66 & $-2.35$ & $[-3.11,\,-1.57]$ & $<\!.0002$ \\
 &  & Miami & 78.48 & 72.61 & $-5.87$ & $[-7.20,\,-4.59]$ & $<\!.0002$ \\
 &  & Phoenix & 72.89 & 70.55 & $-2.34$ & $[-3.53,\,-1.09]$ & $<\!.0002$ \\
\cmidrule(l){1-8}
CrossEarth & FDA & Chicago & 85.01 & 82.55 & $-2.46$ & $[-3.44,\,-1.41]$ & $.0002$ \\
 &  & Miami & 78.48 & 71.49 & $-6.99$ & $[-8.51,\,-5.55]$ & $<\!.0002$ \\
 &  & Phoenix & 72.89 & 71.35 & $-1.53$ & $[-2.60,\,-0.44]$ & $.0068$ \\
\cmidrule(l){1-8}
CrossEarth & SIB & Chicago & 85.01 & 79.31 & $-5.70$ & $[-6.54,\,-4.88]$ & $<\!.0002$ \\
 &  & Miami & 78.48 & 73.93 & $-4.55$ & $[-5.72,\,-3.48]$ & $<\!.0002$ \\
 &  & Phoenix & 72.89 & 69.80 & $-3.08$ & $[-4.26,\,-1.90]$ & $<\!.0002$ \\
\cmidrule(l){1-8}
CrossEarth & SAR & Chicago & 85.01 & 82.66 & $-2.35$ & $[-3.11,\,-1.57]$ & $<\!.0002$ \\
 &  & Miami & 78.48 & 72.61 & $-5.88$ & $[-7.21,\,-4.60]$ & $<\!.0002$ \\
 &  & Phoenix & 72.89 & 70.55 & $-2.34$ & $[-3.53,\,-1.09]$ & $<\!.0002$ \\
\cmidrule(l){1-8}
CrossEarth & BSM & Chicago & 85.01 & 81.42 & $-3.59$ & $[-4.41,\,-2.81]$ & $<\!.0002$ \\
 &  & Miami & 78.48 & 74.76 & $-3.73$ & $[-4.67,\,-2.79]$ & $<\!.0002$ \\
 &  & Phoenix & 72.89 & 71.06 & $-1.83$ & $[-3.21,\,-0.40]$ & $.0108$ \\
\bottomrule
\end{tabular}
\end{table}

\subsection{Pooled Summaries}

\begin{table}[h]
\centering
\small
\caption{
\textbf{LOCO degradation by held-out city,} pooled across architectures and
methods ($n = 42$ cells per city).
}
\label{tab:loco-per-city}
\begin{tabular}{lcccc}
\toprule
\textbf{Held-out city} & $\Delta_{\text{mean}}$ & $\Delta_{\text{best}}$ & $\Delta_{\text{worst}}$ & \textit{Sig./42} \\
\midrule
Chicago & $-11.33$ & $-2.81$  & $-28.50$ & $42/42$ \\
Miami   & $-7.23$ & $-2.14$  & $-17.75$ & $42/42$ \\
Phoenix & $-8.50$ & $+0.22$  & $-24.06$ & $41/42$ \\
\bottomrule
\end{tabular}
\end{table}

\begin{table}[h]
\centering
\small
\caption{
\textbf{LOCO degradation by native resolution,} pooled across architectures,
methods, and held-out cities ($n = 63$ cells per resolution).
}
\label{tab:loco-per-resolution}
\begin{tabular}{lcccc}
\toprule
\textbf{Resolution} & $\Delta_{\text{mean}}$ & $\Delta_{\text{best}}$ & $\Delta_{\text{worst}}$ & \textit{Sig./63} \\
\midrule
High-res ($0.3$\,m/px) & $-8.99$ & $+0.22$ & $-24.06$ & $62/63$ \\
Mid-res  ($0.6$\,m/px) & $-9.04$ & $-2.14$ & $-28.50$ & $63/63$ \\
\bottomrule
\end{tabular}
\end{table}

\begin{table}[h]
\centering
\small
\caption{
\textbf{Method-level safety summary,} pooled across architectures, cities,
and resolutions ($n = 18$ cells per method). Vanilla (no adaptation) is the
no-intervention reference. The worst cell is MRFP+ on OGLANet/Chicago at mid
resolution ($-28.50$) and the best is SegDeSiC on DINOv3/Phoenix at high
resolution ($+0.22$, not significant).
}
\label{tab:loco-method-summary}
\begin{tabular}{lcccc}
\toprule
\textbf{Method} & \textit{Sig./18} & $\Delta_{\text{mean}}$ & $\Delta_{\text{best}}$ & $\Delta_{\text{worst}}$ \\
\midrule
Vanilla   & $18$ & $-6.68$  & $-2.14$ & $-15.70$ \\
FDA       & $18$ & $-6.93$  & $-2.88$ & $-20.58$ \\
SegDeSiC  & $17$ & $-6.68$  & $+0.22$ & $-21.37$ \\
IIM       & $18$ & $-8.41$  & $-2.33$ & $-24.06$ \\
MRFP+     & $18$ & $-9.74$  & $-2.80$ & $-28.50$ \\
FADA      & $18$ & $-10.36$ & $-4.57$ & $-27.35$ \\
ISW       & $18$ & $-14.32$ & $-6.11$ & $-23.28$ \\
\bottomrule
\end{tabular}
\end{table}

\subsection{Histogram Matching Intervention}
\label{app:pixel-intervention}

This appendix expands on the pixel-space test reported in Section~\ref{sec:feature-space}, which rules out input-preprocessing as a recovery strategy under LOCO.

\textbf{Procedure.} For each (architecture, holdout city) cell under vanilla LOCO, we compute the aggregate intensity histogram (the mean of the three channels) of the source-city training images. We map each test image's intensity CDF onto this histogram, restore each pixel's ratio between its channels and its intensity, and run the trained model on the transformed image with no parameter changes. The intervention is parameter-free, applied at test time, and isolates the pixel-statistics hypothesis: if LOCO failure lives at the pixel level, aligning the marginal pixel distribution should recover at least part of the within-city upper bound.

\textbf{Recovery and inflation.} Section~\ref{sec:feature-space} reports the per-architecture median recovery ratio $R$; all three medians lie in $[-2.72, -1.55]$ with $95\%$ CIs below zero. At the median, matched images lie $2.5$ to $3.7$ times as far from the upper bound as the LOCO predictions. False-positive pixel counts rise on $5$ of the $9$ cells and fall on the other $4$. The largest rise is on MAMNet/Chicago, from $201$K LOCO false positives to $15.4$M after matching ($76\times$). Counts are pooled over the $150$ test images with the boundary-tolerant band of the original scorer. $R$ excludes images whose upper-minus-LOCO IoU is below $0.02$ or undefined, and OGLANet values carry the nondeterminism of Appendix~\ref{app:oglanet-dffm}.

\textbf{Implication.} The destruction is uniform and large: $0$ of $9$ cells improve, and the inflation pattern indicates the model's features are tied to the joint distribution of intensity, texture, and spatial pattern in source-city imagery. Marginal pixel-distribution alignment breaks this coupling without replacing it. Any recovery strategy that operates only on input statistics (gamma correction, contrast normalization, single-channel histogram standardization) faces the same structural problem.

\subsection{Selective-Prediction Gap: Per-Cell Details}
\label{app:sp-details}

This appendix expands the SP-gap analysis of Section~\ref{sec:sp-gap}, which characterizes the LOCO failure as class-asymmetric foreground-confidence collapse.

\textbf{Bootstrap procedure.} For each of the $63$ LOCO cells (architecture $\times$ method $\times$ holdout city, high-resolution only), we compute per-image risk-coverage curves on three pixel populations: foreground (\texttt{gt\_pos}, confidence $= p$), background (\texttt{gt\_bg}, confidence $= 1-p$), and all pixels (confidence $= \max(p, 1-p)$). The SP gap is $\text{AURC}_{\text{LOCO}} - \text{AURC}_{\text{upper}}$; image-level paired bootstrap ($B = 1{,}000$) gives $95\%$ CIs on the gap. Cells are paired by image because both upper and LOCO models evaluate on the same $150$ test images per cell.

\textbf{Class-asymmetry distribution.} The pooled means reported in Section~\ref{sec:sp-gap} (foreground gap $+0.136$, background gap $+0.006$) reflect a stable per-cell pattern: the foreground gap exceeds the background gap in $55$ of $63$ cells (sign-test $p = 4.9 \times 10^{-10}$). The per-method breakdown shown in Figure~\ref{fig:diagnostics}(b) confirms the asymmetry holds for every adaptation method individually; no method produces a foreground gap smaller than its background gap on average.

\textbf{Logit and ECE corroboration.} Two independent per-pixel measurements track the same asymmetry. At the logit level, true-positive logits drop by $0.85$ nats on average while false-positive logits drop by $0.15$ nats; the asymmetry holds in $55$ of $63$ cells. Expected calibration error rises by $0.064$ on predicted-positive pixels and is unchanged ($-0.001$) on predicted-negative pixels; the rise is larger on the predicted-positive side in $44$ of $54$ cells (nine cells excluded for insufficient predicted-positive coverage on the calibration split). The three measurements (AURC gap, logit drift, ECE rise) are independent in the sense that they depend on different functions of the prediction distribution, and all three identify the foreground-class population as the locus of degradation.

\subsection{Cross-Entropy AURC Under-Reporting}
\label{app:zeroone-ce}

This appendix details the methodological finding that standard $0/1$ evaluation under-reports the LOCO calibration failure by $2.3\times$, motivating the cross-entropy framing in Section~\ref{sec:sp-gap}.

\textbf{Why $0/1$-AURC misses the calibration component.} The standard AURC measures per-pixel $0/1$ error: a pixel is correct if its predicted class matches the ground truth, regardless of the predicted probability. Two models that produce identical rank orderings of pixels by confidence yield identical $0/1$ risk-coverage curves, even if one assigns probability $0.99$ to a correct pixel and the other assigns $0.55$. The $0/1$-AURC is invariant to any monotonic rescaling of the probability outputs.

Cross-entropy AURC replaces $0/1$ error with per-pixel cross-entropy: $-\log p$ on foreground pixels and $-\log(1-p)$ on background pixels. CE-AURC depends on the probability values themselves, not just rank order, and therefore detects calibration collapse that leaves the rank ordering intact.

\textbf{Result.} The pooled CE-AURC gap exceeds the pooled $0/1$-AURC gap by a factor of $2.3\times$ on the foreground population ($+0.315$ against $+0.136$). The cell-level direction is consistent: $60$ of $63$ cells have $\text{gap}_{\text{CE}} > \text{gap}_{0/1}$ (one-sided paired Wilcoxon $p = 2.3 \times 10^{-11}$).

\textbf{Implication.} Practitioners who evaluate selective deployment on $0/1$ metrics alone observe less than half of the calibration degradation that cross-entropy scoring measures; the rest is hidden by the rank-invariance of $0/1$ scoring. The under-reporting is not an artifact of the bootstrap or the bin width: it is a structural property of the metric.

\subsection{Coverage-Threshold Recovery and Calibration}
\label{app:sp-recovery}

This appendix details the operating-point recovery analysis and the failure of standard post-hoc calibration that together motivate the deployment-time component of \sib{} (Section~\ref{sec:sib-design}).

\textbf{Recoverable-threshold definition.} For each cell we define $c^* = \max\{c : \text{err}_{\text{LOCO}}^{\text{sel}}(c) \leq \text{err}_{\text{upper}}(c{=}1.0) + 0.005\}$, the largest coverage at which the LOCO selective error on the foreground population matches the within-city full-coverage error within a $0.5\%$ tolerance. Image-level paired bootstrap on the per-image error stack gives a $95\%$ CI on $c^*$. A finite $c^*$ exists in $59$ of $63$ cells. The four exceptions are all on Phoenix under aggressive adaptation methods (IIM, ISW, MRFP$^+$); on these cells, no selective-prediction threshold recovers within-city quality.

\textbf{Required threshold magnitude.} Among the $59$ recoverable cells, the median $c^*$ corresponds to discarding $25\%$ of foreground pixels. On Phoenix the median is $40\%$.

\textbf{Method-ranking stability.} We rank the seven methods (Vanilla and six adaptation methods) on the per-method mean foreground SP gap, separately for each of the three architectures, and compute Kendall's coefficient of concordance across the three rankings: $W = 0.79$ ($p = 0.027$, chi-square approximation over $7$ methods). FDA has the smallest pooled gap ($+0.05$) and ISW the largest ($+0.22$), and the pooled ordering is FDA $<$ Vanilla $<$ FADA $<$ SegDeSiC $<$ IIM $<$ MRFP$^+$ $<$ ISW. FDA ranks first on all three architectures, and ISW and MRFP$^+$ each rank between fifth and seventh. On MAMNet, IIM, ISW and MRFP$^+$ reach $+0.22$, $+0.23$ and $+0.25$.

\textbf{Post-hoc temperature scaling fails under cross-source-city transfer.} Standard practice for calibration repair is to fit a single temperature $T$ on a labeled validation set. We test this directly: per cell, fit $T$ on the source-city validation split (the only labeled data available pre-deployment), apply to the held-out city. The recovery is indistinguishable from no calibration (paired Wilcoxon against zero recovery, $p = 0.19$); on $26$ of $63$ cells, the fitted-on-source temperature actively degrades held-out performance. Coverage thresholding is the only label-free recipe we identify that recovers within-city quality on the held-out city.

\subsection{INRIA Cross-Task Replication}
\label{app:inria}

This appendix details the cross-task replication on INRIA building footprint extraction~\cite{inria}, used in Section~\ref{sec:sp-gap} to test whether the SP-gap findings reflect properties of geographic LOCO or of shadow appearance.

\textbf{Dataset and protocol.} INRIA provides 5{,}000$\times$5{,}000-pixel aerial tiles at $0.3$\,m GSD over five cities; we use Austin, Chicago, and Vienna to match the three-city LOCO structure of \shadow. From each city's labeled tiles we sample $750$ non-overlapping $384\times384$ patches under a fixed seed, split into $450$ training, $150$ validation, and $150$ test images: identical sample sizes to \shadow's per-city splits at high resolution. Patches with more than $30\%$ nodata are rejected before sampling. The three LOCO folds (holdout Austin, Chicago, Vienna) follow the same train-on-two/test-on-one structure as Section~\ref{sec:loco}.

\textbf{Training matched to shadow.} We train MAMNet from scratch using the identical hyperparameters as the shadow MAMNet (Adam $\text{lr} = 10^{-4}$, batch $8$, $100$ epochs, $\text{ReduceLROnPlateau}$ with patience $3$, early stopping patience $10$, ImageNet-pretrained $4$-channel ResNet-$34$ encoder with the contrast channel initialized from RGB mean, MAMNet auxiliary loss with weight $0.4$, $\pm 2$\,px boundary tolerance). No hyperparameter is re-tuned for the new task. The class-balancing weight is recomputed from building label frequency using the same formula applied to shadow frequency. The resulting upper-bound mIoU lands in $[0.74, 0.81]$ across cities, confirming the model is not in a data-starved regime for the task.

\textbf{Replication outcomes.} Pooling the three shadow MAMNet-vanilla cells with the three INRIA MAMNet-vanilla cells gives $n = 6$ for the cross-task test, where the smallest $p$-value any sign test can return is $0.016$. Three findings transfer cleanly at the cell level: the foreground SP gap, the foreground-vs-background asymmetry, and the cross-entropy under-reporting each go in the same direction on all $6$ cells, with cell-level effect sizes Cohen's $d > 1.1$ and sign-test $p = 0.016$ each. The predicted-positive logit erosion transfers in direction on $5$ of $6$ cells (Wilcoxon $p = 0.031$, $d = 1.5$). One signature does not transfer: the direction of class-stratified ECE flips between tasks. On INRIA Vienna, ECE rises by $0.138$ on predicted-negative pixels and only $0.034$ on predicted-positive pixels, the opposite of the shadow pattern.

\textbf{Interpretation.} The class-stratified ECE direction is task-dependent, tracking which class is the population majority (sparse foreground in shadow, dense in INRIA), rather than reflecting a property of geographic LOCO. The other findings depend on properties of the foreground class under shift and replicate across both task semantics. We therefore retain the foreground-asymmetry, under-reporting, logit-erosion, and recoverability claims as properties of geographic LOCO across dense prediction, and explicitly scope the class-stratified ECE result to shadow.

\subsection{Component-Retraining Experiments}
\label{app:locus-retraining}

This appendix details the two controlled retraining experiments underlying Section~\ref{sec:architecture-locus}, which localize the recoverable failure component to the decoder for DINOv3, the encoder for OGLANet, and both for MAMNet.

\textbf{Experiment A (decoder retraining).} For each (architecture, holdout city) pair under vanilla LOCO, we freeze the encoder weights, reinitialize the decoder, and retrain only the decoder for $30$ epochs on a randomly sampled $25\%$ of the holdout-city training set ($\approx 110$ images). The encoder receives no gradient updates. The retrained model is evaluated on the held-out city's $150$ test images by per-image foreground IoU with a boundary-tolerant band.

\textbf{Experiment B (encoder retraining).} The inverse setup: freeze the decoder, reset the encoder to its pretrained initialization (ImageNet for the CNN encoders, original DINOv3 weights for the ViT), and retrain only the encoder on the same $25\%$ subset. The decoder receives no gradient updates.

\textbf{Recovery ratio and exclusion.} Recovery is measured by the per-image ratio $R$. To avoid division by zero on images where the within-city gap is negligible, we exclude images with $\text{IoU}^{\text{upper}} - \text{IoU}^{\text{LOCO}} < 0.02$. Pooling per architecture across the three holdout cities, we obtain $335$--$351$ valid $R$ values per architecture per experiment. Image-cluster bootstrap ($B = 10{,}000$) gives the medians and CIs reported in Section~\ref{sec:architecture-locus}; the error bars of Figure~\ref{fig:diagnostics}(c) come from a separate bootstrap ($B = 1{,}000$) of the same per-image values. Each cell is one training run, and OGLANet values carry the nondeterminism of Appendix~\ref{app:oglanet-dffm}. The pattern, summarized in Table~\ref{tab:locus-summary}, shows the central architecture-dependent locus claim: each architecture is recovered by one experiment and harmed or unaffected by the other.

\begin{table}[h]
\centering
\small
\setlength{\tabcolsep}{6pt}
\caption{\textbf{Recovery ratio $R$ per architecture, both experiments.} Median per-image $R$ with image-cluster bootstrap $95\%$ CI. Decoder retraining (Exp.~A) recovers DINOv3 and MAMNet but not OGLANet; encoder retraining (Exp.~B) recovers MAMNet and OGLANet but actively harms DINOv3.}
\label{tab:locus-summary}
\begin{tabular}{lcc}
\toprule
\textbf{Architecture} & \textbf{Exp.~A: decoder} & \textbf{Exp.~B: encoder} \\
\midrule
DINOv3   & $0.69\ [0.60,\ 0.74]$ & $-0.13\ [-0.27,\ -0.03]$ \\
MAMNet   & $0.67\ [0.63,\ 0.72]$ & $0.88\ [0.83,\ 0.90]$    \\
OGLANet  & $0.12\ [0.05,\ 0.21]$ & $0.86\ [0.80,\ 0.89]$    \\
\bottomrule
\end{tabular}
\end{table}

\textbf{Why DINOv3's encoder retraining harms.} The negative DINOv3 Exp.~B result reflects tight encoder-decoder coupling: the LOCO decoder was trained jointly with a pretrained ViT encoder that barely moved during fine-tuning (whole-encoder relative $L_2$ change of $0.019$ from initialization, Appendix~\ref{app:armL-expA}). Resetting the encoder to its pretrained initialization changes the feature distribution slightly, and the frozen decoder cannot adapt to the new feature space. The performance drops below the LOCO baseline because the decoder is calibrated to a feature space the new encoder no longer produces.

\textbf{Implication for adaptation method choice.} The architecture-dependent locus directly explains the catastrophic failures in Table~\ref{tab:loco-main}. MRFP$^+$ perturbs encoder features uniformly without targeting OGLANet's encoder-locus failure, so it costs OGLANet $-14.0$ mIoU against $-4.9$ on DINOv3. ISW whitens encoder feature covariance, which corrupts the pretrained representation DINOv3 depends on, so it costs DINOv3 $-18.5$ mIoU against $-11.1$ on OGLANet. The architecture-dependent recovery component predicts which adaptation methods will fail catastrophically on a given architecture before the method is run.

\section{Alternative Explanations for the LOCO Gap}
\label{app:alternatives}

This appendix reports the protocol, full results and caveats of the four
tests that address alternative explanations of the gap: training-set scale
(\S\ref{app:ladder}), appearance shortcuts (\S\ref{app:aug-stress}),
fine-tuning of the encoder (\S\ref{app:frozen}), and the pretraining and
fine-tuning regime (\S\ref{app:armL-expA}).  All four use two-class
boundary-tolerant mIoU ($\pm 2$\,px) at high resolution.

\subsection{Training-set scale}
\label{app:ladder}

Increasing the LOCO training set from $225$ to $900$ images narrows the gap by
$0.80$ mIoU ($95\%$ CI $[0.46, 1.14]$) and leaves the upper bound significantly
ahead in every $900$-image cell.

\paragraph{Protocol.} We train the Vanilla LOCO model and the within-city upper
bound of MAMNet, OGLANet and DINOv3 on $225$, $450$ and $900$ training images
(rungs), for three held-out cities and three seeds.  A held-out city contains
$450$ high-resolution training images, so at rung $900$ the LOCO model trains on
$900$ images and its upper bound on $450$; we report that quantity as the
budget-asymmetric gap.  Every model selects its checkpoint on $150$ validation
images at every rung, and every gap is paired per image on the same $150$ test
images per city.  Per-rung and per-architecture intervals resample cells
($B{=}10{,}000$).

\begin{table}[h]
\centering
\caption{\textbf{Gap by rung.} Upper bound minus LOCO, mean over cells with a
cell-level bootstrap $95\%$ CI and the number of cells in parentheses.  Rung
$900$ is the budget-asymmetric gap.  Collapsed models are excluded (see below).}
\label{tab:ladder}
\small
\setlength{\tabcolsep}{4pt}
\begin{tabular}{lccc}
\toprule
& Rung 225 & Rung 450 & Rung 900 \\
\midrule
MAMNet & $7.31$ $[6.01, 8.51]$ (9) & $6.68$ $[4.92, 8.97]$ (9) & $6.48$ $[4.98, 8.18]$ (9) \\
OGLANet & $7.44$ $[4.32, 9.81]$ (8) & $7.31$ $[5.77, 8.88]$ (9) & $5.81$ $[4.74, 7.26]$ (9) \\
DINOv3 & $4.31$ $[3.61, 5.30]$ (7) & $4.15$ $[2.79, 5.63]$ (9) & $4.20$ $[2.38, 6.41]$ (9) \\
\midrule
All & $6.48$ $[5.22, 7.66]$ (24) & $6.04$ $[5.00, 7.20]$ (27) & $5.50$ $[4.51, 6.53]$ (27) \\
\bottomrule
\end{tabular}
\end{table}

\paragraph{Per-cell gaps and contrasts.} Every per-cell gap differs from zero
after Holm correction (largest raw $p = 0.0388$).  All are positive except one
rung-$225$ cell, OGLANet/Phoenix seed 1, where LOCO exceeds the upper bound by
$2.24$ mIoU ($p = 0.003$).  The $225$-to-$900$ contrast is $0.80$ mIoU ($95\%$ CI $[0.46, 1.14]$), the $450$-to-$900$
contrast $0.55$ mIoU ($95\%$ CI $[0.32, 0.77]$), and the
$225$-to-$450$ contrast $0.14$ mIoU ($95\%$ CI $[-0.17, 0.44]$) ($p = 0.41$).
These contrasts pool per-image differences of differences across
(architecture, city, seed) triples and resample images, so their intervals treat
images as independent within a triple.

\paragraph{Collapsed models.} We exclude a gap when its LOCO or upper-bound
model predicts shadow on more than half of the valid region, pooled over the
cell's images.  We set this threshold from the ground truth: shadow occupies at
most $0.086$ of the valid region in any city, so a model above the threshold has
foreground IoU of at most $17.2$.  The rule excludes three rung-$225$ gaps
(OGLANet/Phoenix seed 2, DINOv3/Miami seed 0, DINOv3/Phoenix seed 0) and none at
rungs $450$ or $900$.  With these gaps included, the rung-$225$ mean is $2.12$
($[-4.39, 6.81]$, $27$ cells, $3$ sign changes); the other rungs are unchanged.

\paragraph{Appearance alignment.} The grayscale histogram matching of
\S\ref{sec:feature-space} improves none of the $27$ ladder models it was applied
to.  One of them is the collapsed DINOv3/Miami rung-$225$ seed-$0$ model, whose
matched prediction is degenerate.  This count carries no interval or
$p$-value.

\paragraph{Scope.} The ladder covers high resolution, Vanilla training and
budgets up to $900$ images, the most our three cities supply.

\subsection{Training-time appearance augmentation}
\label{app:aug-stress}

\paragraph{Protocol.} We retrain both the LOCO model and its within-city upper
bound with one strong appearance-only augmentation (colour, illumination,
contrast, blur and noise; geometry preserved) on the $9$ high-resolution
(architecture, held-out city) cells at $450$ training images, with one to three
seeds per side.  The gap here is LOCO minus upper bound, so negative values
mean LOCO is below its upper bound.  Before the grid ran, we fixed the rule: a
cell's gap counts as changed if augmentation moves it by more than twice that
cell's standard deviation of the Vanilla gap over three seeds.

\begin{table}[h]
\centering
\caption{\textbf{Appearance augmentation, per cell.} Gap = LOCO minus upper
bound (mIoU points).  SD: Vanilla gap standard deviation over three seeds.
Threshold $= 2\times$SD.  $|\Delta|$: absolute change of the gap under
augmentation.  Seeds: augmented LOCO/upper.}
\label{tab:aug-stress}
\small
\setlength{\tabcolsep}{4pt}
\begin{tabular}{llccccccc}
\toprule
Arch. & City & Vanilla gap & SD & Aug.\ gap & Threshold & $|\Delta|$ & Seeds & Changed \\
\midrule
MAMNet & Chicago & $-4.52$ & $0.90$ & $-3.76$ & $1.80$ & $0.75$ & 1/1 & no \\
MAMNet & Miami & $-5.68$ & $1.57$ & $-6.33$ & $3.13$ & $0.65$ & 2/1 & no \\
MAMNet & Phoenix & $-9.83$ & $4.26$ & $-5.62$ & $8.52$ & $4.21$ & 3/1 & no \\
OGLANet & Chicago & $-9.08$ & $2.47$ & $-7.64$ & $4.95$ & $1.43$ & 3/2 & no \\
OGLANet & Miami & $-8.33$ & $0.95$ & $-7.06$ & $1.90$ & $1.27$ & 1/1 & no \\
OGLANet & Phoenix & $-4.50$ & $0.24$ & $1.80$ & $0.48$ & $6.30$ & 1/1 & yes \\
DINOv3 & Chicago & $-3.77$ & $0.90$ & $-3.15$ & $1.80$ & $0.62$ & 1/1 & no \\
DINOv3 & Miami & $-1.92$ & $0.84$ & $-3.93$ & $1.68$ & $2.01$ & 1/1 & yes \\
DINOv3 & Phoenix & $-6.76$ & $1.43$ & $-1.25$ & $2.86$ & $5.51$ & 2/1 & yes \\
\bottomrule
\end{tabular}
\end{table}

\paragraph{Result.} Six of nine gaps do not change (Table~\ref{tab:aug-stress}).
The mean gap over the nine cells moves from $-6.04$ to $-4.11$ mIoU, and the
most negative augmented gap is OGLANet/Chicago at $-7.64$.  On OGLANet/Phoenix the
augmented LOCO model exceeds its upper bound by $1.80$ mIoU, against a Vanilla
gap of $-4.50$ with a seed SD of $0.24$.  In the three changed cells, most of the
change comes from the LOCO side (LOCO mIoU $+3.56$, $-2.20$ and $+5.79$ on
OGLANet/Phoenix, DINOv3/Miami and DINOv3/Phoenix).

\paragraph{Robustness of the verdicts.} Rerunning every model that reached the
100-epoch limit with a 200-epoch limit leaves all nine verdicts unchanged; on
DINOv3 the longer run also changes the learning-rate schedule.  Replicating
OGLANet/Phoenix at three seeds per side, one augmented upper-bound seed
collapsed (foreground IoU $1.31$); without it the augmented gap is $+1.43$ and
the cell still counts as changed.

\paragraph{Caveats.} The rule compares against the Vanilla seed spread and
carries no confidence interval, $p$-value or multiplicity correction.  Seed
counts on the augmented side are one to three and unequal.  The test uses one
augmentation strength, so an unchanged gap is not attributable to a weak dose.
Phoenix has the largest Vanilla seed variance of the three cities and also
ranks first on 12 of 20 appearance-distance measures; with three cities, those two
explanations of the Phoenix changes cannot be separated.  OGLANet cells here are
inferred at batch size 1, as are the Vanilla cells they are compared with.
CrossEarth was added to the project after this test ran and is not included.

\subsection{Frozen DINOv3 encoder}
\label{app:frozen}

Freezing the DINOv3 encoder leaves a gap in all three cities, so fine-tuning the encoder on
the source cities is not necessary for the gap. We freeze the DINOv3 encoder
(dinov3\_vits16) and train only the decoder, for both the LOCO model and the within-city
upper bound, on $450$ training images at high resolution with one seed.

\paragraph{Gap.} The upper-minus-LOCO gap is $7.103$ mIoU ($95\%$ CI $[5.876, 8.348]$) in
Chicago, $2.669$ ($[1.729, 3.599]$) in Miami and $2.754$ ($[1.652, 3.939]$) in Phoenix, each
Holm $p < 10^{-4}$ over the three cities, and the city mean is $4.175$ (two-class
boundary-tolerant mIoU at $K{=}2$, 150 paired test images per city). Images with empty
ground-truth foreground (11, 38 and 21 of 150) are retained on both sides of each gap. With
one seed, these gaps carry no seed-to-seed interval, and we do not test them against the
fine-tuned DINOv3 gap of $4.15$ (Appendix~\ref{app:ladder}).

\paragraph{Decoder retraining with a frozen encoder.} Retraining the decoder on target
labels with the encoder held fixed gives per-city median per-image recovery ratios $R$ of
$-2.674$ (Chicago), $-7.087$ (Miami) and $-0.956$ (Phoenix) when the decoder is
reinitialized; in Chicago and Miami the retrained decoder predicted background only and
scored below the LOCO model. Initializing the decoder from the LOCO checkpoint gives
$R = 0.743$, $0.602$ and $0.646$. With the encoder frozen, the decoder is the only trained
component, so recovery through the decoder is expected by construction, and this
experiment does not locate the failure; Appendix~\ref{app:armL-expA} tests the locus. Both
runs differ from Experiment~A of Section~\ref{sec:architecture-locus} in the frozen encoder,
and the second also in the decoder initialization, so their $R$ values are not comparable
with Table~\ref{tab:locus-summary}. Images whose upper-minus-LOCO IoU difference is below
$0.02$ are excluded, leaving 128, 93 and 107 of 150, and per-image $R$ values were not
stored, so the medians carry no interval.

\subsection{Pretraining and fine-tuning regime}
\label{app:armL-expA}

\paragraph{Design.} DINOv3 differs from our CNNs in self-supervised
pretraining and a low fine-tuning learning rate.  We separate the two on MAMNet
with a ResNet-50 encoder in a $2\times2$ design: ImageNet-1k supervised or DINO
self-supervised pretraining, each fine-tuned at learning rate $10^{-4}$ (high) or
$10^{-5}$ (low), for three held-out cities and three seeds at $450$ training
images.  We run Experiments~A (decoder reinitialized and retrained on held-out
labels) and B (encoder retrained) of \S\ref{sec:architecture-locus} on every
cell and report the per-image recovery ratio $R$, pooled over cities and seeds
with an image-cluster bootstrap ($B{=}10{,}000$).  Images whose headroom (upper
minus LOCO) is below $0.02$ are excluded because $R$'s denominator vanishes; this
removes $31\%$ to $41\%$ of images per version.

\paragraph{Experiment B.} Encoder retraining recovers all four versions: median
$R$ is $0.82$ ($95\%$ CI $[0.79, 0.86]$) for supervised/high, $0.85$
($[0.81, 0.88]$) for supervised/low, $0.87$ ($[0.84, 0.90]$) for
self-supervised/high and $0.80$ ($[0.76, 0.83]$) for self-supervised/low.  The
self-supervised/low version recovers less than supervised/high (mean per-image
difference $-0.13$, $[-0.22, -0.05]$, Holm $p = 0.005$) and than
self-supervised/high ($-0.16$, $[-0.24, -0.08]$, Holm $p < 0.001$), and is level
with supervised/low ($-0.04$, $[-0.10, +0.04]$, $p = 0.32$); Holm is over these
three contrasts.  The main effects of pretraining ($-0.003$) and learning rate
($-0.018$) on median $R$ are small.

\paragraph{Experiment A.} Decoder retraining gives median $R$ of $0.64$
($[0.59, 0.67]$) for supervised/high, $-0.14$ ($[-0.24, -0.03]$) for
supervised/low, $0.58$ ($[0.54, 0.60]$) for self-supervised/high and $0.12$
($[0.03, 0.18]$) for self-supervised/low.  The two low-learning-rate versions
recover less through the decoder.  In this design the retrained decoder uses
the cell's own fine-tuning learning rate, and $29$ of the $36$ Experiment~A
cells reach the 30-epoch budget, so the low Experiment~A values may reflect a
slower decoder rather than the regime.

\paragraph{Encoder drift.} The whole-encoder relative $L_2$ change from
initialization has median $0.017$ (supervised/low) and $0.032$
(self-supervised/low), against $0.019$ for the fine-tuned DINOv3 LOCO encoders,
and $0.22$ and $0.38$ for the two high-learning-rate versions.  Drift tracks the
learning rate rather than the pretraining objective.

\paragraph{Caveats.} The design varies the pretraining objective at fixed data:
DINO-ResNet-50 is pretrained on ImageNet-1k ($1.28$M images) and DINOv3 on
LVD-1689M, and no public self-supervised CNN closes that gap, so pretraining
scale remains confounded.  Experiments~A and B use held-out-city labels and are
diagnostic, not a deployable adaptation.  The design covers one architecture.

\section{External Benchmark: S-EO}
\label{app:seo}

The gap appears on S-EO~\cite{seo}, an external benchmark that differs from ShadowTransfer
in sensor (WorldView-3), label source (shadows derived from a digital surface model, DSM)
and cities.

\paragraph{Protocol.} The three S-EO cities are Omaha, Jacksonville and San Diego. For each
held-out city, the LOCO model trains on 3,500 patches from the two other cities and the
within-city upper bound on 3,500 patches from the held-out city. We train MAMNet, OGLANet,
DINOv3 and CrossEarth with three seeds each and evaluate on 1,637 patches from 42 areas of
interest (AOIs) in Omaha, 2,642 from 72 in Jacksonville and 1,870 from 52 in San Diego. The
metric is two-class boundary-tolerant mIoU at $K{=}2$ with pixels the DSM did not observe
excluded. Intervals are AOI cluster bootstrap intervals ($B{=}10{,}000$) with seed as a
stratum, and $p$-values come from AOI-level sign-flip permutation tests with Holm correction
over the 12 (architecture, city) pairs. OGLANet runs at inference batch size 1, because its scores depend on
the inference batch size in our reimplementation.

\paragraph{Gap.} The upper-minus-LOCO gap is positive in all 12 (architecture, city) pairs
(Holm $p = 0.0012$ each), from $2.10$ mIoU (OGLANet, Omaha) to $7.95$ (CrossEarth, Omaha). Pooled within architecture, the gap is $6.98$ ($95\%$ CI
$[6.50, 7.46]$) for MAMNet, $6.93$ ($[6.44, 7.42]$) for CrossEarth, $4.28$ ($[3.97, 4.59]$)
for DINOv3 and $3.45$ ($[3.19, 3.71]$) for OGLANet, and $5.41$ ($[5.07, 5.75]$, 664 AOI
units) over all four. We do not pool or test the S-EO gaps against the ShadowTransfer gaps,
because the two benchmarks differ in sensor, labels, training budget and test unit.

\paragraph{Diagnostics of Section~\ref{sec:evaluation}.} We repeat three diagnostics on
S-EO. (i) Mean predicted shadow probability on ground-truth shadow, $\bar{p}_S$, is lower
under LOCO than under the upper bound in 10 of 12 pairs (Holm $p < 0.05$) and higher in
none, with seed-averaged differences from $-0.1210$ (MAMNet, San Diego) to $+0.0022$
(OGLANet, Omaha). The background control $\bar{p}_B$ also differs in 10 of 12 pairs, so on
S-EO the confidence shift is not confined to the shadow class. The foreground change is
smallest in Omaha, a mean absolute $\Delta\bar{p}_S$ of $0.0161$ against $0.0889$ in the
other two cities. (ii) The foreground-minus-background AURC gap (LOCO minus upper bound;
positive means higher selective risk under LOCO) is positive in 12 of 12 pairs and
Holm-significant in 8, pooling to $+0.0479$ ($[+0.0422, +0.0539]$, 660 AOI units).
(iii) The sign of the decoder-retraining recovery (Experiment~A) separates by architecture
in the same direction as on ShadowTransfer: positive in 18 of 18 MAMNet and DINOv3 cells and
negative in 9 of 9 OGLANet cells, under both foreground-only and two-class differencing. A cell's gap
predicts the magnitude of its recovery ($R^2 = 0.6165$), so gap size is an alternative account
of recovery magnitude; the sign separation holds where the DINOv3 and OGLANet gaps overlap. CrossEarth has no Experiment~A
implementation.

\paragraph{Appearance and sun geometry.} Grayscale histogram matching of target to source
improves 0 of 36 S-EO cells. Regressing the per-patch gap on the patch's sun-elevation
distance to the LOCO training pool, with city and architecture fixed effects, gives a
pooled slope of $-0.189$ with $p = 0.30$ under city-clustered standard errors (3 clusters)
and $p = 0.0013$ under ordinary least squares, which ignores the clustering. The data give no
support to a sun-geometry account, and with three clusters the test has little power.

\paragraph{Adaptation methods.} Of 27 (architecture, method) pairs, 21 score below Vanilla
LOCO and 5 are mixed. MAMNet with MRFP+ fails to train in all nine of its cells, and no
method cell collapses at test time.

\paragraph{\sib{}.} At seed 0, \sib{}'s worst (architecture, city) cell is OGLANet/San Diego at
$61.15$ mIoU, $1.57$ below Vanilla's worst cell, OGLANet/Omaha at $62.72$. \sib{} scores
significantly below Vanilla in 7 of 12 cells and above it in 1 (Holm over 12).

\paragraph{Decomposition of \sib{}'s S-EO deficit.} On a single-seed, two-architecture
subset of the S-EO grid, the DINOv3 \sib{} configuration with FDA sits $1.10$ mIoU below
Vanilla on average over the three cities, and the \sib{} core without FDA sits $0.66$ below.
On MAMNet, adding skip attention gates (SAG) to the core changes the score by $+0.28$. These
values suggest that part of the deficit comes with the FDA component; the grid was not
designed as an ablation, and we did not test this.

\paragraph{Caveats.} S-EO labels are generated from a DSM; both sides of every gap use the
same labels on the same patches, so label noise affects absolute levels more than the gap.
108 test patches (29 in Omaha, 79 in San Diego) are entirely DSM-unseen, score 0 in every
cell of their city, and contribute a difference of exactly 0 to every gap. Five cells
stopped at the epoch ceiling before early stopping triggered. 53 of the 72 Vanilla cells
carry no inference-noise band, so no AURC value has a run-to-run interval.

\section{SIB Implementation Details}
\label{app:sib-impl}

\sib{} sits between the encoder and the decoder of any backbone. Encoder features $F \in \mathbb{R}^{C \times H \times W}$ are decomposed via a single-level 2D Haar wavelet into four subbands $\{F_{LL}, F_{LH}, F_{HL}, F_{HH}\}$, each in $\mathbb{R}^{C \times H/2 \times W/2}$. The Haar transform uses no learned parameters and is perfectly invertible; gradients propagate cleanly through the forward and inverse operations. Compressed subbands are recombined by inverse Haar before being passed to the decoder.

\paragraph{Content VIB on $F_{LL}$.} The content subband carries surface appearance and city-specific texture. We compress it through a per-pixel variational information bottleneck. Two $1{\times}1$ convolutional MLPs map $F_{LL}$ to per-pixel mean $\mu_{LL}$ and log-variance $\log\sigma_{LL}^2$. At training time we sample $Z_{LL} = \mu_{LL} + \epsilon \cdot \sigma_{LL}$ with $\epsilon \sim \mathcal{N}(0,I)$; at inference we use the mean only. The KL penalty against the standard Gaussian prior $p(Z_{LL})=\mathcal{N}(0,I)$ is weighted by an intensity-conditioned coefficient
\begin{equation}
\beta_{\text{content}}(x) = \beta_{\text{base}} + \beta_{\text{scale}} \cdot \sigma\!\left(\text{MLP}(I_{\text{local}}(x))\right),
\end{equation}
where $I_{\text{local}}(x)$ is the mean pixel intensity in a local window around spatial location $x$, computed from the original RGB input and downsampled to the bottleneck resolution. The intensity MLP is a small $1 \to 16 \to 1$ stack with ReLU.

\paragraph{Edge VIB on $F_{LH}$ and $F_{HL}$.} The horizontal and vertical boundary subbands carry shadow geometry. They share a single Edge VIB module with reduced channel width ($C/4$ hidden) and a small fixed weight $\beta_{\text{edge}} = 10^{-4}$ ($10^{-5}$ on MAMNet). The diagonal subband $F_{HH}$ passes through unmodified.

\paragraph{Skip Attention Gates.} For U-Net architectures (MAMNet, OGLANet), each skip connection passes through a channel-wise sigmoid gate computed by global average pooling followed by a $C \to C$ linear layer. Gate biases initialize at $+2.0$ so that sigmoid output starts near $0.88$, making the gate a near-identity at initialization. The model learns to suppress city-specific channels routed around the bottleneck. SAG is not applied for DINOv3.

\paragraph{CNN-only preprocessing.} For CNN architectures, we apply Fourier Domain Adaptation as input-level preprocessing with attenuated spectral radius $\beta_{\text{FDA}} = 0.005$, and append a contrast channel as a fourth input channel computed as the local intensity range over a $3{\times}3$ window. ViT inputs receive neither addition.

\paragraph{Coverage thresholding at deployment.} For each cell, we fit $c^*_{val}$ on the source-city validation split as the largest coverage $c$ at which selective shadow error satisfies $\overline{\text{err}}_{val}(c) \leq (1-\tau) \cdot \overline{\text{err}}_{val}(1.0)$, with $\tau = 0.5$ globally. The fitted $c^*_{val}$ is then applied at inference on the held-out city. The coverage grid is $c \in \{0.10, 0.15, \ldots, 1.00\}$. No held-out labels enter the procedure.

\paragraph{Training loss.} The full training objective combines the boundary-tolerant segmentation loss with the two band-specific KL terms:
\begin{equation}
\mathcal{L}_{\sib} = \mathcal{L}_{\text{seg}} + \mathbb{E}_x\!\left[\beta_{\text{content}}(x) \cdot \text{KL}\!\left(q(Z_{LL}|F_{LL}) \,\|\, p(Z_{LL})\right)\right] + \beta_{\text{edge}} \cdot \text{KL}\!\left(q(Z_{\text{edge}}|F_{\text{edge}}) \,\|\, p(Z_{\text{edge}})\right).
\end{equation}
The KL terms ramp from $0$ to their target weight linearly over the first $10\%$ of training iterations to let the encoder-decoder pair settle before compression activates.

\paragraph{Hyperparameters.} Table~\ref{tab:sib-hparams} lists all hyperparameters. The CNN-specific row is applied for MAMNet and OGLANet only; the ViT row applies to DINOv3. The reported \sib{} runs switch off the intensity adaptation of $\beta_{\text{content}}$ and the content augmentation; Appendix~\ref{app:sib-ablations} tests both against a reference that enables them.

\begin{table}[h]
\centering
\small
\setlength{\tabcolsep}{6pt}
\caption{\textbf{\sib{} hyperparameters of the reported runs.} Values are fixed across all $9$ evaluation cells; per-architecture values are given where they differ.}
\label{tab:sib-hparams}
\begin{tabular}{lll}
\toprule
\textbf{Group} & \textbf{Parameter} & \textbf{Value} \\
\midrule
Content VIB    & $\beta_{\text{content}}$ (constant) & $10^{-3}$ (MAMNet, OGLANet); $10^{-2}$ (DINOv3) \\
               & VIB hidden width             & $C/2$ \\
Edge VIB       & $\beta_{\text{edge}}$        & $10^{-4}$ ($10^{-5}$ on MAMNet) \\
               & VIB hidden width             & $C/4$ \\
SAG (U-Net)    & gate bias init               & $+2.0$ \\
CNN preproc    & FDA spectral radius          & $0.005$ \\
               & contrast window              & $3{\times}3$ \\
Deployment     & $\tau$ (coverage criterion)  & $0.5$ \\
               & coverage grid step           & $0.05$ \\
KL warmup      & fraction of total iters      & $0.10$ \\
\midrule
Optimizer      & AdamW, weight decay          & $0.01$ (OGLANet: Adamax, $10^{-4}$) \\
LR             & MAMNet                       & $1\!\times\!10^{-4}$ \\
               & OGLANet                      & $3\!\times\!10^{-4}$ \\
               & DINOv3                       & $5\!\times\!10^{-5}$ \\
Schedule       & cosine, warmup epochs        & $5$ (OGLANet: reduce on plateau) \\
Batch size     & per GPU                      & $8$ \\
Epochs         & max (early stop on val mIoU) & $100$ \\
\bottomrule
\end{tabular}
\end{table}

\section{Method Comparison by Gap to the Upper Bound}
\label{app:sib-comparison}
\label{app:method-profile}

Table~\ref{tab:method-profile} compares every method by its gap to the
within-city upper bound over the nine cells of \S\ref{sec:sib-results}.

\begin{table}[h]
\centering
\caption{\textbf{Gap to the within-city upper bound for every method.} Nine cells (MAMNet,
OGLANet, DINOv3 $\times$ Chicago, Miami, Phoenix), high resolution, two-class boundary-tolerant
mIoU ($\pm 2$\,px) in percentage points, rows ordered by mean gap. Access: S source-only, U
unlabeled target imagery, D deployment-time. Mean: mean gap with a cell-level bootstrap $95\%$
CI ($B{=}10{,}000$). Worst: largest gap, with its cell. Harm/benefit: cells significantly worse
or better than Vanilla (paired image bootstrap, Holm over the method's cells). Largest harm:
smallest method-minus-Vanilla cell mean. SD: spread of the per-cell gap. \sib{}$-$method:
\sib{}'s worst-case gap minus the method's, with an image-cluster bootstrap $95\%$ CI and
Holm-adjusted $p$ over the nine comparisons; negative values favour \sib{}.}
\label{tab:method-profile}
\setlength{\tabcolsep}{3pt}
\resizebox{\linewidth}{!}{%
\begin{tabular}{llclccccll}
\toprule
Method & Access & $n$ & Mean [95\% CI] & Worst (cell) & Harm/benefit & Largest harm & SD & \sib{}$-$method [95\% CI] & Holm $p$ \\
\midrule
\sib{} & S/U & 9 & $6.42$ $[5.01, 7.96]$ & $10.41$ (OGLANet/Mia) & 2/2 & $-3.79$ & $2.45$ & -- & -- \\
FDA & U & 9 & $6.47$ $[5.34, 7.54]$ & $9.10$ (DINOv3/Phx) & 1/4 & $-6.51$ & $1.83$ & $+1.31$ $[-1.01, +3.65]$ & $0.227$ \\
SegDeSiC & U & 9 & $6.84$ $[4.25, 9.33]$ & $13.20$ (OGLANet/Mia) & 2/3 & $-5.49$ & $4.09$ & $-2.80$ $[-5.16, -0.13]$ & $0.141$ \\
SAR & D & 9 & $6.86$ $[4.81, 9.05]$ & $13.02$ (DINOv3/Chi) & 2/1 & $-0.01$ & $3.52$ & $-2.61$ $[-4.18, -0.16]$ & $0.141$ \\
Vanilla & S & 9 & $6.91$ $[4.83, 9.11]$ & $13.01$ (DINOv3/Chi) & -- & -- & $3.55$ & $-2.61$ $[-4.18, -0.16]$ & $0.141$ \\
BSM & S & 8 & $8.05$ $[5.14, 11.46]$ & $17.40$ (OGLANet/Chi) & 6/2 & $-6.74$ & $4.99$ & $-6.99$ $[-9.01, -4.01]$ & $\le 0.001$ \\
FADA & S & 9 & $9.05$ $[6.98, 11.17]$ & $13.84$ (MAMNet/Mia) & 5/1 & $-9.39$ & $3.44$ & $-3.43$ $[-6.17, -1.24]$ & $0.016$ \\
IIM & S & 9 & $9.75$ $[6.37, 14.01]$ & $24.06$ (MAMNet/Phx) & 4/1 & $-14.54$ & $6.19$ & $-13.66$ $[-16.38, -10.19]$ & $\le 0.001$ \\
MRFP+ & S & 9 & $10.11$ $[6.57, 13.75]$ & $18.59$ (MAMNet/Phx) & 7/1 & $-9.07$ & $5.77$ & $-8.18$ $[-10.86, -5.52]$ & $\le 0.001$ \\
ISW & S & 9 & $13.83$ $[10.22, 17.69]$ & $23.28$ (MAMNet/Phx) & 7/0 & $-13.76$ & $6.16$ & $-12.87$ $[-15.54, -10.45]$ & $\le 0.001$ \\
\bottomrule
\end{tabular}}
\end{table}

\paragraph{Relative gaps.} As a percentage of the upper bound, \sib{}'s mean gap is
$8.52\%$ and FDA's $8.53\%$; their worst-case relative gaps are $14.08\%$ and $13.42\%$.

\paragraph{\sib{} against FDA per cell.} \sib{} is significantly closer to the upper bound than
FDA in 2 of 9 cells, farther in 2, and not separable in 5 (Holm over 9).

\paragraph{Cells that reach the upper bound.} Every method's gap is significant in every cell
except SegDeSiC on DINOv3/Phoenix, the one cell whose method score exceeds its upper bound.

\paragraph{Scope and caveats.} We chose this profile after the pre-committed comparison of
worst-case gaps had been read, so it is exploratory. The grid is the one behind
Table~\ref{tab:loco-main}: the 27 cells that originally trained on 900 images (MAMNet Vanilla
and its six adaptation methods, OGLANet SegDeSiC and OGLANet \sib{}) are 450-image retrains, and
every other cell is the published run. Each cell is one training seed, except MAMNet-\sib{},
whose value is the mean of three seeds; the bootstraps resample test images only. The three
OGLANet-\sib{} cells reached the 100-epoch limit with validation still improving, so they are
lower bounds on the recipe; their FDA donor pool is the held-out city's 450 unlabeled training
images. OGLANet inference depends on batch size and random state through a defect in our DFFM
reimplementation (Appendix~\ref{app:oglanet-dffm}; up to $0.66$ mIoU across seeds on an OGLANet-\sib{} checkpoint), and every
OGLANet value is a single pass. BSM has 8 cells because its DINOv3/Chicago model collapsed. SAR
rows are Vanilla plus SAR's within-pass change, so SAR's harm/benefit counts test only whether
that change differs from zero. The worst-case difference is a difference of two maxima and has no
unbiased bootstrap, so its interval is descriptive. Harm, benefit and significance counts depend
on $\alpha = 0.05$ and the stated Holm families.

\section{Extended \sib{} Component Ablations}
\label{app:sib-ablations}

This appendix extends Table~\ref{tab:sib-ablations} with every ablation run during \sib{} development, $10$ rows per CNN architecture and $8$ for DINOv3. Table~\ref{tab:sib-ablations} reports the five rows that isolate architecture-specific gating (SAG), CNN input preprocessing (FDA), spectral decomposition (Haar), content compression ($F_{LL}$ VIB) and subband targeting (the inverse-evidence row).

\paragraph{Ablation methodology.} Each ablation switches off (or, for the inverse-evidence row, replaces) one design choice of a full-model reference trained in the same batch. Each $\Delta$ is the ablation's two-class boundary-tolerant mIoU minus the reference's, per image, pooled over the three held-out cities at high resolution ($n = 450$ paired images). The interval and $p$ come from a paired image bootstrap that resamples $150$ images within each city ($B = 10{,}000$, two-tailed, $p = \min(1, 2(k+1)/(B+1))$ with $k$ replicates past zero). One bootstrap draw is shared by all tests, so intervals of different rows are correlated. Holm correction runs over all $28$ (row, architecture) tests. $10$ of the $28$ tests sit at the bootstrap floor $2/(B+1) \approx 0.0002$, so their $p$ is resolution-limited.

\paragraph{Reference and training differences.} The batch reference differs from the published \sib{} runs: the reference enables intensity-adaptive $\beta$ and content augmentation, which the published runs switch off, and on MAMNet and OGLANet the reference uses early-stopping patience $15$ where the published runs and every ablation use $25$. Each CNN $\Delta$ therefore also carries this patience change. The published configuration minus the reference ranges from $-1.12$ to $+2.80$ mIoU per (architecture, city) and changes sign across cities within OGLANet and DINOv3. Three rows switch more than one flag. The $-$FDA rows also disable the contrast stage. The MAMNet $-$All VIB row also disables intensity-adaptive $\beta$. MAMNet has no run that removes only the $F_{LL}$ VIB, and its \sib{} has no cross-source mixing.

\paragraph{Training and scoring conditions.} Each cell is one training run with no recorded seed, and the intervals resample test images only. The OGLANet ablations and their reference trained on a $900$/$300$/$150$ train/validation/test pool, not the $450$-image protocol of the main tables. The MAMNet and DINOv3 runs record no training count. OGLANet predictions come from a single fp16 inference pass at batch size $8$ and carry the nondeterminism of Appendix~\ref{app:oglanet-dffm}. The MAMNet and OGLANet runs name the held-out city as FDA donor root. OGLANet's trainer does not pass that root to the dataset, and we did not establish whether the MAMNet runs used held-out-city images as donors. Cells are scored from the stored prediction masks. The checkpoints are no longer available, so no cell could be re-inferred, and no probability-based metric is reported.

\begin{table}[h]
\centering
\footnotesize
\setlength{\tabcolsep}{3pt}
\caption{\textbf{Full \sib{} component ablations.} Mean $\Delta$ mIoU against each architecture's batch reference with $95\%$ CI, pooled over the three held-out cities ($n = 450$ paired images). Stars: Holm-adjusted $p$ over $28$ tests, $^{*}\,p<0.05$, $^{**}\,p<0.01$. N/A: component absent from that architecture's \sib{}. --: not run. Uniform $\beta$: no intensity adaptation. Symmetric VIB: the same $\beta$ on $F_{LL}$, $F_{LH}$ and $F_{HL}$. Edge VIB: the VIB on $F_{LH}/F_{HL}$. $-$Haar: one VIB on the undecomposed $F$. Inv.: VIB on $F_{HL}$ in place of $F_{LL}$. $^{\ddagger}$Collapsed run (text).}
\label{tab:sib-ablations-full}
\begin{tabular}{l|ccc}
\toprule
\textbf{Variant} & \textbf{MAMNet} & \textbf{OGLANet} & \textbf{DINOv3} \\
\midrule
Reference (mIoU) & $73.60$ {\scriptsize$[72.17,\,75.03]$} & $67.79$ {\scriptsize$[66.66,\,68.93]$} & $70.13$ {\scriptsize$[68.91,\,71.33]$} \\
\midrule
$-$ VIB on $F_{LL}$ & -- & $-1.01^{**}$ {\scriptsize$[-1.52,\,-0.50]$} & $-0.71$ {\scriptsize$[-1.30,\,-0.19]$} \\
$-$ All VIB & $+0.29$ {\scriptsize$[-0.33,\,+0.94]$} & -- & -- \\
$-$ Edge VIB & $+0.04$ {\scriptsize$[-0.49,\,+0.60]$} & -- & -- \\
Uniform $\beta$ & $+0.03$ {\scriptsize$[-0.61,\,+0.70]$} & $-1.57^{**}$ {\scriptsize$[-2.11,\,-1.01]$} & $-0.26$ {\scriptsize$[-0.72,\,+0.22]$} \\
Symmetric VIB & $-0.76$ {\scriptsize$[-1.37,\,-0.17]$} & $-0.56$ {\scriptsize$[-1.09,\,-0.02]$} & $-0.27$ {\scriptsize$[-0.78,\,+0.23]$} \\
$-$ Content aug. & $+0.30$ {\scriptsize$[-0.27,\,+0.87]$} & $-1.68^{**}$ {\scriptsize$[-2.21,\,-1.15]$} & $-0.94^{**}$ {\scriptsize$[-1.37,\,-0.51]$} \\
$-$ Cross-source mixing & N/A & $-2.18^{**}$ {\scriptsize$[-2.80,\,-1.57]$} & $-0.48$ {\scriptsize$[-0.95,\,-0.03]$} \\
Aug.\ on all subbands & $+0.16$ {\scriptsize$[-0.38,\,+0.74]$} & $-1.43^{**}$ {\scriptsize$[-2.03,\,-0.81]$} & $-0.01$ {\scriptsize$[-0.46,\,+0.44]$} \\
$-$ SAG & $+0.23$ {\scriptsize$[-0.40,\,+0.87]$} & $-2.06^{**}$ {\scriptsize$[-2.58,\,-1.55]$} & N/A \\
$-$ FDA, contrast & $-4.42^{**}$ {\scriptsize$[-5.37,\,-3.49]$} & $-0.87$ {\scriptsize$[-1.51,\,-0.23]$} & N/A \\
$-$ Haar & $+0.11$ {\scriptsize$[-0.59,\,+0.79]$} & $-0.63$ {\scriptsize$[-1.18,\,-0.08]$} & $-23.49^{**\ddagger}$ {\scriptsize$[-24.83,\,-22.09]$} \\
VIB on $F_{HL}$ (inv.) & $-0.29$ {\scriptsize$[-0.84,\,+0.27]$} & $-1.99^{**}$ {\scriptsize$[-2.53,\,-1.45]$} & $-0.74$ {\scriptsize$[-1.25,\,-0.24]$} \\
\bottomrule
\end{tabular}
\end{table}

\paragraph{Reading the table.} OGLANet shows the most significant rows: removing the content VIB, uniform $\beta$, removing content augmentation, removing cross-source mixing, augmenting all subbands, removing SAG and compressing $F_{HL}$ in place of $F_{LL}$. On MAMNet only the $-$FDA row is significant. On DINOv3 removing content augmentation and removing the Haar decomposition are significant. The DINOv3 $-$Haar runs predict almost no shadow. On Chicago and Miami the predicted-positive fraction is $0.0000$ against ground-truth fractions of $0.0864$ and $0.0802$, $100\%$ and $99\%$ of test images contain no predicted shadow, and foreground IoU is $0.0$ against the reference's $56.1$ and $49.3$. On Phoenix the fraction is $0.0026$ against $0.0347$, and foreground IoU is $0.2$ against $25.8$. This row therefore compares a collapsed training run with the reference.

\section{Coverage Threshold $\tau$ Robustness}
\label{app:tau-sweep}

The coverage thresholding recipe of \S\ref{sec:sib-design} is governed by a single global hyperparameter $\tau$ that sets the source-val error reduction criterion. The main-text reports use $\tau = 0.5$ (halve the source-val full-coverage error). This appendix verifies that the result is stable under perturbation of $\tau$.

\paragraph{Source-val operating-point verification.} Before testing on the held-out city, we confirm that the chosen $c^*_{val}$ achieves the $\tau$-criterion on the source-val split for every cell. Table~\ref{tab:tau-cstar-val} reports per-cell $c^*_{val}$, the source-val full-coverage error, the source-val error at $c^*_{val}$, and the achieved reduction percentage. All $9$ cells achieve at least the $\tau = 0.5$ target ($\geq 50\%$ reduction); reductions cluster slightly above $50\%$ ($50.1\%$ to $60.6\%$) because $c^*_{val}$ lands on the smallest grid value crossing the criterion. No cell falls back to the trivial $c = 1.0$.

\begin{table}[h]
\centering
\small
\setlength{\tabcolsep}{8pt}
\caption{\textbf{Source-val operating-point verification at $\tau = 0.5$.} For each cell: fitted $c^*_{val}$, source-val full-coverage error, source-val error at $c^*_{val}$, and achieved reduction.}
\label{tab:tau-cstar-val}
\begin{tabular}{llccccc}
\toprule
\textbf{Arch} & \textbf{Holdout} & \textit{$c^*_{val}$} & \textit{val err.\ at $c=1$} & \textit{val err.\ at $c^*_{val}$} & \textit{reduction} \\
\midrule
MAMNet  & Phoenix & $0.81$ & $0.239$ & $0.094$ & $60.6\%$ \\
MAMNet  & Miami   & $0.67$ & $0.362$ & $0.165$ & $54.5\%$ \\
MAMNet  & Chicago & $0.72$ & $0.334$ & $0.167$ & $50.1\%$ \\
OGLANet & Phoenix & $0.76$ & $0.315$ & $0.134$ & $57.6\%$ \\
OGLANet & Miami   & $0.62$ & $0.453$ & $0.204$ & $55.0\%$ \\
OGLANet & Chicago & $0.67$ & $0.399$ & $0.178$ & $55.5\%$ \\
DINOv3  & Phoenix & $0.72$ & $0.373$ & $0.163$ & $56.4\%$ \\
DINOv3  & Miami   & $0.57$ & $0.518$ & $0.257$ & $50.5\%$ \\
DINOv3  & Chicago & $0.62$ & $0.456$ & $0.221$ & $51.5\%$ \\
\bottomrule
\end{tabular}
\end{table}

\paragraph{$\tau$ sweep on the held-out city.} We re-fit $c^*_{val}$ at $\tau \in \{0.3, 0.5, 0.7\}$ on the source-val split and apply each fitted operating point to the held-out city. Table~\ref{tab:tau-sweep} reports the population-level test at each $\tau$. All three settings produce $9/9$ cells improving with Wilcoxon $p = 0.00195$ (the $n = 9$ floor under one-sided rank-sum), with the mean held-out reduction $\bar\Delta$ scaling monotonically with the criterion's aggressiveness. No cell falls back to the trivial coverage at any $\tau$.

\begin{table}[h]
\centering
\small
\setlength{\tabcolsep}{8pt}
\caption{\textbf{Population-level coverage recovery across $\tau$.} For each $\tau$: mean held-out selective shadow error reduction across $9$ cells, one-sided Wilcoxon $p$-value (alternative: $\Delta < 0$), cells improving, and median $c^*_{val}$.}
\label{tab:tau-sweep}
\begin{tabular}{ccccc}
\toprule
$\tau$ & \textit{Mean $\bar\Delta$ err.} & \textit{Wilcoxon $p$} & \textit{Cells improving} & \textit{Median $c^*_{val}$} \\
\midrule
$0.3$ & $-0.121$ & $0.00195$ & $9 / 9$ & $0.81$ \\
$0.5$ & $-0.199$ & $0.00195$ & $9 / 9$ & $0.67$ \\
$0.7$ & $-0.264$ & $0.00195$ & $9 / 9$ & $0.57$ \\
\bottomrule
\end{tabular}
\end{table}

\paragraph{Per-cell behavior across $\tau$.} Table~\ref{tab:tau-percell} reports per-cell $c^*_{val}$ and held-out selective shadow error reduction for each $\tau$. The pattern is consistent across cells: smaller $\tau$ yields a less aggressive operating point (larger $c^*_{val}$, smaller magnitude held-out reduction); larger $\tau$ yields a more aggressive operating point (smaller $c^*_{val}$, larger magnitude reduction). The DINOv3-Phoenix cell is the smallest in magnitude across all three settings, consistent with this cell's small within-city to LOCO gap (per Appendix~\ref{app:sib-comparison}); even on this cell, all $\tau$ settings produce a non-trivial reduction.

\begin{table}[h]
\centering
\scriptsize
\setlength{\tabcolsep}{4pt}
\caption{\textbf{Per-cell coverage recovery across $\tau$.} Held-out selective shadow error at $c^*_{val}$ minus error at $c = 1$, with $c^*_{val}$ in parentheses.}
\label{tab:tau-percell}
\begin{tabular}{ll|ccc}
\toprule
\textbf{Arch} & \textbf{Holdout} & $\tau = 0.3$ & $\tau = 0.5$ & $\tau = 0.7$ \\
\midrule
MAMNet  & Phoenix & $-0.046\ (0.91)$ & $-0.099\ (0.81)$ & $-0.128\ (0.76)$ \\
MAMNet  & Miami   & $-0.148\ (0.81)$ & $-0.230\ (0.67)$ & $-0.264\ (0.38)$ \\
MAMNet  & Chicago & $-0.115\ (0.86)$ & $-0.212\ (0.72)$ & $-0.271\ (0.24)$ \\
OGLANet & Phoenix & $-0.085\ (0.86)$ & $-0.131\ (0.76)$ & $-0.171\ (0.67)$ \\
OGLANet & Miami   & $-0.119\ (0.76)$ & $-0.228\ (0.62)$ & $-0.370\ (0.48)$ \\
OGLANet & Chicago & $-0.111\ (0.81)$ & $-0.226\ (0.67)$ & $-0.361\ (0.53)$ \\
DINOv3  & Phoenix & $-0.050\ (0.81)$ & $-0.085\ (0.72)$ & $-0.156\ (0.57)$ \\
DINOv3  & Miami   & $-0.225\ (0.72)$ & $-0.312\ (0.57)$ & $-0.349\ (0.34)$ \\
DINOv3  & Chicago & $-0.187\ (0.76)$ & $-0.268\ (0.62)$ & $-0.301\ (0.29)$ \\
\bottomrule
\end{tabular}
\end{table}

\paragraph{mIoU at threshold $0.5$ is preserved by construction.} Coverage thresholding does not modify the $0/1$ decision boundary at $p = 0.5$; it abstains on low-confidence pixels. The mIoU at threshold $0.5$ reported in Appendix~\ref{app:sib-comparison} is therefore identical for $\sib{}$ at $c = 1$ and at any $c^*_{val}$ in the tested range. The $\tau$ sweep characterizes the magnitude of selective-error recovery; deployment selects $\tau = 0.5$ as a single fixed global hyperparameter.

\end{document}